\pdfoutput=1
\documentclass
[11pt]{article}
\usepackage{CJKutf8}
\usepackage[final]{acl}
\usepackage{times}
\usepackage{latexsym}
\usepackage{algorithm}
\usepackage{algorithmic}
\usepackage{amsfonts}
\usepackage{amssymb}
\usepackage{amsmath}
\usepackage{graphicx}
\usepackage{bm}
\usepackage{multirow}
\usepackage{multicol}
\usepackage{dashrule}
\usepackage{arydshln}
\usepackage{booktabs}
\usepackage{framed}
\usepackage{enumitem}
\usepackage{xcolor}
\usepackage[T1]{fontenc}
\usepackage[utf8]{inputenc}

\usepackage{microtype}

\usepackage{inconsolata}

\usepackage{amsthm}
\usepackage{amsmath}
\newtheorem{insight}{Insight}
\usepackage[inkscapelatex=false]{svg}

\title{
    Balancing Speciality and Versatility: A Coarse to Fine Framework for Mitigating Catastrophic Forgetting in Large Language Models
}

\renewcommand\footnotemark{}
\author{
Hengyuan Zhang\textsuperscript{\rm1\dag}\thanks{\dag\ This work was done during internship at SenseTime Research.},
Yanru Wu\textsuperscript{\rm1},
Dawei Li\textsuperscript{\rm3},
\textbf{Sak Yang}\textsuperscript{\rm4}, 
\textbf{Rui Zhao}\textsuperscript{\rm2},\\
\textbf{Yong Jiang\textsuperscript{\rm1*}, 
Fei Tan\textsuperscript{\rm2*\$}}\thanks{*\ Corresponding author}\thanks{\$\ Project lead}  \\
\textsuperscript{1}Tsinghua University \quad
\textsuperscript{2}SenseTime Research \\ 
\textsuperscript{3}University of California, San Diego \quad \textsuperscript{4}Independent Researcher\\
\texttt{\{zhang-hy22,wu-yr21\}@mails.tsinghua.edu.cn} \quad dal034@ucsd.edu \\ sakyang@outlook.com \quad \texttt{\{zhaorui,tanfei\}@sensetime.com}
}
\begin{document}
\maketitle
\begin{abstract}
% Aligned Large Language Models (LLMs) have demonstrated remarkable \textit{versatility}, capable of handling diverse real-world tasks. To better enhance the \textit{speciality} of aligned LLMs for specialized applications, a typical practice is to fine-tune them with extra supervised data. 
% However, fine-tuning often leads to catastrophic forgetting (CF) of previously acquired versatility, hindering the model’s performance on various tasks. 
% In response to this challenge, we propose \textit{CoFiTune}, a coarse to fine framework in an attempt to strike the balance between speciality and versatility. 
% Specifically, at the coarse-grained level, an empirical tree-search algorithm is utilized to pinpoint and update specific modules within a roughly predefined layer range that are crucial for speciality, while keeping other parameters frozen to maintain versatility; at the fine-grained level, the framework introduces a soft-masking mechanism to regulate the update to the LLMs, further mitigating the CF issue without harming speciality. 
% In an overall evaluation of both speciality and versatility, \textit{CoFiTune} consistently outperforms baseline methods across diverse tasks and model scales. 
% We also conduct further experiments to provide more insights in this field.
% The code is available at \url{https://anonymous.4open.science/r/CoFiTune-542C}.
Aligned Large Language Models (LLMs) showcase remarkable \textit{versatility}, capable of handling diverse real-world tasks.
Meanwhile, aligned LLMs are also expected to exhibit \textit{speciality}, excelling in specific applications.
However, fine-tuning with extra data, a common practice to gain speciality, often leads to catastrophic forgetting (CF) of previously acquired versatility, hindering the model’s performance across diverse tasks.
In response to this challenge, we propose \textit{CoFiTune}, a coarse to fine framework in an attempt to strike the balance between speciality and versatility. 
At the coarse-grained level, an empirical tree-search algorithm is utilized to pinpoint and update specific modules that are crucial for speciality, while keeping other parameters frozen; at the fine-grained level, a soft-masking mechanism regulates the update to the LLMs,  mitigating the CF issue without compromising speciality. 
In an overall evaluation of both speciality and versatility, \textit{CoFiTune} consistently outperforms baseline methods across diverse tasks and model scales. 
% Compared to the full-parameter SFT, \textit{CoFiTune} leads to about an average $14\%$ versatility improvement and marginal speciality loss across diverse tasks and model scales. 
When compared to the full-parameter SFT, \textit{CoFiTune} offers an average versatility improvement of 14\%, while only incurring a marginal loss in speciality.
Lastly, based on further analysis, we provide a speculative insight into the information forwarding process in LLMs, which helps explain the effectiveness of the proposed method.
 The code is available at \url{https://github.com/rattlesnakey/CoFiTune}.

\end{abstract}

\section{Introduction}
\label{sec:intro}
Aligned LLMs mainly undergo a two-step procedure: initial pre-training on web-scale text corpora, followed by fine-tuning on diverse instructions to align with human intentions. 
% They exhibit remarkable \emph{versatility}, showcasing their ability to handle various real-world tasks, especially those of high complexity, such as reasoning, knowledge acquisition, and instruction following~\citep{zhao2023survey,achiam2023gpt}.
They exhibit remarkable \textbf{versatility}, \emph{showcasing their ability to handle various real-world tasks, such as reasoning, common sense question-answering, and instruction following}~\citep{zhao2023survey,achiam2023gpt,lu2022makes}.
% ,achiam2023gpt

% Despite the versatility, aligned LLMs still fall short in certain \textbf{speciality} including mathematics~\citep{gou2023tora}, finance~\citep{li2023large}, and law~\citep{cui2023chatlaw}.
% This limitation impedes their application in varied professional fields and is commonly addressed through the practice of fine-tuning.
Despite the versatility, aligned LLMs still \emph{fall short in certain tasks or domains, such as mathematics}~\citep{gou2023tora}, \emph{finance}~\citep{li2023large}, \emph{and law}~\citep{cui2023chatlaw}.
To bolster performance in these particular tasks or domains, i.e., to gain \textbf{speciality}, a common practice is fine-tuning.
% To enhance its task or domain performance, i.e., gaining \textbf{speciality}, a common practice is fine-tuning. 
\begin{figure}[!t]
    \centering  \centerline{\includegraphics[width=\columnwidth]{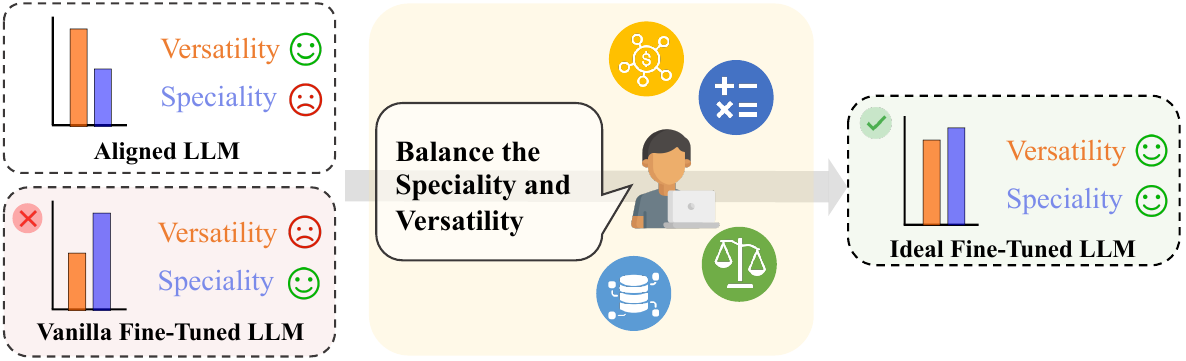}}
    \caption{An illustration of our objective: achieving effective speciality without significantly compromising versatility.}
    % \vspace{-0.1cm}
    \label{fig:intro}
    \vspace{-0.4cm}
\end{figure}
% \begin{figure}[ht]
%     \centering
%     \centerline{\includesvg[width=1\columnwidth]{figs/intro.svg}}
%     % \includegraphics[width=\columnwidth]{}
%     \caption{\textbf{The Trade-Off between Model Speciality and Versatility.} }
%     \label{fig:intro}
% \end{figure}
% 在这边讲 versatility 损害的危害
% The loss in versatility not only hampers the practical application of aligned LLMs for improved user service but also hinders the subsequent specialty learning process from leveraging it effectively.
% This loss of versatility hinders the real-world application of aligned LLMs to serve the users better and also hinders the process leveraged by the subsequent speciality learning~\citep{jie2022alleviating,luo2023investigating}.
However, during the fine-tuning process, the modification of model parameters often leads to catastrophic forgetting (CF), thereby causing a noticeable loss of versatility \citep{lin2023speciality}.
This loss adversely affects the performance of fine-tuned models across various real-world tasks~\citep{cheng2023adapting, dong2023abilities}, propelling several works to investigate and contribute solutions to the CF in LLM versatility~\citep{lin2023speciality,wang2023trace}.
% This loss not only inhibits the performance of fine-tuned model on diverse tasks~\citep{cheng2023adapting,dong2023abilities} but also hinders the effectiveness of transferring knowledge when gaining speciality~\citep{jie2022alleviating,luo2023investigating}, propelling several works to investigate and contribute solutions to the CF in LLMs~\citep{lin2023speciality,wang2023trace}.
% panigrahi2023task,
% learning from it.
% learning  subsequent speciality learning process from effectively leveraging it.
% This loss not only limits the practical use of aligned LLMs for better user service but also hinders the subsequent speciality learning process from effectively leveraging it.
% Several works have pioneered in investigating and contributing solutions to the CF in LLMs~\citep{panigrahi2023task,lin2023speciality,wang2023trace}. 

As a relatively new problem, the CF in LLMs remains under-explored. We categorize the existing studies into regularization-based, weight-based, and architecture-based methods\footnote{See Appendix~\ref{sec:baselines_descriptions} for more details about these methods.}.
Regularization-based methods~\citep{lin2023speciality,smith2023continual} add extra terms into loss function to penalize parameter changes.
% Approaches belonging to the former category share the consensus that, to mitigate CF, the fine-tuned model should closely align with the pre-trained model~\citep{wortsman2022robust,lin2023speciality}.
% Weight-based methods~\citep{wortsman2022robust,ke2022continual} aim to constrain the parameter update of the pre-trained model via a weighted mechanism.
Weight-based methods~\citep{wortsman2022robust,ke2022continual} introduce weight coefficients for parameters to regulate their updates.
% design a weight coefficient for parameters to regulate their updating.
% Weight-based methods~\citep{wortsman2022robust,ke2022continual} update the parameters of the original model using a weighted mechanism.
% align the fine-tuned model closely with the pre-trained model via a weight 
% \citet{wortsman2022robust} proposed the Wise-FT strategy, which interpolates the pre-trained and fine-tuned model parameters to derive the mixed model.
% \citet{lin2023speciality} added L1 and L2 regularization respectively as penalties to minimize the parameter differences between the fine-tuned and pre-trained models.
% as penalties to the model parameter discrepancy.
% Architecture-based methods~\citep{wang2023rehearsal,razdaibiedina2023progressive} focus on designing extra modules outside the original model to mitigate interference.
% Architecture-based methods~\citep{wang2023rehearsal,razdaibiedina2023progressive} focus on designing extra modules outside the original model, fine-tuning them exclusively.
% wang2023orthogonal
Architecture-based methods~\citep{wang2023rehearsal,razdaibiedina2023progressive} design and exclusively fine-tune extra modules outside the original model.
% The latter category aims to constrain the parameter update of the pre-trained model~\citep{hu2021lora,ke2022continual}.
% \citet{ke2022continual} limited the gradients of parameters during the fine-tuning process based on their corresponding importance.
% While \citet{hu2021lora} inserts additional low-rank matrix (LoRA) to the model without updating the original parameters of LLMs. 
% However, these methods are limited in either of the two aspects: 
However, these methods still have limitations:
1) Regularization- and weight-based methods involve fine-tuning all parameters, which poses a significant challenge to preventing CF in versatility. Particullary,  CF deteriorates as the training iteration proceeds~\citep{luo2023empirical}.
% deteriorates~\citep{luo2023empirical}.
% CF becomes severer~\citep{luo2023empirical}.
% especially given the accumulation of CF with the growth of iteration number. 
2) Architecture-based methods only update the inserted parameters, which blocks the learning of speciality, particularly on challenging tasks~\citep{lin2023speciality}.
These limitations raise a key question: 
\emph{How can the model gain speciality while preserving the versatility, thereby enhancing its overall abilities?} (shown in Fig.~\ref{fig:intro}).
% These limitations require a
% method to gain speciality while preserving the versatility, thereby enhancing the overall model abilities (shown in Fig.~\ref{fig:intro}).
% 开始分析
% 感觉 redundancy 好像和 importance 是差不多概念的？
% Recent studies reveal the presence of redundant parameters in the model and variations in importance across different modules.
Recent research threads highlight two pivotal findings: the presence of redundant parameters in the model~\citep{bansal2023rethinking,sun2023simple} and the distinct role played by different modules and layers~\citep{geva2021transformer,meng2022locating}.
These suggest that it is highly feasible to enhance speciality by updating specific parameters within a defined layer range while keeping the remainder frozen to maintain versatility.\footnote{The results in Sec. \ref{sec:main_results} and \ref{sec:further_analysis}, along with the insights \ref{insight:1}, \ref{insight:3}, and \ref{insight:4}, further support our speculation.}

% \footnote{The results in Sec. \ref{sec:main_results} and \ref{sec:further_analysis}, along with the insights in \ref{insight:1}, \ref{insight:3}, and \ref{insight:4}, further support our speculation.}.
% We experimentally explore the fine-tuning of different modules and layer ranges (results and discussions in Sec. \ref{sec:main_results} and \ref{sec:further_analysis}), and the findings, concluded in Insight \ref{insight:1}, \ref{insight:3} and \ref{insight:4}, further support our speculation.
% These indicate that, a method is highly feasible to effectively gain speciality by updating specific parameters within a defined layer range, while freezing the rest to preserve versatility.
% Recent works have proved the presence of redundant parameters and the different importance of parameters in the model, which indicates that a method to update specific parameters in the model to gain speciality effectively while freezing its rest of parameters to preserve versatility, thereby enhancing the overall abilities is high feasible.
% we argue these limitations require a method to update specific parameters in the model to gain speciality effectively, due to the redundancy and distinct function of parameters for speciality, while freezing its rest of parameters to preserve versatility, thereby enhancing the overall abilities (shown in Fig.~\ref{fig:intro}).
% 这边是不是写一些观点来支撑这个论点
% These challenges necessitate a
% method to gain speciality while preserving the versatility,
% thereby enhancing the overall model abilities (shown in Fig.~\ref{fig:intro}).
Therefore, we propose a \underline{\textbf{Co}}arse to \underline{\textbf{Fi}}ne framework, i.e., \emph{CoFiTune}.
At the coarse-grained level, we perform an empirical tree-search algorithm to identify and update specific modules within a defined layer range that are crucial for speciality without significant versatility penalty.
Simultaneously, the remaining parameters are frozen to further preserve versatility.
% Subsequently, recognizing that not all neurons within the module identified at the coarse level affect versatility equally
Subsequently, recognizing that not all units\footnote{\emph{Units} denote the attn. heads in MHA or neurons in FFN.} in a module are evenly important for the versatility~\citep{michel2019sixteen}, we utilize a fine-grained soft-masking mechanism to regulate the backward gradient flow based on their importance values for versatility.
This further mitigates the CF issue while not jeopardizing the speciality.
To summarize, our contributions are as follows:

\vspace{0.2em}
\quad \textbf{1)} We present \emph{CoFiTune}, a framework striking a delicate balance between versatility and speciality. We also lead in creating a comprehensive Chinese CF setting, contributing to CF research in the Chinese domain. Extensive experiments demonstrate the effectiveness of \emph{CoFiTune} across diverse tasks and model scales.
% \quad \textbf{1)} We present \emph{CoFiTune}, a framework striking a delicate balance between versatility and speciality. We also lead in creating a comprehensive Chinese CF setting, contributing to CF research in the Chinese domain. Extensive experiments demonstrate \emph{CoFiTune}'s effectiveness across diverse tasks and model scales, achieving over 95\% versatility and 90\% speciality compared to original and full SFT models on average.

\vspace{0.4em}
\quad \textbf{2)} Our \emph{CoFiTune} achieves over 95\% versatility and 90\% speciality compared to original and full SFT models on average, and we find that only tuning the FFN module in the mid-to-lower layer range achieves satisfactory speciality without significantly affecting versatility.

\vspace{0.4em}
\quad \textbf{3)} We conduct extra experiments to give more insights, demonstrating the module's importance for gaining speciality and exploring the function of three crucial areas within the model.

\section{Related Work}
\label{sec:related_work}
\paragraph{CF in LLM}
% CF~\cite{} has been wildly observed on deep neural networks in continual learning~\cite{kirkpatrick2017overcoming,seff2017continual} or domain adaptation~\cite{ke2022continual} scenarios.
% The previous study about the CF problem typically focused on the CF of previous training tasks under a continual learning manner that learns a sequence of tasks~\cite{seff2017continual,madotto2020continual,wu2021pretrained}.
The previous studies about the CF problem typically focused on the CF of prior training tasks in a continual learning context, where a sequence of tasks are learned~\cite{madotto2021continual,wu2021pretrained}.
% Recently, several studies have focused on the CF problem in the general abilities of LLMs.
Recently, attention has shifted towards investigating the CF problem in the general abilities of LLMs.
\citet{luo2023empirical} first explored the CF problem in LLMs, revealing a pronounced CF phenomenon in general knowledge and reasoning.
% \citet{luo2023empirical} first explored the CF problem in LLMs under a continual learning manner, revealing a pronounced CF phenomenon in general knowledge and reasoning.
Subsequently, \citet{wang2023trace} established a benchmark comprising a sequence of challenging tasks to facilitate CF research in LLMs.
% Subsequently, \citet{wang2023trace} established a benchmark comprising a sequence of challenging tasks, including domain-specific tasks and mathematical reasoning, to facilitate CF research in LLMs.
Different from them, \citet{lin2023speciality} proposed several solutions to address the trade-off between versatility and speciality in LLMs under a single task fine-tuning setting.
In this study, we follow \citet{lin2023speciality} to strike a balance between the speciality and versatility of LLMs.
% While all the aforementioned studies have been centered around the English domain, our work pioneers the establishment of a Chinese CF setting, following the setting proposed by \citet{lin2023speciality} to 
% follow \citet{lin2023speciality} to underscore the trade-off between the generalized and specialized abilities of LLMs.
% All the above works are focused on the English domain, in this work, we first establish a Chinese CF setting and follow \citet{lin2023speciality} to underscore the trade-off between the generalized and specialized abilities of LLMs.
% \citet{fu2023specializing} demonstrate the trade-off between LLMs' multi-dimensional abilities and successfully adapt smaller LLMs (<10B in parameter size) to a domain-specific model by paying the price of decreased generic ability.
% \citet{dong2023abilities} study the effectiveness of supervised fine-tuning to the LLMs' performance and devise a dual-stage mixed fine-tuning (DMT) strategy to keep the general capability of LLMs by adding an additional learning stage.
% There are also other works that present the effectiveness of various low-rank adaptation (LoRA) methods~\cite{wang2023orthogonal,dou2023loramoe} in addressing the CF issues in LLMs.
\paragraph{Key Components in Transformer}
\label{para:key_comp}
% (FFN, attn role, 66b ICL, knowledge editing (only edit some param))
% 2.important param and layer in Transformer (FFN, attn role, 66b ICL, knowledge editing (only edit some param))
% todo: adatper blog
Previous works demonstrate the presence of substantial redundant parameters in Transformer-based models~\citep{an2020repulsive,xia2022structured,bansal2023rethinking}.
% Interpret models' mechanisms correctly can boost efficient improvement and revision of them.
Therefore, it's crucial to pinpoint the key components and precisely comprehend their underlying mechanisms.
% Recent endeavors try to analyze key layers and modules in it.
Recent endeavors try to analyze the involved layers and modules.
\citet{elhage2021mathematical} developed mathematical tools to unveil an information-copying behavior within the attention module.
\citet{geva2021transformer} analyzed the feed-forward network (FFN), considering the up and down projection matrices as the key and value of the memories. 
% and viewed the up and down projection matrices inside it as the key and value of the memories.
% 在 LLM 时代，trainable activation SwiGLU is used prevelantly, xxx demonstrate
\citet{mirzadeh2023relu} demonstrated that SwiGLU, a common trainable activation function in the FFN of LLMs, can be replaced with non-trainable ReLU to reduce computation without affecting performance.
% SwiGLU is a common trainable activation function within the FFN of LLMs. 
% However, \citet{mirzadeh2023relu} showed that replacing it with non-trainable ReLU reduces computation without affecting performance.
% However, \citet{mirzadeh2023relu} demonstrated that replacing it with non-trainable ReLU significantly reduces computation without sacrificing performance.
% Trainable activation function SwiGLU has been prevalantly used in LLM, while zhang demonstrate that replacing it with ReLU (without trainblae parameter) significantly reduces computation while not harming performance
% \citet{mirzadeh2023relu} demonstrate that replacing the SwiGLU with ReLU (without trainblae parameter) activation function significantly reduces computation while not harming performance, which indicates that the SwiGLU
% as for the trainable activation SwiGLU, xx demonstrate 
% 有人尝试了不同的 activation SwiGLU 
% 有人尝试了不同的 activation , which indicate that the activation has small influence on performance.
% Several works~\cite{li2021bert,meng2022locating,geva2023dissecting} scrutinized various variants of the Transformer, providing unique insights into each specific model.
% scrutinize diverse variants of the Transformer, offering distinct insights into each specific model.
% 我们没有 borrow，而是提供 insights
In light of the aforementioned insights, in this work, we explore the key components, i.e., attention and FFN\footnote{For simplicity, starting from this section, FFN only refers to the combination of up \& down projection.} modules in certain layer range that are integral to gaining speciality and preserving versatility.
% In this study, we follow the aforementioned works to explore the key components, i.e., attention and FFN modules within the certain layer range that are crucial for gaining speciality and preserving versatility.
% also provide our insights about the layers and modules that that we find in the corase level selection process 
% borrow the insight of hierarchically analyzing the layers and functions of LLMs that play important roles in storing the general and specific capabilities.
% In this study, we also borrow the insight of hierarchically analyzing the layers and functions of LLMs that play important roles in storing the general and specific capabilities.
% Additionally, by strategically positioning the knowledge stored in specific layers of models, several works~\cite{de2021editing,dai2022knowledge} propose efficient knowledge editing approaches for Transformer-based models~\cite{zhu2020modifying}.
% Additionally, several works propose efficient knowledge editing approaches for Transformer-based models~\cite{zhu2020modifying}.
% Furthermore, various works propose efficient knowledge editing approaches for Transformer-based models~\cite{zhu2020modifying}. In their work, \cite{de2021editing} develops a knowledge editor (KE) hyper-network designed to fine-tune a model by incorporating new facts derived from textual descriptions.
% \citet{dai2022knowledge} suggests that individual neurons within MLP layers encode distinct facts and introduce an attribution method for identifying neurons associated with a specific fact.
% % add a recent paper

\section{The Framework}
We start with the task formulation (Sec.~\ref{sec:task_formulation}) and the backbone we used in this work (Sec.~\ref{sec:backbone_arch}).
Then, we outline our \emph{CoFiTune} framework, covering: 1) an empirical tree-search algorithm to identify modules within a defined layer range that balances speciality and versatility effectively (Sec.~\ref{sec:coarse_level}); 2) a Fine-SoftMask mechanism regulating the backward gradient flow based on units' importance for versatility to further mitigate the CF issue (Sec.~\ref{sec:fine-grained_level}). 
\subsection{Task Formulation}
\label{sec:task_formulation}
Different from earlier researches, which focus on the CF in previously learned abilities when fine-tuning on a sequence of tasks~\citep{liu2021continual,zhang2022continual}, we align with \citet{lin2023speciality}, emphasizing the trade-off between speciality and versatility in fine-tuning original LLM for a single task.
% This is because the original versatility of LLM is more crucial and will be progressively undermined at each fine-tuning step.
This is because the original versatility of LLM is crucial and undergoes a progressive reduction at each fine-tuning step~\citep{luo2023empirical}.
Additionally, we investigate a potential strategy to replace sequential fine-tuning for better performance.
% We present our discussion of the setting in Appendix~\ref{sec:setting_discussion}.
% We further discuss our setting in Appendix~\ref{sec:setting_discussion}.
Further discussion of our setting is in Appendix~\ref{sec:setting_discussion}.

During supervised fine-tuning (SFT), given an input token sequence $\bm{x}=(x_0, x_1, \ldots)$, the model is trained to predict the next token $x_i$ in an autoregressive manner:
% in an autoregressive manner
% \begin{align}
%     \small \mathcal{L}_{\textrm{SFT}}(\Theta) &= \mathbb{E}_{\bm{x}\sim\mathcal{D}_{\textrm{SFT}}} \left[ -\sum_{i}\log p(x_i|x_0,\ldots,x_{i-1};\Theta)\right]
% \end{align}
% \begin{equation}
%     \label{eq:sft_loss}
%     % \resizebox{.89\hsize}{!}{$\mathcal{L}_{\textrm{SFT}}(\Theta) = \mathbb{E}_{\bm{x} \sim\mathcal{D}_{\textrm{SFT}}} \left[ -\sum_i \log p(x_i|x_0,\ldots,x_{i-1};\Theta)\right]$}
%     \begin{aligned}
%         &\mathcal{L}_{\textrm{SFT}}(\theta) = \\
%         &\mathbb{E}_{\bm{x} \sim\mathcal{D}_{\textrm{SFT}}} \left[ -\sum_i \log p(x_i|x_0,\ldots,x_{i-1};\theta)\right]
%     \end{aligned}    
% \end{equation}
\vspace{-0.3cm}
\begin{equation}
\vspace{-0.1cm}
\hspace{-2mm}
    \label{eq:sft_loss}
    \mathcal{L}_{\textrm{SFT}}(\theta) = \mathbb{E}_{\bm{x} \sim\mathcal{D}_{\textrm{SFT}}} \left[ -\sum_i \log p(x_i|x_{<i};\theta)\right]
\end{equation}
\vspace{-0.3cm}

% The loss is only calculated on the \emph{\{output\}} part of the input sequence and can be expressed as:
\noindent where $\theta$ represents the parameters of aligned LLM,  $\mathcal{D}_{\textrm{SFT}}$ is the fine-tuning dataset.

% After obtaining the fine-tuned model $\hat{\theta}$, we assess its speciality (Spec.) and versatility (Vers.) scores.
% A unified (Uni.) score is further defined to comprehensively evaluate its overall ability, i.e., both Spec. and Vers.
After obtaining the fine-tuned model $\hat{\theta}$, we assess its speciality (Spec.) and versatility (Vers.) scores.
A unified (Uni.) score is further defined to evaluate its overall ability, i.e., both Spec. and Vers..
Refer to Sec.~\ref{sec:metrics} for more details of scores.

\subsection{Backbone Architecture}
\label{sec:backbone_arch}
In this work, we utilize the Llama-based~\citep{touvron2023llama} aligned LLM as our backbone, which consists of an Embedding module, an LM Head, and a stack of Llama Layers.
Each Llama Layer integrates a multi-head attention (MHA), followed by a feed-forward network (FFN) with residual connections.
For an input $\mathbf{x}$, the Llama Layer generates output $\mathbf{y}$ based on the following equations:

% \vspace{-0.3cm}
\begin{equation}
\label{eq:llama_block}
\begin{aligned}
\mathbf{x}' &= \mathbf{x} + \text{MHA}(\text{Norm}(\mathbf{x})) \\
\mathbf{y} &= \mathbf{x}' + \text{FFN}(\text{Norm}(\mathbf{x}')) 
\end{aligned}
\end{equation}
Each head in the MHA module can be defined as: 

\vspace{-0.5cm}
\begin{equation}
\text{head}_i = \text{Attention}(\mathbf{x}\mathbf{W}^Q_i, \mathbf{x}\mathbf{W}^K_i, \mathbf{x}\mathbf{W}^V_i) \label{eq:attn}
\end{equation}
\vspace{-0.5cm}

\noindent where $\mathbf{W}^Q_i$, $\mathbf{W}^K_i$, and $\mathbf{W}^V_i$ are the weight matrices of query, key, and value for the $i$-th head. The FFN module is parameterized by an up projection (up\_proj) and a down projection (down\_proj):

\vspace{-0.6cm}
\begin{align}
    \text{FFN}(\mathbf{x}') =  \sigma(\mathbf{x}'\mathbf{W}^1) \mathbf{W}^2\label{eq:FFN}
\end{align}
\vspace{-0.6cm}

\noindent where $\sigma$ is the SwiGLU activation function~\citep{shazeer2020glu}, $\mathbf{W}^1$ and $\mathbf{W}^2$ are up\_proj and down\_proj respectively.
More details of backbone information are illustrated in Appendix~\ref{sec:swiglu_rmsnorm}.

\begin{figure}[!t]
    \centering
    \centerline{\includegraphics[width=\columnwidth]{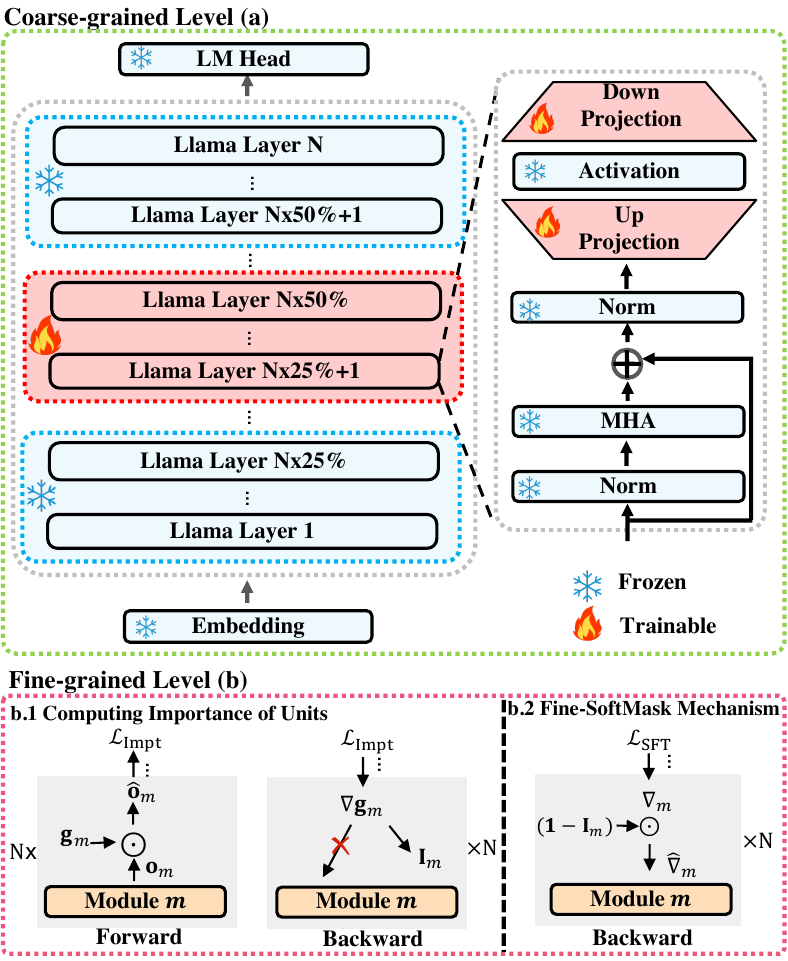}}
    % \vspace{-0.3cm}
    \caption{An illustration of our \emph{CoFiTune} framework. $N$ denotes the number of layers. At the coarse-grained level, we pinpoint the module (e.g., FFN) within a defined layer range (e.g., 10th - 20th layers) that gains speciality effectively without harming versatility much. At the fine-grained level, we selectively update the parameters within the region identified at the coarse-grained level and leverage $\mathbf{I}_m$ to control their gradient flow.}
    \label{fig:overview}
    \vspace{-0.4cm}
\end{figure}
\subsection{Coarse-grained Level}
\label{sec:coarse_level}
\begin{figure*}[!t]
% \vspace{-0.1cm}
    \centering
 \centerline{\includegraphics[width=1.4\columnwidth]{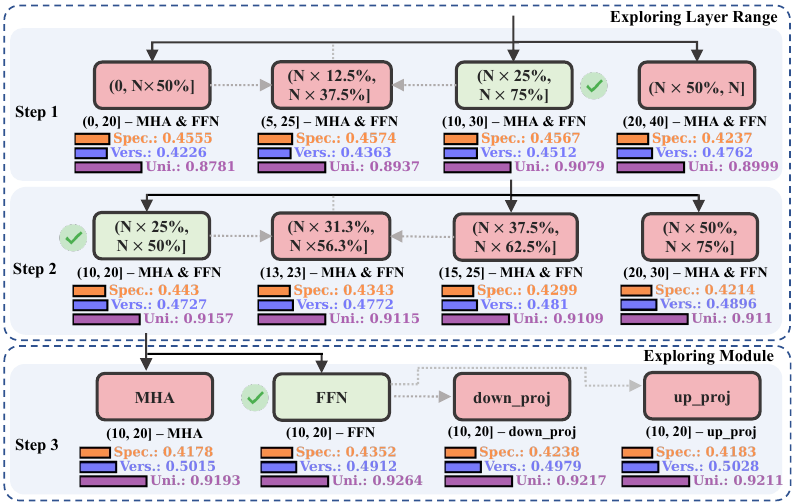}}
    \caption{The exploration process of Finance task in the 13B model. $N$ denotes the number of layers and in this case, $N=40$. For simplicity, we denote the model fine-tuned in our exploration as ``layer range - module'', e.g., the model fine-tuned with FFN module within the layer range $(10, 20]$ denoted as ``$(10, 20]$ - FFN''.}
    \label{fig:search}
    \vspace{-0.4cm}
\end{figure*}

Following the analysis in paragraph~\ref{para:key_comp} of Sec.~\ref{sec:related_work}, which reveals the presence of parameter redundancy and key components (MHA and FFN) in LLMs, we aim to pinpoint the module (e.g., FFN) within a defined layer range (e.g., $10$th - $20$th layers) target for speciality without significantly affecting versatility.
Subsequently, we exclusively train the identified module within the layer range, while freezing the remaining parameters to further preserve the versatility.

Our exploration process encompasses three steps: 1) identifying the optimal broad layer range $lr\_broad$; 2) narrowing down from $lr\_broad$ to obtain the optimal narrow layer range $lr\_narrow$; 3) exploring the best modules within the narrow layer range $lr\_narrow$.
The best solution is up to the best Uni. score through three steps. 
Note that in steps 1 and 2, all modules, including both MHA and FFN are trainable in the specific exploring layer range; in step 3, only the selective module is trainable.\footnote{We exclusively train the up\_proj $\mathbf{W}^1$ and down\_proj $\mathbf{W}^2$ within the FFN (shown in Eq.~\ref{eq:FFN}) and head's query $\mathbf{W}^Q_i$, key $\mathbf{W}^K_i$, value $\mathbf{W}^V_i$ in the MHA (shown in Eq.~\ref{eq:attn}).}

Take the exploration process for Finance task in the 13B model as an example (illustrated in Fig.~\ref{fig:search})\footnote{The details of tasks and models are illustrated in Sec.~\ref{sec:data_exp_setting}.}. 
% Here, $N$ represents the number of layers, and for the 13B model, $N=40$.
% We denote the layer number as $N$, where $N=40$ in the 13B model.
In step 1, we first obtain the Uni. scores of three broad layer ranges: $(0, 20]$, $(20, 40]$, and $(10, 30]$.
For a more accurate search, we further obtain the Uni. score of the layer range $(5, 25]$ between $(10, 30]$ with best Uni. score and $(0, 20]$ with second-best Uni. score.
% between the layer ranges achieving the best $(10, 30]$ and second-best $(0, 20]$ Uni. score.
In step 2, we first split the optimal broad layer range $lr\_broad$ $(10, 30]$ into three narrow layer ranges: $(10, 20]$, $(20, 30]$, $(15, 25]$, and follow the exploration procedures from step 1 to identify the optimal narrow layer range $lr\_narrow$ $(10, 20]$.

In step 3, we separately explore the Uni. scores for the modules within the layer range $(10, 20]$, i.e., ``$(10, 20] - \text{MHA}$'' and ``$(10, 20] - \text{FFN}$'' and obtain the best module ``$(10, 20] - \text{FFN}$''.
To delve deeper, we further examine the Uni. scores of sub-modules (up\_proj and down\_proj) within the best module (FFN). 
% For a finer search, we further attempt the Uni. scores of sub-modules (up\_proj and down\_proj) of the better module (FFN). 
Here, we opt not to explore smaller layer ranges within $lr\_narrow$ (e.g., $(10, 15]$ and $(15, 20]$) due to a substantial speciality gap compared to the all-layer range.
See Appendix~\ref{sec:narrower_range_result} for the results of smaller layer ranges.
% Results for narrower layer ranges are shown in Appendix~\ref{sec:narrower_range_result}.
% Here, we do not further attempt narrower layer ranges (e.g., $(10, 15]$ and $(15, 20]$), because the speciality in narrower layer range has a big gap between the all layer range. 

We select the best Uni. score among three steps as the final solution of the coarse-grained level.
The entire exploration is illustrated in Algorithm~\ref{algo:exploration_algorithm}.

\subsection{Fine-grained Level}
\label{sec:fine-grained_level}
Recognizing that not all units (attention heads or neurons) in a module contribute equally to the LLM versatility~\citep{michel2019sixteen}, 1) we first compute their importance for LLM versatility; 2) Subsequently, we utilize a fine-grained soft-masking (Fine-SoftMask) mechanism to control the backward gradient flow based on their importance values, aiming to further mitigate the CF issue.

% In this section, we use ``module'' or $m$ to denote the best module within the optimal layer range identified at the coarse-grained level.
In this section, we use ``module'' or $m$ to denote the module from the final solution identified at the coarse-grained level.
% MHA or FFN modules, as well as their sub-modules within the layer range identified at the coarse-grained level.
Note that the Fine-SoftMask is exclusively applied to these modules, e.g., the FFN modules within the layer range (10, 20].
% In this section, we use ``module'' or $m$ to denote MHA or FFN modules, as well as their sub-modules within the layer range identified at the coarse-grained level.
% Note that the Fine-SoftMask is exclusively applied to these modules, e.g., the FFN modules within the layer range (10, 20].
% \footnote{In this section, we use ``module'' or $m$ to denote MHA and FFN modules, as well as their sub-modules within the layer range identified at the coarse level.}.
% In this section, we use ``module'' or $m$ to indicate any of MHA and FFN modules and their sub-modules within the specific layer that identified at the coarse level, because the procedure of applying Fine-SoftMask on these modules' is similar.
% Fine-SoftMask includes two steps: 1) Computing the importance of units in the module $m$, i.e., $\mathbf{I}_m$; 2) Based on the $\mathbf{I}_m$ to control the backward gradient flow.

% \noindent \textbf{Computing Importance Vector of Module:}
\paragraph{Computing Importance of Units}
Before directly fine-tuning the module, we follow \citet{ke2022continual} to use a proxy based on robustness, i.e., KL-divergence loss, to compute units' importance for versatility without accessing the external data. This is achieved by employing a gradient-based importance detection method~\citep{michel2019sixteen}:

\vspace{-0.6cm}
\begin{align}
& \hat{\mathbf{o}}_m = \mathbf{g}_m \odot \mathbf{o}_m \nonumber \\
& \mathbf{I}_m = \frac{1}{K} \sum_{k=1}^K \left| \frac{\partial \mathcal{L}_{\text{Impt}}}{\partial \mathbf{g}_m} \right| \label{eq:imp_vector} \\
& \mathcal{L}_{\text{Impt}} = \mbox{KL}(f^1_{\text{LLM}}(\mathbf{x}), f^2_{\text{LLM}}(\mathbf{x})) \nonumber
\end{align}
% \vspace{-0.5cm}

\noindent where $\mathbf{o}_m$ refers to the output of module $m$. 
$K$ is the data size of $\mathcal{D}_{\textrm{SFT}}$.
$\mathbf{g}_m$, an all 1's \emph{virtual parameter}, remains unchanged during the computing process, as we only need its gradient $\nabla \mathbf{g}_m$ (the term within $\left| \right|$ in Eq.\ref{eq:imp_vector}) on each parameter to get the importance of corresponding unit.
% that helps to get the gradient of the unit in $\mathbf{o}_m$.
% Parameter $g_{m,i}$ in $\mathbf{g}_m$ with higher gradient value, its corresponding unit $o_{m,i}$ is considered more important.
% $\odot$ denotes element-wise multiplication, i.e., each $g_{m,i}$ in $\mathbf{g}_m$ corresponding to a unit $o_{m,i}$ in the module.
$\odot$ denotes element-wise multiplication, i.e., each $g_{m,i}$ in $\mathbf{g}_m$ corresponding to a unit in the module.
The overall importance of units $\mathbf{I}_m$ is computed by averaging the gradient values of $\mathbf{g}_{m}$, where $\mathbf{I}_m$ is of the same size as $\mathbf{g}_m$.
% The resulting $\mathbf{I}_m$ is of the same size as $\mathbf{g}_m$.
% Parameter $g_{m,i}$ in $\mathbf{g}_m$ with higher gradient value, their corresponding unit $o_{m,i}$ are considered more important, and the overall importance of the unit $o_{m,i}$ can be obtained by computing the average gradient of $g_{m,i}$.
% The resulting $\mathbf{I}_m$ is of the same size as $\mathbf{g}_m$.

$f^1_{\text{LLM}}$ and $f^2_{\text{LLM}}$ are the LLM with different dropout masks\footnote{As the Llama layer lacks dropout module, we manually add it after the output of down\_proj and the attention weights of MHA when computing the importance of units.}.
By simply feeding the same $\mathbf{x} \in \mathcal{D}_{\textrm{SFT}}$ to the model with different dropout masks twice, we obtain two representations. 
The difference between these two representations, computed by the  KL-divergence loss, can measure the unit's importance for the model.
% The difference, computed by the KL-divergence loss, between these two representations, can measure the unit's importance for the model.
% The rationale is that if a unit's change in different dropout masks significantly alters the LLM's output representation, it is considered an important unit in the module.
The rationale is as follows: if an unimportant neuron is dropped out, the resulting representation will show minimal alteration. Therefore, we will obtain a "normal type" representation similar to the one without any dropout. Conversely, if an important neuron is dropped out, the resulting representation will significantly change, resulting in an "abnormal type" representation (since dropout simulates noise addition). The difference between abnormal and normal output representations will be considerable. By utilizing KL divergence loss, we can effectively capture this difference and assign high gradient values to determine the importance of neurons. In contrast, the difference between normal representations will be marginal, resulting in small gradient values and, consequently, small importance values.

\paragraph{Fine-SoftMask Mechanism}
\label{sec:finetune_with_impt}
Then, we regulate the backward gradient flow based on the computed $\mathbf{I}_m$.
% After obtaining the $\mathbf{I}_m$, we restrict the update intensity of units crucial for model versatility by reducing their gradient.
We initially derive the original gradient $\nabla_m$ by employing $\mathcal{L}_{\text{SFT}}$ loss defined in Eq.~\ref{eq:sft_loss}.
Subsequently, we apply the $\mathbf{I}_m$ to obtain the modified gradient $\hat{\nabla}_m$ for updating:
% \vspace{-0.7cm}
\begin{align}
    & \hat{\nabla}_m = (1 - \mathbf{I}_m) \odot \nabla_m
    \label{eq:gradient_change}
\end{align}
% \vspace{-0.5cm}

\noindent Here, we expand (by copying) the $\mathbf{I}_m$ to match the dimensions of $\nabla_m$ to apply it to all associated parameters. 
% This is \emph{soft-masking} as each element in $\mathbf{I}_m$ is a real number in [0, 1] and 
This mechanism further mitigates the CF issue by regulating the update intensity of parameters based on $\mathbf{I}_m$ for versatility.
Note that the Fine-SoftMask is only applied in the backward pass during fine-tuning and the virtual parameter $\mathbf{g}_m$ mentioned in the computing stage will be discarded in this process.
% and the virtual parameter $\mathbf{g}_m$ mentioned in the computing stage will be discarded.
% Note that the virtual parameter $\mathbf{g}_m$ mentioned in the computing stage will be discarded under the this procedure.
The overview of our \emph{CoFiTune} is shown in Fig.~\ref{fig:overview}.

% to further alleviate CF in LLM, we employ a importance-based fine-tuning technique to update model's param in a fine-graind manner.

\section{Experiment}
% (3) \textbf{RQ3}: Is it effective to utilize the fine-grained soft-masking technique during fine-tuning?
In this section, we conduct extensive experiments to answer the following questions:
\textbf{1) RQ1}: Training which layers and modules can optimize the trade-off between speciality and versatility?
\textbf{2) RQ2}: How does our \emph{CoFiTune} framework compared to the baseline methods?
\textbf{3) RQ3}: Can the Fine-SoftMask mechanism further alleviate the CF issue in versatility?
\textbf{4) RQ4}: Can our \emph{CoFiTune} be seamlessly applied to models across different model scales and families?
% \textbf{4) RQ4}: Is our \emph{CoFiTune} readily applicable to models of different scales and families?
% \textbf{4) RQ4}: Is our \emph{CoFiTune} readily applicable for the larger-scale model?
% Is our \emph{CoFiTune} capable of generalizing to larger size model?

We also conduct further experimental analysis to delve into more insights for this field in Sec.~\ref{sec:further_analysis}.
% We also conduct further analysis on the results of \textbf{RQ1} to give more findings for this field in Sec.~\ref{sec:further_analysis}. 
\subsection{Datasets and Experiment Settings}
\label{sec:data_exp_setting}
\paragraph{Datasets:} 
Building upon our task formulation in Sec.~\ref{sec:task_formulation}, we establish a Chinese CF setting, aiming to advance research in the Chinese language. 
This fills a crucial gap in existing studies, which have predominantly focused on English~\citep{razdaibiedina2023progressive,lin2023speciality,luo2023empirical}.

Our setting emphasizes both speciality and versatility. Specifically, for LLM's speciality, we select tasks with considerable complexity in two primary categories:
% These tasks involve embedding specific domain knowledge into LLM, requiring the model to generate responses to questions tailored to that domain.
\textbf{1) Knowledge Intensive Tasks}: These tasks require the model to generate responses to questions tailored for specific domain. We create the \textbf{Finance} and \textbf{Law} instruction format datasets based on FiQA~\citep{wang2023fingpt} and LlamaLawyer~\citep{huang2023lawyer};
\textbf{2) Ability Intensive Tasks}: These tasks aim to test LLM's ability to handle challenging and crucial tasks in real-world scenarios.
We create the mathematical problem solving (\textbf{Math}) and Context-Aware Generation~(\textbf{CAG}\footnote{CAG task requires the model to generate answers flexibly based on the given context (See Sec.~\ref{sec:CAG_task} for more details).}) based on MGSM8k~\citep{chen2023breaking} and DuReader~\citep{he-etal-2018-dureader}.

To evaluate the versatility of LLM, we combine the insights from previous studies~\citep{lin2023speciality,luo2023empirical} and focus on three key aspects:
\textbf{1) General Domain Knowledge (Gen-Kn.)}: We adopt CMMLU~\citep{li2023cmmlu} to assess the original world knowledge stored in LLM;
\textbf{2) Generic Reasoning (Gen-Rs.)}: We utilize commonly used reasoning datasets, including LogiQA~\citep{liu2021logiqa}, LogiQA2~\citep{10174688}, OCNLI~\citep{ocnli}, and Zh-Winograd~\citep{muennighoff2022crosslingual}; 
\textbf{3) Instruction Following (Instruct.)}: Following~\citet{lin2023speciality}, we employ a Chinese GPT4-instruct dataset~\citep{peng2023gpt4llm}.
Both Gen-Kn. and Gen-Rs. are evaluated in a zero-shot manner.
Statistics and detailed descriptions of each dataset are provided in Appendix~\ref{sec:static_description_dataset}.

\paragraph{Experimental Setting:}
We use the aligned LLM Chinese-Alpaca-Pro~\citep{cui2023efficient} due to its robust versatility across multiple model scales, ranging from 7B to 33B, as detailed in Table~\ref{tab:model_sizes}.
Implementation details, including learning rates and batch sizes tailored to different model scales and tasks, can be found in Appendix~\ref{sec:implementation_details}. 
Fine-tuning adheres to the instruction prompt provided by~\citet{cui2023efficient}, detailed in Appendix~\ref{sec:instruction_prompt}.
% We fine-tuned the chat version of Chinese-Llama~\citep{cui2023efficient} as the chat version has greater versatility that we want to evaluate aforementioned than the non-chat version, which can facilitate us to better capture the CF phenomenon in versatility.
% The Chinese-Llama has several sizes ranging from 7B to 33B, details of different model sizes can refer to Table~\ref{tab:model_sizes}.
% We mainly conduct our exploration mentioned in Section~\ref{sec:coarse_level} in 7B and 13B.
% We set different learning rates, global batch sizes for different model sizes under different tasks, the implementation details are illustrated in Appendix~\ref{sec:implementation_details}.
% We follow the instruction prompt of the chat version of Chinese Llama to fine-tune, the instruction prompt is illustrated in Appendix~\ref{sec:instruction_prompt}.

\begin{table*}[!th]
    \centering
    % \normalsize
    \fontsize{10}{12}\selectfont 
    \begin{tabular}{lcccclcccc}
    \toprule[1.5pt]
    \multicolumn{1}{l}{}                                         & \multicolumn{4}{c}{\textbf{7B}}                                                &  & \multicolumn{4}{c}{\textbf{13B}}                                               
     \\ \cline{2-5} \cline{7-10} \addlinespace[2pt]
    \multicolumn{1}{l}{}                                            & \textbf{Finance}         & \textbf{Law}             & \textbf{Math}            & \textbf{CAG}             &  & \textbf{Finance}         & \textbf{Law}             & \textbf{Math}            & \textbf{CAG}             \\ \hline \addlinespace[2pt]
    \multicolumn{1}{l}{\textit{Full SFT}}   & 0.8488          & 0.8460          & 0.5184          & 0.9315          &  & 0.8770          & 0.8901          & 0.5912          & 0.9999          \\ 
     \multicolumn{1}{l}{\textit{LoRA} (Arch.)}       & 0.8534          & 0.8744          & 0.5072          & 0.9656          &  & 0.8983          & 0.9226          & 0.5628          & 1.0215          \\ 
     \multicolumn{1}{l}{\textit{Wise-FT} (Weight.)}    & 0.8802          & 0.8898          & 0.5369          & 0.9745          &  & 0.9161          & 0.9313          & 0.6070          & 1.0310          \\ 
      \multicolumn{1}{l}{\textit{V-SoftMask} (Weight.)} & 0.8537          & 0.8515          & 0.5186          & 0.9362          &  & 0.8807          & 0.8955          & 0.5954          & 1.0057          \\ 
   
     \multicolumn{1}{l}{\textit{L1} (Regular.)}         & 0.8580          & 0.8714          & 0.5122          & 0.9541          &  & 0.8988          & 0.9108          & 0.5733          & 1.0084          \\ 
     \multicolumn{1}{l}{\textit{L2} (Regular.)}         & 0.8475          & 0.8616          & 0.5135          & 0.9413          &  & 0.8772          & 0.8960          & 0.5808          & 1.0010          \\ 
     \multicolumn{1}{l}{\textit{CoFiTune} (Ours)}   & \textbf{0.8901} & \textbf{0.8993} & \textbf{0.5597} & \textbf{0.9882} &  & \textbf{0.9296} & \textbf{0.9406} & \textbf{0.6250} & \textbf{1.0511} \\ \bottomrule[1.5pt]
    \end{tabular}
    \vspace{-0.1cm}
    \caption{The Uni. scores of our \emph{CoFiTune} and baseline methods in four tasks under the 7B and 13B models. Arch., Weight. and Regular. represent Architecture-, Weight-, and Regularization-based methods respectively.}
    \vspace{-0.3cm}
    \label{tab:main_results}
    \end{table*}
    \subsection{Baselines and Metrics}
    \label{sec:baseline_metrics}
    \paragraph{Baselines:} We compared our \emph{CoFiTune} with full parameter SFT (Full SFT) and five CF baselines, with detailed descriptions provided in Appendix~\ref{sec:baselines_descriptions}.
These baselines are carefully categorized into three groups:
\begin{itemize}
  \item \textbf{Regularization-based} These methods introduce additional terms into the loss function to constrain changes in model weights. Selected baselines include L1~\citep{panigrahi2023task} and L2 regularization~\citep{lin2023speciality};
  \item \textbf{Weight-based} These methods design weight coefficients for parameters to control their updates. Selected baselines include Wise-FT~\citep{Wortsman2021wiseft} and V-SoftMask ~\citep{ke2022continual};
  \item \textbf{Architecture-based} These methods fine-tune only the external module, leaving the rest of the parameters fixed. Selected baseline includes LoRA~\citep{Hu2021LoRALA}.
\end{itemize}

% We also conduct an analysis of the training costs associated with these methods in Appendix
% \noindent \textbf{1) Regularization-based:} These methods introduce additional terms into the loss function to constrain changes in model weights.
% Selected baselines include L1~\citep{panigrahi2023task} and L2 regularization~\citep{lin2023speciality};

% \noindent \textbf{2) Weight-based:} These methods design weight coefficients for parameters to control their updates.
% % These methods aim to regulate the updating of the model's parameters based on a weighted mechanism.
% Selected baselines include Wise-FT~\citep{Wortsman2021wiseft} and V-SoftMask ~\citep{ke2022continual};

% \noindent \textbf{3) Architecture based:} These methods fine-tune only the external module, leaving the rest of the parameters fixed.
% Selected baseline includes LoRA~\citep{Hu2021LoRALA}.

\paragraph{Metrics:} 
\label{sec:metrics}
The exploration algorithm in Sec.~\ref{sec:coarse_level} requires numerous experiments, demanding a fast, cost-effective, and accurate automatic evaluation strategy.
For Finance, Law, and CAG, we employ automatic generation metrics encompassing both semantic alignment and n-gram matching, including BERTScore~\citep{bert-score}, Rouge~\citep{lin2004rouge}, and BLEU~\citep{papineni2002bleu}.
Additional experiments confirm a strong correlation between the automatic scores and those generated by GPT-4, as well as human annotations.
% , thereby demonstrating the trustworthiness of the chosen automatic generation metrics.
Moreover, a new evaluation strategy is introduced to improve the reliability of the evaluation results.
The details of the correlation evaluation and the new evaluation strategy are provided in Appendix~\ref{sec:appendix_evaluation_validation}.
For Math, we employ a rule-based extraction~\citep{chen2023breaking} method to obtain accuracy.
Utilizing the lm-evaluation-harness framework~\citep{eval-harness}, we assess the accuracy of datasets in general domain knowledge (Gen-Kn.) and generic reasoning (Gen-Rs.). The instruction following (Instruct.) is evaluated using log-likelihood (LL) following the approach of \citet{lin2023speciality}.
To evaluate overall performance in both speciality (Spec.) and versatility (Vers.), we define $\textrm{\fontsize{11}{12}\selectfont Spec.} = \frac{1}{3}(\textrm{\fontsize{11}{12}\selectfont BERTScore} + \textrm{\fontsize{11}{12}\selectfont  BLEU} + \textrm{\fontsize{11}{12}\selectfont Rouge})$ in Finance, Law and CAG, while $\textrm{\fontsize{11}{12}\selectfont Spec.} = \textrm{\fontsize{11}{12}\selectfont accuracy}$ in Math.
% Vers. score is obtained as
% Uni. score is then computed as
We define $\textrm{\fontsize{11}{12}\selectfont Vers.} = \frac{1}{3}(\textrm{\fontsize{11}{12}\selectfont Gen-Kn.} + \textrm{\fontsize{11}{12}\selectfont Gen-Rs.} + \textrm{\fontsize{11}{12}\selectfont Instruct.})$ and $\textrm{\fontsize{11}{12}\selectfont Uni.} = \textrm{\fontsize{11}{12}\selectfont Spec.} + \textrm{\fontsize{10}{12}\selectfont Vers.}$, (Uni. $\in [0, 2]$).
More details of evaluation metrics are available in Appendix~\ref{sec:appendix_evaluation_metrics}.

\subsection{Results}
% In this section, to facilitate our analysis of the result of the model only training the specific module under the curtain layer range, we define the trained model in such settings as ``layer range - module'', e.g., the model only training the FFN under the layer range $(0, 20]$ denoted as ``$(0, 20]$ - FFN''.

\paragraph{Optimal Layer Range and Module (RQ1)}
\label{sec:RQ1_layer_param}
The exploration algorithm mentioned in Sec.~\ref{sec:coarse_level} is carried out on four tasks under 7B and 13B models. 
Due to space limits, we present overall results for the Finance task under the 13B model in Fig.~\ref{fig:search} as an example, with detailed results available in Appendix~\ref{sec:appendix_layer_param} for future analysis in this field.

As depicted in Fig.~\ref{fig:search}, during step 1, the optimal Uni. score is attained with the ``(10, 30] - MHA \& FFN'' configuration.
Notably, the Spec. scores in the bottom and middle layer ranges (e.g., (0, 20], (5, 25], and (10,30]) are relatively high and comparable, whereas the top layer range (e.g., (20, 40]) exhibits the lowest performance. Conversely, the Vers. score is lowest in the bottom layer range and improves as the layer range ascends. Similar findings can also be observed in the 7B model. Further analysis of this observation is provided in Sec.~\ref{sec:speculation_harm}.

In step 2, the best Uni. score is achieved with ``(10, 20] - MHA \& FFN'', where a reduction in the number of trainable layers from 20 to 10 results in an increase in Vers. score and a decrease in Spec. score. 
However, the decline in Spec. score is less pronounced than the increase in Vers. score, summing up to a higher Uni. score in overall performance. Moving to step 3, the Uni. score for ``(10, 20] - FFN'' surpasses all other configurations. Specially, the Spec. score of ``(10, 20] - FFN'' is comparable to ``(10, 20] - MHA \& FFN'' while its Vers. score is superior. Furthermore, the Spec. score of FFN significantly outperforms other modules including MHA, down\_proj, and up\_proj. The module importance for speciality will be further examined in Section~\ref{sec:param_impt_all_layer}.

% Back to the whole picture, upon scrutinizing the results across different tasks and model sizes (\textit{i.e.}, 7B and 13B), we surprisingly observe a consistent pattern:
Back to the whole picture, upon examining the results across various tasks and model scales, we surprisingly observe a consistent pattern:

\begin{insight}
    The ``$(\mathbf{N \times 25\%}, \mathbf{N \times 50\%]}$ - FFN'' configuration yields the best Uni. score on all tasks for both 7B and 13B models.
    \label{insight:1}
\end{insight}

Namely, the optimal Uni. score of the coarse-grained level search for all tasks under the 7B model (\textit{i.e.}, $N=32$) is achieved with ``(8, 16] - FFN'', while for the 13B model (\textit{i.e.}, $N=40$), it is ``(10, 20] - FFN''.
Moreover, for certain special scenarios, we offer an optional solution in Appendix~\ref{sec:optional_solution}.
% Drawing on the above findings and considering the correspondence of 7B and 13B models to the number of Llama layers $N=32$ and $N=40$.
%we tentatively conclude that the \textbf{``$(\mathbf{N \times 25\%}, \mathbf{N \times 50\%]}$ - FFN''} configuration yields the best Uni. score.

\paragraph{Performance of \emph{CoFiTune} (RQ2)}
\label{sec:main_results}
\begin{table}[!t]
% \vspace{-0.4cm}
\centering
% \normalsize
\setlength\tabcolsep{5pt}
\fontsize{10}{13}\selectfont 
\begin{tabular}{lccccc}
% \hline
\toprule[1.5pt]
{\color[HTML]{333333} }                    & \multicolumn{2}{c}{{\color[HTML]{333333} \textbf{Finance}}}            &  & \multicolumn{2}{c}{{\color[HTML]{333333} \textbf{Math}}}               \\ \cline{2-3} \cline{5-6} \addlinespace[2pt]
                                           & {\color[HTML]{333333} \textbf{Spec.}}  & {\color[HTML]{333333} \textbf{Vers.}}  &  & {\color[HTML]{333333} \textbf{Spec.}}  & {\color[HTML]{333333} \textbf{Vers.}}  \\ \hline \addlinespace[2pt]
{\color[HTML]{333333} \textit{ZeroShot}}   & {\color[HTML]{333333} 0.3766} & {\color[HTML]{333333} 0.5201} &  & {\color[HTML]{333333} 0.0760} & {\color[HTML]{333333} 0.5201} \\
{\color[HTML]{333333} \textit{Full SFT}}   & {\color[HTML]{333333} 0.4761} & {\color[HTML]{333333} 0.4009} &  & {\color[HTML]{333333} 0.1400} & {\color[HTML]{333333} 0.4512} \\
{\color[HTML]{333333} \textit{LoRA}}       & {\color[HTML]{333333} 0.4179} & {\color[HTML]{333333} 0.4804} &  & {\color[HTML]{333333} 0.0840} & {\color[HTML]{333333} 0.4788} \\
{\color[HTML]{333333} \textit{Wise-FT}}    & {\color[HTML]{333333} 0.4308} & {\color[HTML]{333333} 0.4853} &  & {\color[HTML]{333333} 0.1000} & {\color[HTML]{333333} 0.5070} \\
{\color[HTML]{333333} \textit{V-SoftMask}} & {\color[HTML]{333333} 0.4752} & {\color[HTML]{333333} 0.4055} &  & {\color[HTML]{333333} 0.1400} & {\color[HTML]{333333} 0.4554} \\
{\color[HTML]{333333} \textit{L1}}         & {\color[HTML]{333333} 0.4287} & {\color[HTML]{333333} 0.4701} &  & {\color[HTML]{333333} 0.0920} & {\color[HTML]{333333} 0.4813} \\
{\color[HTML]{333333} \textit{L2}}         & {\color[HTML]{333333} 0.4368} & {\color[HTML]{333333} 0.4404} &  & {\color[HTML]{333333} 0.1080} & {\color[HTML]{333333} 0.4728} \\
{\color[HTML]{333333} \textit{CoFiTune}}   & {\color[HTML]{333333} 0.4351} & {\color[HTML]{333333} 0.4945} &  & {\color[HTML]{333333} 0.1120} & {\color[HTML]{333333} 0.5130} \\ \bottomrule[1.5pt]
\end{tabular}
\vspace{-0.1cm}
\caption{The Spec. and Vers. scores of Finance and Math tasks under the 13B model. \emph{ZeroShot} denotes the aligned LLM without fine-tuning.}
\vspace{-0.4cm}
\label{tab:main_detail_finace_math}
\end{table}

% We conclude the Uni. scores of our \emph{CoFiTune} and the baseline methods across various tasks and model sizes in Table~\ref{tab:main_results}. For further elaboration, the Spec. and Vers. scores for the Finance and Math tasks under the 7B model are detailedly presented in Table~\ref{tab:main_detail_finace_math}.
% Importantly, \emph{CoFiTune} also proves to be more resource-efficient in the LLM setting, requiring the training of only approximately 10.5\% of all model parameters, indicating reduced demands on training resources.

We summarize the Uni. scores of our \emph{CoFiTune} and the competitive methods across different tasks and model scales in Table~\ref{tab:main_results}. For further elaboration, the Spec. and Vers. scores for the Finance and Math tasks under the 13B model are presented in Table~\ref{tab:main_detail_finace_math}.

In Table~\ref{tab:main_results}, \emph{CoFiTune} consistently outperforms all baseline methods. Specifically, in the Finance task of 7B model, it exhibits improvements in Uni. scores of 3.7\%, 4.3\%, and 4.5\% compared to L1, LoRA, and V-SoftMask respectively. Table~\ref{tab:main_detail_finace_math} highlights that \emph{CoFiTune} especially succeeds in the balance of speciality and versatility (e.g., in Finance task, it reaches up to 98.1\% of ZeroShot in Vers. score and up to 91.4\% of Full SFT in Spec. score). 
In particular, Full SFT, V-SoftMask, and L2 exhibit a low Vers. score despite their relatively strong grasp of speciality. On the other hand, LoRA, L1, and Wise-FT fall short in terms of their performance in Spec. score. Similar trends are observed in the remaining results, detailed in Appendix~\ref{sec:appendix_main_results}.
%In particular, Full SFT, V-SoftMask, L1 and L2 suffer from a low Vers. score despite their relatively good acquisition of speciality, while LoRA and Wise-FT lag behind in their speciality performance.

%Wise-FT emerges as the strongest baseline method, boasting the highest Vers. score and competitive Spec. score among all baselines. L2 demonstrates parity with \emph{CoFiTune} in Spec. score but lags in Vers. score. LoRA exhibits the lowest Spec. score, suggesting challenges in adapting to demanding tasks. Similar trends are observed in the remaining results, detailed in Appendix~\ref{sec:appendix_main_results}.

\paragraph{Impact of Fine-SoftMask (RQ3)}
\label{sec:abl_softmask}
% Please add the following required packages to your document preamble:
% \usepackage[table,xcdraw]{xcolor}
% Beamer presentation requires \usepackage{colortbl} instead of \usepackage[table,xcdraw]{xcolor}

% \centering
% \normalsize
% \setlength\tabcolsep{4pt}
% Please add the following required packages to your document preamble:
% \usepackage[table,xcdraw]{xcolor}
% Beamer presentation requires \usepackage{colortbl} instead of \usepackage[table,xcdraw]{xcolor}
\begin{table}[!t]
% \vspace{-0.4cm}
\centering
% \normalsize
\setlength\tabcolsep{5pt}
\fontsize{10}{13}\selectfont 
\begin{tabular}{lccccc}
\toprule[1.5pt]
                                        & \multicolumn{2}{c}{{\color[HTML]{333333} \textbf{\textit{CoFiTune}}}} &  & \multicolumn{2}{c}{{\color[HTML]{333333} \textbf{\textit{\begin{tabular}[c]{@{}c@{}}CoFiTune \\ w/o Fine-SoftMask\end{tabular}}}}} \\ \cline{2-3} \cline{5-6} \addlinespace[2pt]
                                        & {\color[HTML]{333333} \textbf{Spec.}}  & {\color[HTML]{333333} \textbf{Vers.}}  &  & {\color[HTML]{333333} \textbf{Spec.}}                                 & {\color[HTML]{333333} \textbf{Vers.}}                                \\ \hline \addlinespace[2pt]
{\color[HTML]{333333} Finance} & 0.4351                        & 0.4945                       &  & 0.4352                                                       & 0.4912                                                     \\
{\color[HTML]{333333} Law}     & 0.4503                        & 0.4903                       &  & 0.4512                                                       & 0.4855                                                     \\
{\color[HTML]{333333} Math}    & 0.1120                         & 0.5130                        &  & 0.1080                                                        & 0.5102                                                       \\
{\color[HTML]{333333} CAG}     & 0.5406                        & 0.5105                       &  & 0.5397                                                       & 0.5054                                                     \\ \bottomrule[1.5pt]
\end{tabular}
% \vspace{-0.1cm}
\caption{The impact of Fine-SoftMask in \emph{CoFiTune} on Spec. and Vers. scores under the 13B model. ``w/o'' means excluding this technique from \emph{CoFiTune}.}
\vspace{-0.4cm}
\label{tab:abl_softmask}
\end{table}

We conduct additional experiments to demonstrate the effectiveness of the Fine-SoftMask mechanism discussed in Sec.~\ref{sec:fine-grained_level}, and give the conclusion below:

\begin{insight} 
    Fine-SoftMask mechanism effectively mitigates the CF in LLM's versatility without harming the speciality performance.
    \label{insight:2}
\end{insight}
%The results, presented in Table~\ref{tab:abl_softmask} for the 13B setting and in Appendix~\ref{sec:appendix_softmask} for the 7B setting, highlight the impact of the Fine-SoftMask mechanism.
Concretely, we report the results under 13B in Table~\ref{tab:abl_softmask}.
%and the results under 7b in Appendix~\ref{sec:appendix_softmask}.
%As depicted in Table~\ref{tab:abl_softmask}, %the Fine-SoftMask mechanism proves effective in mitigating CF in the versatility of LLMs, without compromising speciality performance. 
% When applying Fine-SoftMask, we observe nearly identical Spec. scores compared to not applying it in Finance and Law
% There's even a slight improvement in CAG and Math.
When applying Fine-SoftMask, we observe nearly identical Spec. scores in the Finance and Law tasks compared to not applying it, and even a slight improvement in CAG and Math tasks.
Moreover, Fine-SoftMask contributes to a Vers. score improvement of 0.7\%, 1\%, and 1.1\% in Finance, Law and CAG tasks respectively. 
Similar conclusions in the 7B model are presented in Table~\ref{tab:abl_softmask_7b}.
%Similar trends and conclusions are evident across various tasks, as detailed in other experiments presented in Appendix~\ref{sec:appendix_softmask}.

% 这边就放 Math 13b 和 Finance 13b 的结果
% We further conduct experiments to show the effectiveness of the Fine-SoftMask mechanism mentioned in Section~\ref{sec:fine-grained_level}.
% We report the results under 13B in Table~\ref{tab:abl_softmask} and the results under 7b in Appendix~\ref{sec:appendix_softmask}.
% As shown in Table~\ref{tab:abl_softmask}, we can observe that the Fine-SoftMask mechanism can further prevent the CF in the generality of LLM while not harming the performance of the specialty.
% % For example, in Math under 13B model, applying the Fine-SoftMask mechanism gets the same Spec. score as the not applying one, while the Vers. score improves by 0.8\%.
% % In CAG under the 13b model, it improves the Vers. score and even can slightly improve the Spec. score.
% For example, applying the Fine-SoftMask mechanism gets almost the same Spec. score as the not applying one in Finance, Law, and Math, even slightly improves in CAG and Math.
% Meanwhile, it brings the improvement of Vers. by 1.2\%, 1.1\%, and 0.8\% in Law, CAG, and Math.
% % while the Vers. score improves by 0.8\%.
% % In CAG under the 13b model, it improves the Vers. score and even can slightly improve the Spec. score.
% Similar conclusions can be found in other tasks in Appendix~\ref{sec:appendix_softmask}.
% % even can slightly improve it.
% % This results aligns with ~\citet{}

% \paragraph{Scaling \emph{CoFiTune}: Performance across Model Scales and Families (RQ4)}
\paragraph{Performance of \emph{CoFiTune} across Models of Different Scales and Families (RQ4)}
\label{sec:large_results}
\begin{table}[!t]
% \vspace{-0.4cm}
\centering
% \normalsize
\setlength\tabcolsep{5pt}
\fontsize{10}{13}\selectfont 

\begin{tabular}{lcccc}
\toprule[1.5pt]
                                           & {\color[HTML]{333333} \textbf{Finance}} & {\color[HTML]{333333} \textbf{Law}}    & {\color[HTML]{333333} \textbf{Math}}  & {\color[HTML]{333333} \textbf{CAG}}   \\\addlinespace[2pt] \hline \addlinespace[2pt]
{\color[HTML]{333333} \textit{Full SFT}}   & {\color[HTML]{333333} 0.9294}  & {\color[HTML]{333333} 0.9392} & {\color[HTML]{333333} 0.6910} & {\color[HTML]{333333} 1.0527} \\
{\color[HTML]{333333} \textit{LoRA}}       & 0.9368                         & 0.9635                        & 0.6427                       & 1.0648                        \\
{\color[HTML]{333333} \textit{Wise-FT}}    & 0.9595                         & 0.9811                        & 0.6876                       & 1.0776                        \\
{\color[HTML]{333333} \textit{V-SoftMask}} & 0.9343                         & 0.9453                        & 0.6911                       & 1.0579                        \\
{\color[HTML]{333333} \textit{L1}}         & 0.9301                         & 0.9536                        & 0.6640                        & 1.0527                        \\
{\color[HTML]{333333} \textit{L2}}         & 0.9170                         & 0.9388                        & 0.6698                       & 1.0556                        \\
{\color[HTML]{333333} \textit{CoFiTune}}   & \textbf{0.9722}                & \textbf{0.9910}                & \textbf{0.6991}              & \textbf{1.0904}               \\ \bottomrule[1.5pt]
\end{tabular}
% \vspace{-0.1cm}
\caption{The Uni. score of our \emph{CoFiTune} and baseline methods under the 33B model.}
% \vspace{-0.1cm}
\label{tab:larger_model}
\end{table}

\begin{table}[!t]
% \vspace{-0.4cm}
\centering
% \normalsize
\setlength\tabcolsep{4.2pt}
\fontsize{10}{12}\selectfont 
\begin{tabular}{lccccc}
\toprule[1.5pt]
                                           & \multicolumn{2}{c}{{\color[HTML]{333333} \textbf{3B}}}                                   &           & \multicolumn{2}{c}{{\color[HTML]{333333} \textbf{7B}}}                                   \\ \cline{2-3} \cline{5-6} \addlinespace[2pt] 
                                           & {\color[HTML]{333333} \textbf{Finance}}         & {\color[HTML]{333333} \textbf{Math}}            &           & {\color[HTML]{333333} \textbf{Finance}}         & {\color[HTML]{333333} \textbf{Math}}            \\ \hline \addlinespace[2pt] 
{\color[HTML]{333333} \textit{Full SFT}}   & {\color[HTML]{333333} 0.7437} & {\color[HTML]{333333} 0.3548} &           & {\color[HTML]{333333} 0.7926} & {\color[HTML]{333333} 0.4365} \\
{\color[HTML]{333333} \textit{LoRA}}       & {\color[HTML]{333333} 0.7514}          & {\color[HTML]{333333} 0.3462}          &           & {\color[HTML]{333333} 0.8073}          & {\color[HTML]{333333} 0.4237}          \\
{\color[HTML]{333333} \textit{Wise-FT}}    & {\color[HTML]{333333} 0.7782}          & {\color[HTML]{333333} 0.3715}          &           & {\color[HTML]{333333} 0.8351}          & {\color[HTML]{333333} 0.4519}          \\
{\color[HTML]{333333} \textit{V-SoftMask}} & {\color[HTML]{333333} 0.7463}          & {\color[HTML]{333333} 0.3564}          &           & {\color[HTML]{333333} 0.7973}          & {\color[HTML]{333333} 0.4383}          \\
{\color[HTML]{333333} \textit{L1}}         & {\color[HTML]{333333} 0.7568}          & {\color[HTML]{333333} 0.3542}          &           & {\color[HTML]{333333} 0.8144}          & {\color[HTML]{333333} 0.4331}          \\
{\color[HTML]{333333} \textit{L2}}         & {\color[HTML]{333333} 0.7505}          & {\color[HTML]{333333} 0.3547}          &           & {\color[HTML]{333333} 0.8017}          & {\color[HTML]{333333} 0.4304}          \\
{\color[HTML]{333333} \textit{CoFiTune}}   & {\color[HTML]{333333} \textbf{0.7876}} & {\color[HTML]{2C3A4A} \textbf{0.3891}} & \textbf{} & {\color[HTML]{2C3A4A} \textbf{0.8463}} & {\color[HTML]{2C3A4A} \textbf{0.4758}} \\ \bottomrule[1.5pt]
\end{tabular}
% \vspace{-0.1cm}
\caption{The Uni. scores of Finance and Math tasks under the 3B and 7B BLOOMZ models.}
\vspace{-0.5cm}
\label{tab:bloomz_results}
\end{table}

To verify the generalization of \emph{CoFiTune} on different model scales and families, we directly apply Insight~\ref{insight:1}, obtained in \textbf{RQ1}, without tree-search exploration to the Chinese-Alpaca-Pro 33B model and another popular aligned model\footnote{In this paper, unless explicitly specified, the model referred to is Chinese-Alpaca-Pro.}, BLOOMZ~\citep{muennighoff2023crosslingual}.
The overall results of the 33B model are shown in Table~\ref{tab:larger_model}, with detailed results in Appendix~\ref{sec:appendix_main_results_33b}. 
For the BLOOMZ model, we conduct experiments on 3B and 7B scales under Finance and Math tasks, the overall results are presented in Table~\ref{tab:bloomz_results}, with detailed results in Appendix~\ref{sec:appendix_main_results_bloomz}.
As indicated in Table~\ref{tab:larger_model}, \emph{CoFiTune} maintains its superior unified performance under the 33B model, exhibiting improvements in Uni. score by 1.2\%, 1.3\%, and 1.6\% in Finance, CAG, and Math tasks compared to the best-performing baseline Wise-FT.
Similarly, Table~\ref{tab:bloomz_results} demonstrates the strong competence of our \emph{CoFiTune} under the BLOOM family model, showcasing improvements in Uni. score by 5.9\% and 6.4\% in the Finance task when compared to Full SFT baseline.
\emph{These results further validate the widespread effectiveness of Insight~\ref{insight:1} and CoFiTune across models of different scales and families}.\footnote{Due to the computing resource limitation, we defer the experimental analysis on even larger LLMs for future work.}

% Due to computing resource limitations, we defer the experimental analyses on even larger LLMs for future work . Nevertheless, we observe alignment between our findings and those from \citet{bansal2022rethinking}, where only the performance of 66B model under ICL settings is explored. 
% To further demonstrate the generalization of \emph{CoFiTune}, we directly utilize Insight~\ref{insight:1} obtained in \textbf{RQ1} to the 33B model. The overall results are listed in Table~\ref{tab:larger_model} and detailed results are in Appendix~\ref{sec:appendix_main_results}.
% As shown in Table~\ref{tab:larger_model}, \emph{CoFiTune} still achieves the best unified performance under 33B model, with an improvement of Uni. score by 1.3\%, 1.29\%, and 1.18\% in Finance, Law, and CAG compared to the best-performing baseline Wise-FT. 
% Such results further support the generalization of Insight~\ref{insight:1} as well as the effectiveness of \emph{CoFiTune}. 

%However, we observed alignment between our findings and those from a related study conducted on the 66B model under ICL settings~\citep{bansal2022rethinking}. 

% Due to the computing resource limitation, we haven't conducted on the 65B model, but we find that the observation from the work that conducted only on 66B model under ICL setting is aligned with our conclusion~\citep{bansal2022rethinking}. We will leave this in the future work.

\subsection{Further Analysis}
\label{sec:further_analysis}
\begin{figure}[!thbp]
    \centerline{\includegraphics[height=5cm,width=\columnwidth]{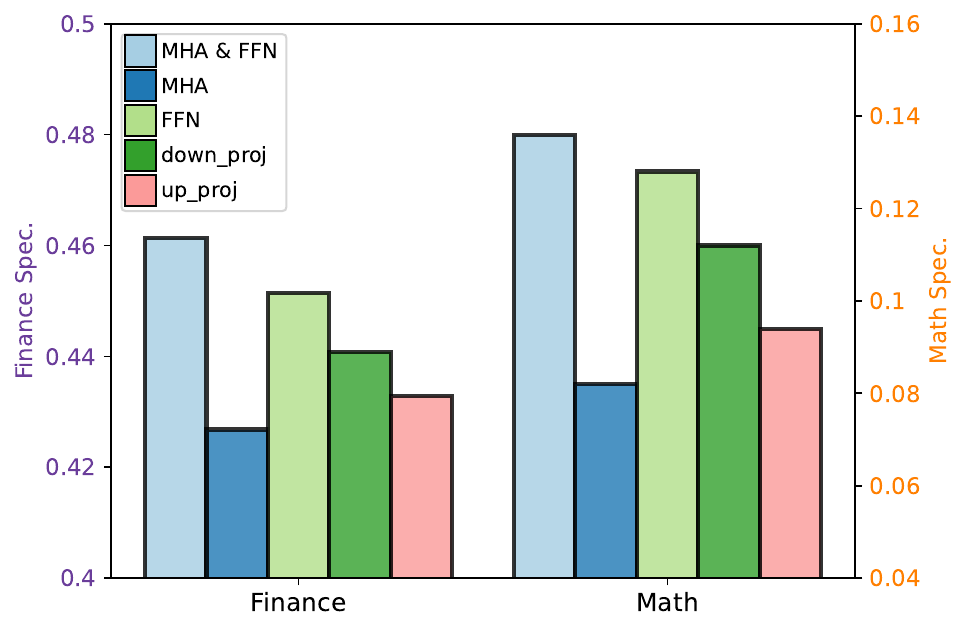}}
    % \vspace{-0.2cm}
    \caption{Spec. scores for Finance and Math tasks under the 13B model across modules trained in all layers.}
    \label{fig:all_layer_impt}
    \vspace{-0.2cm}
\end{figure}

\begin{figure*}[!t]
    \centering
    \includegraphics[width=1.45\columnwidth]{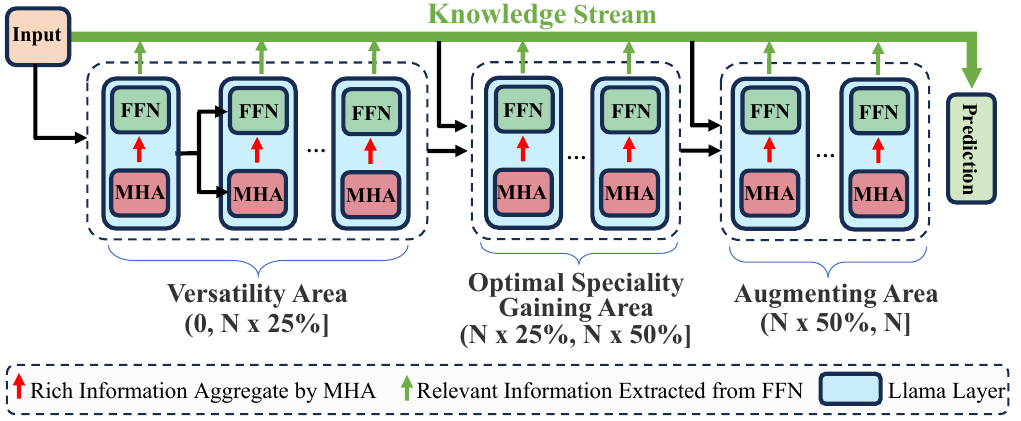}
    \vspace{-0.2cm}
    \caption{An illustration of our speculation on the process of information forwarding.}
    \label{fig:knowledge_flow}
    \vspace{-0.4cm}
\end{figure*}

\begin{figure}[!t]    \centerline{\includegraphics[height=5cm,width=\columnwidth]{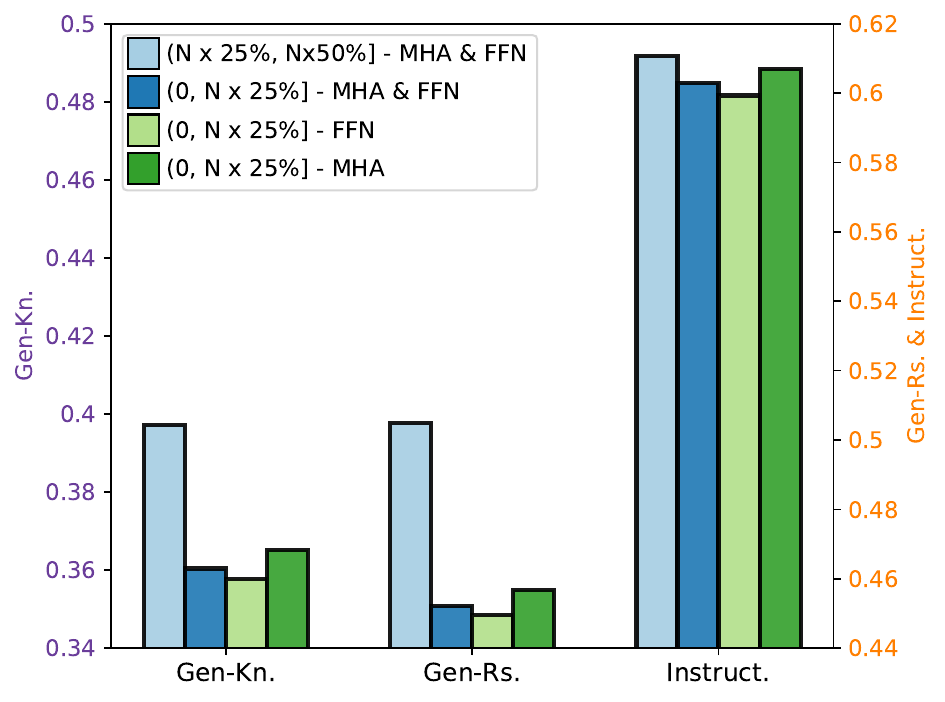}}
    \vspace{-0.2cm}
    \caption{The Gen-Kn., Gen-Rs., and Instruct. scores in Math task under the 13B model.}
    \label{fig:bottom_layer_general_math_13b}
    \vspace{-0.4cm}
\end{figure}

\paragraph{Module Importance for Speciality}
\label{sec:param_impt_all_layer}
To delve deeper into module significance for gaining speciality, we separately experimented with MHA, FFN, down\_proj, and up\_proj modules across all $N$ layers in both the 7B and 13B models.
Results for the Finance and Math tasks under the 13B model are shown in Fig.~\ref{fig:all_layer_impt}.
% As shown in Fig.~\ref{fig:all_layer_impt}, FFN achieves the highest Spec. score, followed by down\_proj and up\_proj, while MHA performing the least favorably.
% Similar trends are observed under the 7B model in Fig.~\ref{fig:all_layer_impt_7b}.
As shown in Fig.~\ref{fig:all_layer_impt}, FFN achieves a Spec. score most comparable to MHA \& FFN, followed by down\_proj and up\_proj, while MHA performs the least favorably.
Similar trends are observed under the 7B model in Fig.~\ref{fig:all_layer_impt_7b}.
Based on these observations, we draw the following insight:

\begin{insight}
    FFN, especially the down\_proj in it, is more crucial than MHA when gaining speciality.
    \label{insight:3}
\end{insight}

This\footnote{Notably, the number of trainable parameters in down\_proj and MHA are of the same scale.} aligns, to some extent, with views from \citet{geva2021transformer} and \citet{dai2022knowledge}, suggesting FFN operates as key-value memories. 
The prediction distributions over the output vocabulary are determined by the down\_proj, while MHA captures more superficial linguistic and pattern information~\citep{manning2020emergent,rogers2021primer}.
\paragraph{Exploring CF in LLM's Versatility}
\label{sec:speculation_harm}
% We further explore the observation from Sec.~\ref{sec:RQ1_layer_param} which unveils a notable gap in Vers. scores between the bottom $(0, N\times50\%]$ and middle $(N\times25\%, N\times75\%]$ layer ranges in both 7B and 13B models, while their Spec. scores remain comparable.
We further explore the observation from Sec.~\ref{sec:RQ1_layer_param} which unveils a notable gap in Vers. scores between the bottom $(0, N\times50\%]$ and middle $(N\times25\%, N\times75\%]$ layer ranges in both 7B and 13B models.
% mid-to-lower $(N\times25\%, N\times50\%]$ layer ranges in both 7B and 13B models.
Note that there are two non-overlapping layer ranges, $(0, N \times 25\%]$ and $(N \times 50\%, N \times 75\%]$, situated between $(0, N \times 50\%]$ and $(N \times 25\%, N \times 75\%]$.
However, the Vers. score of $(N \times 50\%, N \times 75\%]$ is favorable (shown in Fig.~\ref{fig:search} and Appendix~\ref{sec:appendix_layer_param}); therefore, we suspect that it is the $(0, N \times 25\%]$ that harms LLM's versatility. 
% Given the presence of the non-overlapping layer ranges $(0, N \times 25\%]$ and $(N \times 50\%, N \times 75\%]$ between $(0, N \times 50\%]$ and $(N \times 25\%, N \times 75\%]$, and the Vers. score of layer range $(N \times 50\%, N \times 75\%]$ is favorable, therefore, 
% we suspect that it is the $(0, N \times 25\%]$ layer range that harms LLM's versatility. 

Thus, we conduct experiments in this range with detailed Vers. score of Math task under 13B model in Fig.~\ref{fig:bottom_layer_general_math_13b}.
In Fig.~\ref{fig:bottom_layer_general_math_13b}, we observe that:
1) The detailed Vers. scores of ``$(0, N \times 25\%]$ - MHA \& FFN'' significantly lag behind ``$(N \times 25\%, N \times 50\%]$ - MHA \& FFN'';
2) In the layer range $(0, N \times 25\%]$, compared with ``FFN \& MHA'', FFN further impairs versatility while MHA mitigates it.
Similar findings across various tasks and model scales are shown in Appendix~\ref{sec:appendix_speculate_harm}.
Hence, we derive the following insight:

\begin{insight}
    The LLM's versatility may predominantly reside in the layer range $(0, N\times25\%]$, particularly within the FFN module.
    \label{insight:4}
\end{insight}

% \footnote{Their studies examine the functions of MHA and FFN in the prediction process.}
Furthermore, combining the hints from \citet{geva2021transformer}, \citet{meng2022mass}, \citet{allenzhu2023physicspart3.2}, \citet{allenzhu2023physicspart3.1}, and \citet{chuang2023dola} with our findings in this work, we speculate the forward process in Fig.~\ref{fig:knowledge_flow}:
1) MHA modules in \emph{Versatility Area} aggregate information, enriching input for corresponding FFN. FFN modules utilize this enriched input information to extract relevant knowledge, passing it through residual connections to \emph{Optimal Speciality Gaining Area};
2) \emph{Optimal Speciality Gaining Area} handles both versatility and speciality information, forwarding it to \emph{Augmenting Area}.
3) Processed information from the previous two areas is forwarded to \emph{Augmenting Area} through residual connections, focusing on  information enhancement for final prediction.
% Processed information from the bottom and middle layer ranges is forwarded to the top through residual connections, focusing on enhancing information for the final prediction, i.e., the \emph{Augmenting Area}.
% \footnote{This speculation stems from studies highlighting the top layers' emphasis on augmentation~\citep{geva2021transformer, meng2022mass, chuang2023dola}.}.

In this context, modifying modules in \emph{Versatility Area} impairs the original memory and abilities of LLM, impeding the flow of versatility information passing through. 
In contrast, updating modules in \emph{Optimal Speciality Gaining Area} allows versatility information to be recalled and passed through residual connections. 
Extra experiments are scheduled to validate these speculations in the future.
% Further experiments are planned to validate these speculations.
% In this context, modifying modules in the \emph{Versatility Area} impairs the original memory and abilities of LLM, impeding the flow of versatility information. 
% In contrast, updating modules in the \emph{Optimal Speciality Gaining Area} allows versatility information to be recalled and passed through residual connections. 
% Further experiments are planned to validate these speculations.

% \begin{figure}[!t]
%     \centering
%     \includegraphics[width=\columnwidth]{latex/figs/knowledge_flow.pdf}
%     % \vspace{-0.4cm}
%     \caption{An illustration of our speculation on the process of information forwarding.}
%     \label{fig:knowledge_flow}
%     % \vspace{-0.6cm}
% \end{figure}

\section{Conclusion}
\label{sec:conclusion}
% \begin{figure}[!thbp]
%     \centerline{\includegraphics[height=5cm,width=\columnwidth]{latex/figs/bottom_layer_general.pdf}}
%     \vspace{-0.2cm}
%     \caption{The Spec. score of different modules trained in all layers for Finance and Math under the 13B model.}
%     \label{fig:all_layer_impt}
%     \vspace{-0.4cm}
% \end{figure}
In this work, we strive to optimize the balance between LLM's speciality and versatility. 
Our proposed \emph{CoFiTune} framework employs an empirical tree-search exploration algorithm to identify the module within a defined layer range that is crucial for gaining speciality without significantly affecting versatility. 
Additionally, the Fine-SoftMask mechanism is applied to further alleviate CF issue without impairing the speciality. 
We introduce a Chinese CF setting to advance the research in Chinese domain. 
The experimental results demonstrate that our \emph{CoFiTune} outperforms all baselines across various tasks and model scales. Our in-depth analysis provides valuable insights for future work.

\section{Limitations}
\label{sec:limitation}
In this study, we opt not to use the rehearsal-based method, which involves replaying a small portion of the general dataset during fine-tuning~\citep{rolnick2019experience,de2019episodic}. 
This decision stems from the uncertainty surrounding the replay data ratio and strategy in this domain. 
Increasing the number of replay samples during fine-tuning would also incur higher training resource costs. 
Moreover, we believe that the rehearsal-based method can be integrated with our proposed \emph{CoFiTune} framework, and we plan to explore this in future work.

Additionally, the model employed in this study is mainly based on the Llama architecture. This choice is driven by the fact that Llama family models demonstrate remarkable dominance and performance in the current NLP research area (constituting over 90\% on the LLM leaderboard\footnote{\url{https://huggingface.co/spaces/HuggingFaceH4/open_llm_leaderboard}}).
% Furthermore, Llama is built upon the representative and classical transformer decoder-only architecture.
Besides, based on the fact that most current LLMs share the same architecture as Llama, i.e., transformer decoder-only architecture, we are confident that our identified universal optimal range can also be applied to a majority of models within the same architecture (the experiment results of BLOOMZ in \textbf{RQ4} have validated this point). Therefore, our \emph{CoFiTune} framework can be directly adopted in these models using \textbf{Insight~\ref{insight:1}}, without incurring any tree-search exploration overhead.
Furthermore, we plan to explore other model architectures (e.g., encoder-decoder) in future research.
\bibliography{acl_latex}

\newpage
\onecolumn
\appendix
\section{Appendix}
\label{sec:appendix}
\subsection{Detailed Information of Llama Backbone}
\label{sec:swiglu_rmsnorm}
The MHA operation is a crucial component of the transformer, defined as:
\begin{equation}
\text{MHA}(\mathbf{Q}, \mathbf{K}, \mathbf{V}) = [\text{head}_1; \dots; \text{head}_h]\mathbf{W}^O
\end{equation}

\noindent where $;$ is the concatenation operation, $\mathbf{Q}$, $\mathbf{K}$, and $\mathbf{V}$ are the query, key, and value matrices, respectively, and $\mathbf{W}^O$ is a learnable output matrix unique to the MHA module. Each head is computed as:
\begin{equation}
\begin{split}
\text{head}_i = \text{Attention}(\mathbf{x}\mathbf{W}^Q_i, \mathbf{x}\mathbf{W}^K_i, \mathbf{x}\mathbf{W}^V_i) \\
\text{Attention}(\mathbf{Q}_i, \mathbf{K}_i, \mathbf{V}_i) = \text{Softmax} \left(\frac{\mathbf{Q}_i\mathbf{K}_i^T}{\sqrt{d_k}}\right)\mathbf{V}_i
\end{split}
\end{equation}
\noindent where $\mathbf{W}^Q_i$, $\mathbf{W}^K_i$, and $\mathbf{W}^V_i$ are the corresponding weight matrices of query, key and value for the $i$-th head.

The FFN module is parameterized by an up projection matrix (up\_proj), followed by a down projection matrix (down\_proj):
\begin{equation}
\begin{split}
& \text{FFN}(\mathbf{x}') = \bigl( \text{SwiGLU}(\mathbf{x}') \otimes (\mathbf{x}'\mathbf{W}^1)\bigl) \mathbf{W}^2\\
& \text{SwiGLU}(\mathbf{x}') = \text{SiLU}(\mathbf{x}' \mathbf{W}) \\
& \text{RMSNorm}(\mathbf{x}') = \frac{w \odot \mathbf{x}'}{\sqrt{\text{Var}(\mathbf{x}') + \epsilon}}
\end{split}
\end{equation}
where $\mathbf{x}'$ is the input tensor of FFN, $\mathbf{W}$ is the learnable weight matrix without bias (a.k.a gated projection matrix), $\mathbf{W}^1$ and $\mathbf{W}^2$ are up\_proj and down\_project respectively. $\text{SiLU}(\mathbf{x}') = \mathbf{x}' \otimes \text{Sigmoid}(\mathbf{x}')$. 
$\otimes$ is the element-wise multiplication operation.
$w$ represents the learnable weight parameter, $\text{Var}(\mathbf{x}')$ denotes the variance of $\mathbf{x}'$ across the last dimension, and $\epsilon$ is a small constant introduced for numerical stability.

As the Llama layer lacks dropout module, we manually add it after the output of down\_proj and the attention weights of MHA when computing the importance of units in Sec.~\ref{sec:fine-grained_level}.

Given \citet{mirzadeh2023relu}'s demonstration that SwiGLU, a commonly used trainable activation function in the FFN of LLMs, can be substituted with non-trainable ReLU to decrease computation without compromising performance, we have chosen not to investigate SwiGLU in our exploration outlined in Sec.~\ref{sec:coarse_level}. Specifically, our exploration focuses solely on the up\_proj and down\_proj components within the FFN.
% As zhang demonstrated that SwiGLU, a common trainable activation function in the FFN of LLMs, can be replaced with non-trainable ReLU to reduce computation without affecting performance, therefore, we opt not to explore the SwiGLU in our exploration mentioned in Sec.~\ref{sec:coarse_level}, i.e., we only explore the up\_proj and down\_proj in FFN.
% $w$ is the learnable weight parameter, $\text{Var}(x)$ is the variance of $x$ across the last dimension, and $\epsilon$ is a small constant for numerical stability.
\nocite{shang2024incremental}
\nocite{shang2024understanding}
\nocite{zhang2024improving}
\nocite{zhang2024question}

% where $\otimes$ denotes element-wise multiplication, 

\subsection{Smaller Exploring Layer Range Results}
\label{sec:narrower_range_result}
\begin{table}[!h]
\centering
\setlength\tabcolsep{4.5pt}
\fontsize{10}{14}\selectfont 
\begin{tabular}{lccccc}
\toprule[1.5pt]
                                      & \multicolumn{2}{c}{7B} &  & \multicolumn{2}{c}{13B} \\ \cline{2-3} \cline{5-6} 
                                      & Finance    & Math      &  & Finance     & Math      \\ \hline
(0, N{]} MHA \& FFN                   & 0.4432     & 0.0960    &  & 0.4614      & 0.136     \\
(N x 25\%, N x 37.5\%{]} - MHA \& FFN & 0.3921     & 0.0360    &  & 0.4183      & 0.068     \\
(N x 37.5\%, N x 50\%{]} - MHA \& FFN & 0.3923     & 0.0400    &  & 0.4189      & 0.064     \\ \bottomrule[1.5pt]
\end{tabular}
\caption{The speciality performance of Finance and Math under 7B and 13B within a smaller layer range.}
\end{table}

There is a substantial speciality gap between the smaller layer range and the all-layer range.
For example, in Math under the 7B model, the smaller layer range $(N\times 25\%, N\times 37.5\%$ accounts for only 37.5\% of $(0, N]$ in Accuracy.

\subsection{Exploration Algorithm in Coarse-grained Level}
\label{sec:appendix_exploration_algo}
The exploration at coarse level is illustrated in Algorithm~\ref{algo:exploration_algorithm}.
The variables $results_1, results_2, results_3$ are dictionary data structures intended to store results from steps 1, 2, and 3, respectively. 
The variables $left, right, pivot$ represent the start index of their corresponding layer range. 
$\alpha$ denotes the number of layers targeted for coverage. For instance, in a 13B model ($N=40$), $\alpha = 20$ in step 1 and $\alpha = 10$ in step 2.
% $\alpha$ is the number of layers that attempt to cover. For example, under 13B model ($N=40$), in step 1, $\alpha = 20$, in step 2, $\alpha = 10$.
The \text{GetBestLayers} function aims to identify the layer range that performs best in terms of the unified score. The \text{GetSubModules} function returns the sub-modules of a specific module; for example, the sub-modules of FFN are up\_proj and down\_proj. The functions \text{GetMax} and \text{GetSecond} are responsible for obtaining the start index of the layer range that achieves the best and second-best performance.
% The \text{GetBestLayers} function aims to find the layer range that performs the best in terms of unified score.
% The \text{GetSubModules} function will return the sub modules of specific module, e.g., the sub modules of FFN are up\_proj and down\_proj.
% \text{GetMax} and \text{GetSecond} functions are responsible for getting the start index of the layer range that gets the best and second best performance.
$f_{*\_layer\_range}$ denotes the parameters of FFN and MHA in the $*\_layer\_range$ that are trainable in the model, while others are frozen. Similarly, $f^{m}_{*\_layer\_range}$ represents the parameters of a specific module in the $*\_layer\_range$ that are trainable in the model, while others are frozen. The \text{Max} function returns the optimal strategy that achieves the highest unified score among all three steps.
\begin{algorithm}[!h]
% \floatplacement{algorithm}{!h}
\caption{Exploration Algorithm in Coarse Level}
\label{algo:exploration_algorithm}
\begin{algorithmic}[1]
\STATE \textbf{Input:} Llama Model $f(\cdot)$, $N$ layers, evaluation data $D$
\STATE \textbf{Initialize:} $results_1, results_2, results_3 \gets \{\}$, $left \gets 0$, $right  \gets  N \times 50\%, \alpha \gets N \times 50\%$
\STATE \textbf{Function:} \\
\textbf{def} GetBestLayerRange($left$, $right$, $\alpha$, $result$):
\STATE $\hspace{0.5cm}$ $pivot \gets (left + right) 
 \hspace{0.05cm} // \hspace{0.05cm} 2$
\STATE $\hspace{0.5cm}$ $left\_layer\_range \gets (left, left + \alpha]$
\STATE $\hspace{0.5cm}$ $right\_layer\_range \gets (right, right + \alpha]$
\STATE $\hspace{0.5cm}$ $pivot\_layer\_range \gets (pivot, pivot + \alpha]$
\STATE $\hspace{0.5cm}$ $result[left] \gets$ Eval($f_{left\_layer\_range}(\cdot), D$)
\STATE $\hspace{0.5cm}$ $result[right]\gets$ Eval($f_{right\_layer\_range}(\cdot), D$)
\STATE $\hspace{0.5cm}$ $result[pivot] \gets$ Eval($f_{pivot\_layer\_range}(\cdot), D$)
\STATE $\hspace{0.5cm}$ $left \gets$ GetMax($result$)
\STATE $\hspace{0.5cm}$ $right \gets$ GetSecond($result$)
\STATE $\hspace{0.5cm}$ $pivot_2 \gets (left + right) 
 \hspace{0.05cm} // \hspace{0.05cm} 2$
\STATE $\hspace{0.5cm}$ $pivot_2\_layer\_range \gets (pivot_2, pivot_2 + \alpha]$
\STATE $\hspace{0.5cm}$ $result[pivot_2] \gets$ Eval($f_{pivot_2\_layer\_range}(\cdot), D$)
\STATE $\hspace{0.5cm}$ return GetMax($result$)

\textbf{\# Step 1:} 
% \STATE $\alpha \gets N \times 50\%$
\STATE $lr\_broad\_start\_idx \gets$ GetBestLayerRange($left$, $right$, $\alpha$, $results_1$)

\textbf{\# Step 2:} 
% \STATE $\alpha \gets N \times 25\%$
\STATE $left \gets lr\_broad\_start\_idx$; $right \gets lr\_broad\_start\_idx + \alpha$; $\alpha \gets \alpha // 2$
\STATE $lr\_narrow\_start\_idx \gets$ GetBestLayerRange($left$, $right$, $\alpha$, $results_2$)

\textbf{\# Step 3:}
\STATE $lr\_narrow \gets (lr\_narrow\_start\_idx, lr\_narrow\_start\_idx + \alpha]$
% \FOR{$m \gets [\text{MHA}, \text{FFN}, \text{down\_proj}, \text{up\_proj}]$}
% \STATE $results_3[m] \gets$ Eval($f^m_{best_2\_layer\_range}(\cdot), D$)
% \ENDFOR
\FOR{$m \gets [\text{MHA}, \text{FFN}]$}
\STATE $results_3[m] \gets$ Eval($f^m_{lr\_narrow}(\cdot), D$)
\ENDFOR
\STATE $best\_module \gets$ Max$(results_3)$
\FOR{$m_{sub} \gets$ GetSubModules($best\_module$)}
\STATE $results_3[m_{sub}] \gets$ Eval($f^{m_{sub}}_{lr\_narrow}(\cdot), D$)
\ENDFOR
\STATE \textbf{return} Max($results_1, results_2, results_3$)
\end{algorithmic}
\end{algorithm}
% $f_{*\_layer\_range}$ denotes the parameters of FFN and MHA in the $*\_layer\_range$ are trainable in model while others are frozen.
% $f^{m}_{*\_layer\_range}$ denotes the parameters of specific module in the $*\_layer\_range$ are trainable in model while others are frozen.
% \text{Max} function is to return the best strategy that achieves the best unified score among all three steps.

\subsection{Discussion of our setting}
\label{sec:setting_discussion}
% \begin{figure}[H]
%     \centering
%     \includegraphics[width=0.7\linewidth]{fig/Old_Procedure.png}
%     \caption{Typical setting on continual learning}
%     \label{fig:conventional_setting}
% \end{figure}
\def\cT{\mathcal{T}}
% 这边我们是想表达 sequential training 的话可能会大幅度破坏掉原来模型的能力
% Recall that the typical settings in continual learning considers sequentially fine-tuning a pre-trained model $f_{\theta_0}$ on tasks [$\mathcal{T}_1$, $\mathcal{T}_2$, ..., $\mathcal{T}_K$], and evaluation of forgetting is by measuring the model's performance on $\mathcal{T}_i$ after trained on $\mathcal{T}_j$ ($i < j$) 
% Differently, we only investigate the forgetting in generalized ability of the original model during fine-tuning a single task.
% We adopt this new setting because: 
Recall the conventional settings in continual learning, where a pre-trained model $f_{\theta_0}$ undergoes sequential fine-tuning on tasks [$\mathcal{T}_1$, $\mathcal{T}_2$, ..., $\mathcal{T}_K$], and forgetting evaluation is based on the model's performance on $\mathcal{T}_i$ after being trained on $\mathcal{T}_j$ ($i < j$).

In contrast, our investigation focuses solely on forgetting in the versatility of the original model during the fine-tuning of a single task.
This setting is motivated by two key considerations:
\begin{itemize}
    % \item The intrinsic capabilities of LLM are crucial and powerful, having been established through extensive pre-training on a massive amount of data, ranging from tens of billions to even hundreds of billions of samples. Adopting a sequential fine-tuning approach poses a significant risk of progressively undermining the original model's capabilities at each fine-tuning step~\citep{luo2023empirical,wang2023trace}. Moreover, the loss of the model's inherent capabilities is detrimental to subsequent task adaptation.
    % The versatility of aligned LLMs, established through extensive pre-training on a massive dataset ranging from tens of billions to even hundreds of billions of samples, is crucial and powerful. 
    \item Aligned LLMs have demonstrated remarkable versatility, achieved through extensive pre-training on a massive dataset comprising tens of billions to even hundreds of billions of samples. This enables them to effectively handle a wide range of real-world tasks~\citep{chowdhery2023palm,achiam2023gpt}.
    However, adopting a sequential fine-tuning method poses a significant risk of progressively undermining the model's versatility at each fine-tuning step~\citep{luo2023empirical, wang2023trace}. Additionally, the severe decline in versatility not only inhibits the fine-tuned
    performance of the model on diverse tasks~\citep{cheng2023adapting,dong2023abilities} but also hinders the effectiveness of transferring knowledge from it when gaining speciality~\citep{jie2022alleviating,luo2023investigating}.
    % is detrimental to the model's deployment for user-serving purposes and hinders the effectiveness of the speciality transfer learning procedure~\citep{jie2022alleviating, luo2023investigating}.
    % \item Furthermore, recent studies have shown that it is possible to replace the traditional approach of fine-tuning a sequence of tasks by independently fine-tuning each task on pre-trained model and subsequently combining their weights through interpolation \cite{ilharco2022editing, li2022branch, liu2023tangent,yu2023language}. This weight interpolation approach significantly enhances performance and mitigates catastrophic forgetting compared to conventional methods that sequentially fine-tune one model on $[\cT^{1}, \cT^{2}, \cT^{3}...]$. 
    \item Recent studies suggest an alternative to the traditional sequential fine-tuning approach. Instead of fine-tuning tasks sequentially, each task is independently fine-tuned on a pre-trained model, and their weights are subsequently combined through interpolation \cite{ilharco2022editing, li2022branch, liu2023tangent, yu2023language}. This weight interpolation approach significantly enhances performance and mitigates catastrophic forgetting compared to conventional methods that sequentially fine-tune one model on $[\mathcal{T}^{1}, \mathcal{T}^{2}, \mathcal{T}^{3}...]$.
\end{itemize}

\subsection{Baseline Descriptions}
\label{sec:baselines_descriptions}
% \subsection{Baselines}
In this Section, we describe the baseline method in our setting in detail.
We carefully categorize them into three classes:
\subsubsection{Regularization-based Methods}

% Approaches belonging to the 068
% former category share the consensus that in order to 069
% prevent CF, the finetuned model should be forced 070
% close to the pretrained model
% Model Consistency-Based Methods aim to prevent CF via forcing the parameters of fine-tuned model close to the pre-trained model.
% Model Consistency-Based Methods aim to mitigate catastrophic forgetting by constraining the parameters of the fine-tuned model to remain close to those of the pre-trained model.
% The selected baselines include Wise-FT~\citep{Wortsman2021wiseft}, L1 regularization~\citep{panigrahi2023task}, and L2 regularization~\citep{xuhong2018explicit}.
% The Wise-FT methodology comprises two distinct stages: initial fine-tuning of the pre-trained model $\theta_0$ on the specific downstream task to get fine-tuned model $\theta$ and subsequent amalgamation of the original pre-trained model $\theta_0$ and fine-tuned model $\theta$ via linear weight interpolation $f_{(1-\alpha) \theta_0 + \alpha \theta}$, where $\alpha$ represents a hyper-parameter ranging from 0 to 1. 
% , denoted as weight-space ensembling.

\paragraph{L1 Regularization}
\citet{panigrahi2023task} introduced explicit L1 regularization (with a strength of 0.001) on the parameter shift $\theta \rightarrow \hat{\theta}$, denoted as $|\hat{\theta} - \theta|$.
This regularization strategy helps alleviate catastrophic forgetting by limiting the extent of tuning.
% \citet{panigrahi2023task}  adds an explicit L1 regularization (with strength 0.001) on the parameter movement $\theta-\theta_0$, i.e., $|\theta - \theta_0|$ to recover the sparse grafts and encourage skill localization among model parameters, which further alleviates CF by constraining the degree of tuning.
% \citeauthor{panigrahi2023task} in \citeyear{panigrahi2023task} adds an explicit L1 regularization (with strength 0.001) on the parameter movement $\theta-\theta_0$, i.e., $|\theta - \theta_0|$ to recover the sparse grafts and encourage skill localization among model parameters, which further alleviates CF by constraining the degree of tuning. 

\paragraph{L2 Regularization}
% To cope with the inconsistency in transfer learning scenarios caused by the abuse of L2 regularization towards the origin, 
% Based on the idea of \citet{xuhong2018explicit}, \citet{lin2023speciality} advocated a coherent parameter regularization approach.
Building upon the concept introduced by \citet{xuhong2018explicit}, \citet{lin2023speciality} utilized a more consistent parameter regularization approach.
Specifically, the divergence between the parameters of the fine-tuned model $\hat{\theta}$ and the pre-trained model $\theta$ serves as the object of the L2 penalty in the optimization process, denoted as $|\hat{\theta} - \theta|_2^2$. This approach helps prevent deviation from the pre-trained model during tuning.

\subsubsection{Weight-based Methods}
\paragraph{Wise-FT}
The Wise-FT~\citep{Wortsman2021wiseft} methodology involves two distinct stages: initially fine-tuning the pre-trained model $\theta$ on a specific downstream task to obtain the fine-tuned model $\hat{\theta}$, followed by the fusion of the original pre-trained model $\theta$ and the fine-tuned model $\hat{\theta}$ using linear weight interpolation $f_{(1-\alpha) \theta + \alpha \hat{\theta}}$, where $\alpha$ is a hyper-parameter ranging from 0 to 1.
Parameter Constraint-Based Methods formulate their approach around parameter-specific strategies, leveraging either parameter efficiency techniques or the estimation of parameter importance. 

\paragraph{Vanilla Soft-masking}
\citet{ke2022continual} proposed Vanilla Soft-masking to address CF in continual domain pre-training of language models. Specifically, a gradient-based detection method is used to compute the importance value of units within the attention and FFN modules across all transformer layers for general domain knowledge. The resulting importance vector is then employed to control the backward gradient flow. 
Importantly, the soft-masking regulation is exclusively applied to the backward process, ensuring that knowledge transfer across domains during tuning remains unaffected.
% and soft-masks them based on their importance values to control the backward gradient flow. 
% Soft-masking contains the regulation to backward only, so that the knowledge transfer across domains during tuning would not be influenced.

\subsubsection{Architecture-based Methods}
\paragraph{LoRA}
Grounded in the assumption that the alteration in weights during model adaptation exhibits a low "intrinsic rank," 
\citet{Hu2021LoRALA} introduced a low-rank matrix into the dense layers within the network. 
This allows the indirect training of these layers by optimizing the decomposition matrices within the low-rank. Throughout training, all the decomposition matrices within the low-rank matrix are trainable, while the pre-trained weights remain frozen to preserve the original general ability of the pre-trained model.

\subsection{Statistics and Descriptions of Datasets}
\label{sec:static_description_dataset}
% Please add the following required packages to your document preamble:
% \usepackage{multirow}
% Please add the following required packages to your document preamble:
% \usepackage{multirow}
% Please add the following required packages to your document preamble:
% \usepackage{multirow}
\begin{table}[!h]
\centering
\begin{tabular}{clccc}
\toprule[1.5pt]
\multicolumn{1}{l}{}                              & \textbf{Dataset Name}            & \textbf{Train} & \textbf{Test}  & \textbf{Avg. Length} \\ \addlinespace[2pt] 
 \hline \addlinespace[2pt]
\multirow{2}{*}{\textbf{Knowledge Intensive Task}} & Finance                 & 14337 & 600   & 270.94      \\ 
                                                   & Law                     & 19400 & 600   & 277.82      \\ \addlinespace[2pt] \hdashline[1pt/1pt] \addlinespace[2pt]
\multirow{2}{*}{\textbf{Ability Intensive Task}}         & Math                    & 7366  & 250   & 216.95      \\ 
                                                   & CAG                     & 19400 & 600   & 331.24      \\ \addlinespace[2pt] \hdashline[1pt/1pt] \addlinespace[2pt]
\textbf{General Domain Knowledge}                         & CMMLU                   & -     & 11582 & 71.89       \\ \addlinespace[2pt] \hdashline[1pt/1pt] \addlinespace[2pt]
\multirow{4}{*}{\textbf{Generic Reasoning}}        & LogiQA                  & -     & 645   & 238.91      \\  
                                                   & LogiQA2                 & -     & 1589  & 224.25      \\  
                                                   & OCNLI                   & -     & 1995  & 82.17       \\  
                                                   & Zh-Winograd             & -     & 499   & 65.59       \\ \addlinespace[2pt] \hdashline[1pt/1pt] \addlinespace[2pt]
\textbf{Instruction Following}                     & Chinese GPT4-Instruct & -     & 1000  & 210.18      \\ \bottomrule[1.5pt]
\end{tabular}
\caption{The statistical information of the datasets involved in our setting. ``Train'' denotes the number of samples in the training set, ``Test'' denotes the number of samples in the test set, and ``Avg. Length'' denotes the average length of the dataset.}
\label{tab:statistics}
\end{table}

% data example 感觉可以先不用
% \nocite{tan2020tnt}
% \nocite{lu2023punifiedner}
% zhang2024improving,zhang2024question,li2024contextualization
Currently, LLM still falls short in certain areas~\citep{li2023multi,lu2023punifiedner}, we mainly select two types of tasks to evaluate its speciality. The overall statistics of the selected task datasets are shown in Table~\ref{tab:statistics}, and the detailed descriptions are as follows:
\subsubsection{Knowledge Intensive Tasks}
\begin{itemize}
    \item \textbf{Finance:} To create a Chinese financial QA instruction format dataset, we adopt the building procedure from~\citet{chen2023disc}. Initially, we translate the instruction content of the English FiQA\footnote{\url{https://huggingface.co/datasets/FinGPT/fingpt-fiqa_qa}} dataset into Chinese.
    Subsequently, we employ the translated Chinese instructions to generate responses that align with China’s national conditions using a robust Chinese FinLM\footnote{\url{https://huggingface.co/Go4miii/DISC-FinLLM}}~\citep{chen2023disc}.
\item \textbf{Law:}
    \citet{huang2023lawyer} initially gather Chinese legal questions from OpenLaw~\citep{DVN/OLO4G8_2018} and JEC-QA~\citep{zhong2019jec}. 
    They then employ ChatGPT to obtain responses, constructing the LlamaLawyer\footnote{\url{https://github.com/AndrewZhe/lawyer-llama}} dataset. This dataset comprises both single-turn and multi-turn instructions. For a fair comparison, we exclusively choose single-turn instructions in our setting.
\end{itemize}

\subsubsection{Ability Intensive Tasks}
\label{sec:CAG_task}
\begin{itemize}
    \item \textbf{Math:}
    \citet{chen2023breaking} establishes a multi-lingual GSM8k dataset (MGSM8k)\footnote{\url{https://mathoctopus.github.io/}} by translating the original English GSM8k~\citep{cobbe2021training} dataset into ten different languages using ChatGPT. In our setting, we specifically select the Chinese version within MGSM8k for both training and evaluation.
    \item \textbf{Context-Aware Generation (CAG\footnote{The CAG task essentially serves as the generation step in Retrieval-Augmented Generation (RAG), flexibly relying on the retrieved context to generate responses.}):}
    To enhance the model's capability of leveraging the context information to generate answers~\citep{tong2023eliminating,Tong2024CanLL}, we follow the methodology RA-DIT mentioned in \citet{lin2023ra} to construct an instruction format dataset based on Dureader~\citep{he-etal-2018-dureader}, a Chinese Reading Comprehension (RC) dataset that includes three types of questions: ``Entity'', ``Description'' and ``YesNo''. In our setting, we specifically choose ``Description'' type questions (Dureader-Desc), because their answers are human-written, presenting a significant difference from the context content. 
    Following the methodology of~\citet{chen2023disc}, we design a prompt (illustrated in Appendix~\ref{sec:rag_prompt}) that allows the model to flexibly generate appropriate responses under contexts with different properties. 
    We select the context that is irrelevant to the corresponding question to form a negative sample. This aims to enable the LLM to ignore misleading context content and lean into its parametric knowledge to respond. In contrast, we form positive samples by selecting context relevant to the corresponding question, intending to evaluate whether the LLM can better utilize relevant background knowledge to make a prediction. During the evaluation of the Context-Aware Generation (CAG) task, we use a subset of negative samples from the training data, paraphrase their instructions during inference, and utilize automatic generation metrics to assess the model's ability to leverage its parametric knowledge for response. We use positive samples outside the training data to evaluate the model's proficiency in utilizing context for response.
    
\end{itemize}
% we select the context that is relevant to the corresponding question to form a positive sample. This aims to enable the LLM to better utilize relevant background knowledge to make a prediction.
% When doing evaluation on CAG task, we select a subset of negative samples from training data and paraphrase their instructions to inference, and use the automatic generation metric to estimate whether the model can use its parametric knowledge to respond. We select the positive samples outside the training data to evaluate whether the model can better utilize the context to respond.

\subsubsection{General Domain knowledge}
We employ CMMLU~\citep{li2023cmmlu} for a zero-shot evaluation of the model's general domain knowledge.
CMMLU\footnote{\url{https://huggingface.co/datasets/haonan-li/cmmlu}} covers a wide range of subjects, comprising 67 topics that span from elementary to advanced professional levels. It includes subjects that require computational expertise, such as physics and chemistry, as well as disciplines within humanities and social sciences.
Notably, many tasks within CMMLU involve contextual nuances and wording that may not easily translate across languages. Moreover, numerous tasks have answers specific to China, making them contextually relevant but potentially not universally applicable or considered correct in other regions or languages.
% We utilize CMMLU~\citep{li2023cmmlu} to evaluate the general knowledge of the model in zero-shot manner.
% CMMLU covers a wide range of subjects, comprising 67 topics that span from elementary to advanced professional levels. It includes subjects that require computational expertise, such as physics and mathematics, as well as disciplines within humanities and social sciences.
% Many of these tasks are not easily translatable from
% other languages due to their specific contextual nuances and wording. Furthermore, numerous tasks within CMMLU have answers that are specific to
% China and may not be universally applicable or considered correct in other regions or languages.
% CMMLU has 67 subjects, including natural science, social sciences, engineering, and humanities, etc. 

\subsubsection{Generic Reasoning}
We assess the generic reasoning ability of LLM through the following datasets in a zero-shot manner:
\begin{itemize}
    \item \textbf{LogiQA\footnote{\url{https://github.com/lgw863/LogiQA-dataset}}:}
    The dataset follows a paragraph-question pair format, each accompanied by four candidate answers. It is sourced from publicly available logical examination papers for reading comprehension, designed by domain experts to assess participants' logical reasoning ability. Therefore, the questions exhibit reliable quality and cover a diverse range of topics.
    
    \item \textbf{LogiQA2\footnote{\url{https://github.com/csitfun/LogiQA2.0}}:}
    This is the second version of the LogiQA dataset, collected from the Chinese Civil Service Entrance Examination. The dataset includes newly released exam questions and practice questions, sourced from approximately 20 provinces in China where the exam is held annually. Exam materials are made publicly available on the Internet after each exam, and practice questions are obtained from various sources.
    \item \textbf{OCNLI\footnote{\url{https://github.com/CLUEbenchmark/OCNLI/}}:} The Original Chinese Natural Language Inference dataset (OCNLI) is the first extensive Natural Language Inference (NLI) dataset for Chinese. Unlike other datasets, OCNLI does not depend on automatic translation or non-expert annotation. Instead, it gathers annotations from native speakers with expertise in linguistics.
    \item \textbf{Zh-Winograd:} XWinograd~\citep{emelin-sennrich-2021-wino}\footnote{\url{https://huggingface.co/datasets/Muennighoff/xwinograd}} is a multilingual Winograd dataset designed for assessing commonsense reasoning and coreference resolution. In our setting, we exclusively choose the Chinese version for evaluation.
\end{itemize}

\subsubsection{Instruction Following Dataset}
Chinese GPT4-Instruct encompasses a variety of user-oriented instructions, spanning areas such as email writing, social media, and entertainment. This dataset was introduced by \citet{peng2023gpt4llm}, who rigorously followed the methodology outlined by \citet{alpaca} to implement the self-instruct strategy~\citep{selfinstruct} with GPT-4.
% Chinese GPT4-Instruct comprises diverse user-oriented instructions, encompassing areas like email writing, social media, and entertainment. 
% This dataset is introduced by \citet{peng2023instruction}, who meticulously followed the methodology introduced by \citet{alpaca} to implement the self-instruct~\citep{selfinstruct} strategy with GPT-4. 
% This approach was employed to craft a Chinese instruction dataset known as Chinese GPT4-Instruct, ensuring a high-quality and diverse collection of instructional content.
% \citet{peng2023instruction} meticulously followed the methodology introduced by \citet{alpaca} to implement the self-instruct~\citep{selfinstruct} strategy with GPT-4. 
% This approach was employed to craft a Chinese instruction dataset known as Chinese GPT4-Instruct, ensuring a high-quality and diverse collection of instructional content.
% \citet{peng2023instruction} follow the procedure of \citet{alpaca} to utilize the self-instruction~\citep{selfinstruct} strategy using GPT-4 to create a high-quality and diverse Chinese instruction dataset (Chinese GPT4-Instruct)\footnote{\url{https://huggingface.co/datasets/c-s-ale/alpaca-gpt4-data-zh}}.

\subsection{Implementation Details}
\label{sec:implementation_details}
% 可以讲 max length, 我们用了哪些框架
% lora 用的 learning rate，lora rank and alpha 要说一下

\begin{table}[!h]
\normalsize
\centering
\setlength \tabcolsep{4.5pt}
\begin{tabular}{lccclccclccc}
\toprule[1.5pt] \addlinespace[2pt]
        & \multicolumn{3}{c}{\textbf{7B}}     &  & \multicolumn{3}{c}{\textbf{13B}}    &  & \multicolumn{3}{c}{\textbf{33B}}    \\  \addlinespace[0.5pt] \cline{2-4} \cline{6-8} \cline{10-12} \addlinespace[2pt]
        & Epochs & Batch & LR &  
        & Epochs & Batch & LR & 
        & Epochs & Batch & LR \\ \hline \addlinespace[2pt]
Finance & 3      & 256        & 4e-5          &  & 5      & 256        & 4e-5          &  & 5      & 256        & 2e-5          \\
Law     & 3      & 256        & 4e-5          &  & 5      & 256        & 4e-5          &  & 5      & 256        & 2e-5          \\
Math    & 3      & 128        & 2e-5          &  & 5      & 128        & 2e-5          &  & 5      & 128        & 1e-5          \\
RAG     & 3      & 256        & 4e-5          &  & 5      & 256        & 4e-5          &  & 5      & 256        & 2e-5          \\ \bottomrule[1.5pt]
\end{tabular}
\caption{The Hyper-parameters involved in our training setting. LR denotes ``learning rate'', Batch denotes ``global batch size''.}
\label{tab:hyp}
\end{table}

We use \emph{DeepSpeed Zero3}~\citep{rajbhandari2021zero} and Adam optimizer~\citep{kingma2014adam} to train all the models.
Our experiments are conducted on a computing platform equipped with 32 V100 GPUs. Each experiment is run with three different random seeds, and the results are averaged to obtain the final outcome.
% Our experiments are conducted on a computing platform equipped with 32 V100 GPUs and each experiment is conducted with three different random seed and average the results to get the final result.
% on a computing platform with 32 V100 GPUs in our experiments.
The maximum sequence length is set to 512 across all four tasks, and we employ a greedy decoding strategy for generating results. A cosine scheduler with a 3\% warm-up period is applied.
In our study, LoRA is applied to five weight matrices: $\mathbf{W}_q$, $\mathbf{W}_k$, $\mathbf{W}_v$ in the MHA module, and $\mathbf{W}_{down}$, $\mathbf{W}_{up}$ in the FFN module. 
Following~\citet{lin2023speciality}, we experiment with $lora\_rank$ values in [4, 8, 16], setting $lora\_alpha = 2 \times lora\_rank$ to achieve optimal performance.
For Wise-FT, consistent with~\citet{lin2023speciality}, we explore $\alpha$ values in [0.4, 0.6, 0.8] to determine the best performance. Different learning rates and global batch sizes are set for training tasks across various model sizes, with detailed information provided in Table~\ref{tab:hyp}.
% We set the maximum sequence length to 512 when training the model in all four tasks and use the greedy decoding strategy for generating results.
% We used a cosine scheduler with a 3\% warm-up period.
% In our study, we apply LoRA to five weight matrices, which are $W_q$, $W_k$, $W_v$ in the MHA module, and $W_{down}$, $W_{up}$ in the FFN module.
% Following~\citet{lin2023speciality}, we try the $lora\_rank$ in [4, 8, 16] and set $lora\_alpha = 2 \times lora\_rank$ to get the best performance as the final results.
% For Wise-FT, following~\citet{lin2023speciality}, we try the $\alpha$ in [0.4, 0.6, 0.8] to get the best performance.
% We set different learning rates and global batch sizes when training different tasks under different model sizes, the detailed information is shown in Table~\ref{tab:hyp}.

% \small
% \centering
% \setlength \tabcolsep{1.8pt}

\subsection{Details of Different Model Scales}
\label{sec:different_size}
All the model adheres to the Apache-2.0 license, the details of different model scales are illustrated in Table~\ref{tab:model_sizes}.
\begin{table}[!h]
\centering
\begin{tabular}{lccc}
\bottomrule[1.5pt]
    & Dimension & Heads   & Layers \\ \addlinespace[2pt] \hline \addlinespace[2pt]
7B  & 4096      & 32      & 32       \\
13B & 5120      & 40      & 40       \\
33B & 6656      & 52      & 60       \\ \bottomrule[1.5pt]
\end{tabular}
\caption{The detailed architectural information for Chinese-Alpaca-Pro across various model scales.}
\label{tab:model_sizes}
\end{table}

\subsection{Prompts}
\label{sec:prompts}
\subsubsection{Instruction Prompt}
\label{sec:instruction_prompt}
\begin{framed}
\begin{quote}\em
Below is an instruction that describes a task. Write a response that appropriately completes the request.
\newline
\newline
\#\#\# Instruction:
\newline
\{instruction\} 
\newline
\{input\}
\newline
\newline
\#\#\# Response: \{output\}
\end{quote}
\end{framed}
We use ``\{instruction\}'', ``\{input\}'', and ``\{output\}'' to replace the specific instruction, input, and output.

\subsubsection{CAG Prompt}
\label{sec:rag_prompt}
\begin{framed}
\begin{quote}\em
\begin{CJK*}{UTF8}{gbsn} 请根据参考材料回答下面的问题。下面的材料可供参考。 \end{CJK*}
\\ ``Please answer the following questions based on the reference materials. The following materials are available for reference.'' 
\\ \begin{CJK*}{UTF8}{gbsn} （注意：1、材料可能与问题无关，请忽略无关材料，并基于已有知识回答问题。 2、尽量不要直接复制材料内容作为答案，而应该将材料内容作为事件的补充与潜在分析，启发思考）；\end{CJK*}
\\ ``(Note: 1. The material may have nothing to do with the question. Please ignore the irrelevant material and answer the question based on existing knowledge. 2. Try not to directly copy the material content as an answer, but use the material content as a supplement and potential analysis of the event to inspire thinking. );''
\\ \begin{CJK*}{UTF8}{gbsn} 参考材料:\end{CJK*} \\
``references:'' \\
\{reference\} \\
\begin{CJK*}{UTF8}{gbsn} 问题: \end{CJK*}
\\ ``question:'' \\
\{question\} 
\end{quote}
\end{framed}

Here, the contents labeled with ``'' contain the corresponding English translations of Chinese.
We use ``\{reference\}'' and ``\{question\}'' to replace the specific reference and question.

\subsection{Evaluation Metrics}
\label{sec:appendix_evaluation_metrics}
% 这边主要得去解释为什么要用 generation score 
% 可以讲一下我们用了哪些库和框架 (感觉也不用)
% RAG:对于需要利用上下文的，我们是不要从 training dataset sample 的，对于不利用上下文的，我们需要从 training dataset 里面 sample (这个在 dataset 那边以及讲过了)
\subsubsection{Instruction Following Score}
\label{App:log-likihood}
Following LMflow \cite{diao2023lmflow}, we assess instruction following performance using log-likelihood:
\begin{align*}
    \mbox{NLL} = &  -\frac{1}{N} \sum_{i=1}^N \log p(\mbox{sentence}_i | \mbox{context}_i) \\
    = & -\frac{1}{N} \sum_{i=1}^N \log p(\mbox{token}_{i, 1}, \mbox{token}_{i, 2}, ...,  \mbox{token}_{i, n_i} | \mbox{context}_i), 
\end{align*}
where $n_i$ is the length of the token in $\mbox{sentence}_i$. 
As the $\mbox{LL}$ results are typically in the order of several hundreds (e.g., 531.27), and the results of Accuracy or automatic generation metrics are values ranging from 0 to 1, they inherently possess disparate scales.
To comprehensively assess both the speciality and versatility of LLM, we initially scale the $\mbox{LL}$ by dividing it by one thousand. Since a lower $\mbox{LL}$ is indicative of superior performance, we then subtract the result from 1 to obtain the final instruction following (Instruct.) score, i.e., $\textrm{Instruct.} = 1- \mbox{LL}/1000$.

\subsubsection{Speciality Score}
\label{sec:specialized_score}
For Finance, Law, and CAG tasks, we follow~\citet{tan-etal-2020-tnt,tan2021bert,zhang2022fine,kong2022blcu,kong2022multitasking,zhang2023assisting} to utilize the automatic metrics to evaluate the quality of generated text~, i.e., BERTScore\footnote{\url{https://github.com/Tiiiger/bert_score}}, BLEU\footnote{\url{https://www.nltk.org/api/nltk.translate.bleu_score.html}}, and Rouge\footnote{\url{https://github.com/pltrdy/rouge}} to evaluate their speciality performance. 
The Rouge metric comprises Rouge-1, Rouge-2, and Rouge-L. We initially average the results of Rouge-1, Rouge-2, and Rouge-L to obtain the overall Rouge result. Subsequently, we further average the results of BERTScore, BLEU, and Rouge to derive the final speciality score. It's worth noting that we employ the F1 score in both Rouge and BERTScore.
\begin{align}
\text{Rouge} &= \frac{\text{Rouge-1} + \text{Rouge-2} + \text{Rouge-L}}{3} \nonumber \\
\text{Spec.} &= \frac{\text{BERTScore} + \text{Rouge} + \text{BLEU}}{3}
\end{align}
For Math, as it only has one metric, i.e., Accuracy, we directly use its Accuracy result as the Spec. score.

\subsubsection{Versatility Score}
\label{sec:generalized_score}
We assess the versatility of LLM in three aspects: general domain knowledge (Gen-Kn.), generic reasoning (Gen-Rs.), and instruction following (Instruct.).
For evaluating the generic reasoning of LLM, we average the results from four datasets to obtain an overall generic reasoning (Gen-Rs.) score.
Subsequently, we average the scores of Gen-Kn., Gen-Rs., and Instruct. (computed in Appendix~\ref{App:log-likihood}) to derive the overall versatility (Vers.) score for LLM.

\begin{align}
\text{Gen-Rs.} &= \frac{\text{LogiQA} + \text{LogiQA2} + \text{OCNLI} + \text{Zh-Winograd}}{4} \nonumber \\
\text{Vers.} &= \frac{\text{Gen-Kn.} + \text{Gen-Rs.} + \text{Instruct.}}{3}
\end{align}

\subsubsection{Unified Score}
% To facilitate the evaluation of LLM on both speciality and versatility, we directly sum the results of the speciality (Spec.) score obtained in Appendix~\ref{sec:specialized_score} and the versatility (Vers.) score obtained in Appendix~\ref{sec:generalized_score} to get the final unified (Uni.) score:
To facilitate the evaluation of LLM in both speciality and versatility, we sum the results of the speciality (Spec.) score (obtained in Appendix~\ref{sec:specialized_score}) and the versatility (Vers.) score (obtained in Appendix~\ref{sec:generalized_score}) to get the final unified (Uni.) score:
\begin{align}
\text{Uni.} &= \text{Spec.} + \text{Vers.} 
\end{align}

\onecolumn\newpage
\subsection{Ablation Results of Fine-SoftMask}
\label{sec:appendix_softmask}
\begin{table}[!h]
\centering
\begin{tabular}{lccccc}
\toprule[1.5pt]
                               & \multicolumn{2}{c}{{\color[HTML]{333333} \textit{CoFiTune}}}                    & \textit{} & \multicolumn{2}{c}{{\color[HTML]{333333} \textit{\begin{tabular}[c]{@{}c@{}}CoFiTune \\ w/o Fine-SoftMask\end{tabular}}}} \\ \addlinespace[2pt] \cline{2-3} \cline{5-6} \addlinespace[2pt]
                               & {\color[HTML]{333333} Spec.}           & {\color[HTML]{333333} Vers.}           &           & {\color[HTML]{333333} Spec.}                                & {\color[HTML]{333333} Vers.}                                \\ \hline \addlinespace[2pt]
{\color[HTML]{333333} Finance} & {\color[HTML]{333333} 0.4218} & {\color[HTML]{333333} 0.4683} &           & {\color[HTML]{333333} 0.4220}                      & {\color[HTML]{333333} 0.4653}                      \\
{\color[HTML]{333333} Law}     & {\color[HTML]{333333} 0.4328}          & {\color[HTML]{333333} 0.4665}          &           & {\color[HTML]{333333} 0.4331}                               & {\color[HTML]{333333} 0.4626}                               \\
{\color[HTML]{333333} Math}    & {\color[HTML]{333333} 0.0800}          & {\color[HTML]{333333} 0.4797}          &           & {\color[HTML]{333333} 0.0800}                               & {\color[HTML]{333333} 0.4765}                               \\
{\color[HTML]{333333} CAG}     & {\color[HTML]{333333} 0.5166}          & {\color[HTML]{333333} 0.4716}          &           & {\color[HTML]{333333} 0.5159}                               & {\color[HTML]{333333} 0.4684}                               \\ \bottomrule[1.5pt]
\end{tabular}
\caption{The impact of Fine-SoftMask in \emph{CoFiTune} on Spec. and Vers. scores under 7B model. ``w/o'' means excluding the technique from \emph{CoFiTune}.}
\label{tab:abl_softmask_7b}
\end{table}

\onecolumn\newpage
\subsection{Exploration Results in Coarse Level}
\label{sec:appendix_layer_param}
\subsubsection{7B Results}
% \onecolumn
\begin{table}[!h]
\centering
\setlength\tabcolsep{4.5pt}
\fontsize{10}{14}\selectfont 
\begin{tabular}{lccc|ccc|c}
\hline
                        & {\color[HTML]{333333} BERTScore} & {\color[HTML]{333333} Rouge} & {\color[HTML]{333333} BLEU} & Rouge-1 & Rouge-2 & Rouge-L & {\color[HTML]{333333} Spec.} \\ \hline
(0, 16{]} - MHA \& FFN  & 0.7283                           & 0.3774                       & 0.1843                      & 0.5314  & 0.2512  & 0.3498  & 0.4300                       \\
(8, 24{]} - MHA \& FFN  & 0.7298                           & 0.3814                       & 0.1884                      & 0.5339  & 0.2575  & 0.3528  & 0.4332                       \\
(16, 32{]} - MHA \& FFN & 0.7162                           & 0.3558                       & 0.1649                      & 0.5048  & 0.2324  & 0.3303  & 0.4123                       \\
(4, 20{]} - MHA \& FFN  & 0.7306                           & 0.3799                       & 0.1887                      & 0.5314  & 0.2562  & 0.3521  & 0.4331                       \\
                        &                                  &                              &                             &         &         &         &                              \\
(8, 16{]} - MHA \& FFN  & 0.7266                           & 0.3721                       & 0.1807                      & 0.5262  & 0.2466  & 0.3434  & 0.4264                       \\
(16, 24{]} - MHA \& FFN & 0.7151                           & 0.3524                       & 0.1629                      & 0.5048  & 0.2273  & 0.3251  & 0.4101                       \\
\hline
\end{tabular}
\caption{The detailed speciality performance of Finance task under the 7B model in different exploring layer ranges and modules.}
\end{table}

\begin{table}[!h]
\centering
\setlength\tabcolsep{2pt}
\fontsize{10}{14}\selectfont 
\begin{tabular}{lccc|cccc|c}
\hline
                        & {\color[HTML]{333333} Instruct.} & {\color[HTML]{333333} Gen-Kn.} & {\color[HTML]{333333} Gen-Rs.} & Logiqa1 & Logiqa2 & OCNLI   & {\color[HTML]{333333} Zh\_Winograd} & {\color[HTML]{333333} Vers.} \\ \hline
(0, 16{]} - MHA \& FFN  & 0.5375                           & 0.3172                         & 0.4269                         & 0.2496  & 0.2801  & 0.4547  & 0.7234                              & 0.4272                       \\
(8, 24{]} - MHA \& FFN  & 0.5447                           & 0.3524                         & 0.4409                         & 0.2775  & 0.3255  & 0.4503  & 0.7104                              & 0.4460                       \\
(16, 32{]} - MHA \& FFN & 0.5579                           & 0.3609                         & 0.4535                         & 0.2759  & 0.3386  & 0.4461  & 0.7535                              & 0.4574                       \\
(4, 20{]} - MHA \& FFN  & 0.5426                           & 0.3414                         & 0.4338                         & 0.2682  & 0.3033  & 0.4484  & 0.7152                              & 0.4392                       \\
                        &                                  &                                &                                &         &         &         &                                     &                              \\
(8, 16{]} - MHA \& FFN  & 0.5770                           & 0.3518                         & 0.4437                         & 0.2774  & 0.3158  & 0.4464  & 0.7356                              & 0.4575                       \\
(16, 24{]} - MHA \& FFN & 0.5858                           & 0.3609                         & 0.4515                         & 0.2729  & 0.3361  & 0.4536  & 0.7434                              & 0.4661                       \\
\hline
\end{tabular}
\caption{The detailed versatility performance of Finance task under the 7B model in different exploring layer ranges and modules.}
\end{table}

\begin{table}[!h]
\centering
\setlength\tabcolsep{4.5pt}
\fontsize{10}{14}\selectfont 
\begin{tabular}{lccc|ccc|c}
\hline
                        & {\color[HTML]{333333} BERTScore} & {\color[HTML]{333333} Rouge} & {\color[HTML]{333333} BLEU} & Rouge-1 & Rouge-2 & Rouge-L & {\color[HTML]{333333} Spec.} \\ \hline
(0, 16{]} - MHA \& FFN  & 0.7313                           & 0.3846                       & 0.2092                      & 0.5198  & 0.2579  & 0.3762  & 0.4417                       \\
(8, 24{]} - MHA \& FFN  & 0.7341                           & 0.3903                       & 0.2152                      & 0.5321  & 0.2645  & 0.3843  & 0.4465                       \\
(16, 32{]} - MHA \& FFN & 0.7200                           & 0.3621                       & 0.1739                      & 0.4951  & 0.2363  & 0.3550  & 0.4187                       \\
(4, 20{]} - MHA \& FFN  & 0.7329                           & 0.3874                       & 0.2126                      & 0.5179  & 0.2615  & 0.3828  & 0.4443                       \\
                        &                                  &                              &                             &         &         &         &                              \\
(8, 16{]} - MHA \& FFN  & 0.7293                           & 0.3848                       & 0.2007                      & 0.5184  & 0.2590  & 0.3772  & 0.4383                       \\
(16, 24{]} - MHA \& FFN & 0.7189                           & 0.3598                       & 0.1711                      & 0.4941  & 0.2340  & 0.3513  & 0.4166                       \\
\hline
\end{tabular}
\caption{The detailed speciality performance of Law task under the 7B model in different exploring layer ranges and modules.}
\end{table}

% Please add the following required packages to your document preamble:
% \usepackage[table,xcdraw]{xcolor}
% Beamer presentation requires \usepackage{colortbl} instead of \usepackage[table,xcdraw]{xcolor}
\begin{table}[!h]
\centering
\setlength\tabcolsep{2pt}
\fontsize{10}{14}\selectfont 
\begin{tabular}{lccc|cccc|c}
\hline
                        & {\color[HTML]{333333} Instruct.} & {\color[HTML]{333333} Gen-Kn.} & {\color[HTML]{333333} Gen-Rs.} & Logiqa1 & Logiqa2 & OCNLI  & {\color[HTML]{333333} Zh\_Winograd} & {\color[HTML]{333333} Vers.} \\ \hline
(0, 16{]} - MHA \& FFN  & 0.4847                           & 0.3189                         & 0.4225                         & 0.2931  & 0.3052  & 0.4064 & 0.6453                              & 0.4087                       \\
(8, 24{]} - MHA \& FFN  & 0.4943                           & 0.3488                         & 0.4460                         & 0.3147  & 0.3205  & 0.4174 & 0.7314                              & 0.4297                       \\
(16, 32{]} - MHA \& FFN & 0.5448                           & 0.3588                         & 0.4533                         & 0.2992  & 0.3247  & 0.4416 & 0.7475                              & 0.4523                       \\
(4, 20{]} - MHA \& FFN  & 0.4902                           & 0.3240                         & 0.4368                         & 0.3013  & 0.3138  & 0.4109 & 0.7217                              & 0.4170                       \\
                        &                                  &                                &                                &         &         &        &                                     &                              \\
(8, 16{]} - MHA \& FFN  & 0.5454                           & 0.3512                         & 0.4487                         & 0.3133  & 0.3236  & 0.4224 & 0.7354                              & 0.4484                       \\
(16, 24{]} - MHA \& FFN & 0.5791                           & 0.3591                         & 0.4578                         & 0.3019  & 0.3327  & 0.4471 & 0.7495                              & 0.4653                       \\
\hline
\end{tabular}
\caption{The detailed versatility performance of Law task under the 7B model in different exploring layer ranges and modules.}
\end{table}

\begin{table}[!h]
\centering
\begin{tabular}{lc}
\hline
                        & {\color[HTML]{333333} Accuracy} \\ \hline
(0, 16{]} - MHA \& FFN  & 0.084                           \\
(8, 24{]} - MHA \& FFN  & 0.088                           \\
(16, 32{]} - MHA \& FFN & 0.064                           \\
(4, 20{]} - MHA \& FFN  & 0.088                           \\
                        &                                 \\
(8, 16{]} - MHA \& FFN  & 0.084                           \\
(16, 24{]} - MHA \& FFN & 0.056                           \\
(12, 20{]} - MHA \& FFN & 0.072                           \\
(10, 18{]} - MHA \& FFN & 0.076                           \\
                        &                                 \\
(8, 16{]} - FFN         & 0.080                           \\
(8, 16{]} - MHA         & 0.056                           \\
(8, 16{]} - down\_proj  & 0.072                           \\
(8, 16{]} - up\_proj    & 0.064                           \\ \hline
\end{tabular}
\caption{The detailed speciality performance of Math task under the 7B model in different exploring layer ranges and modules.}
\end{table}

\begin{table}[!h]
\centering
\setlength\tabcolsep{2pt}
\fontsize{10}{14}\selectfont 
\begin{tabular}{lccc|cccc|c}
\hline
                        & {\color[HTML]{333333} Instruct.} & {\color[HTML]{333333} Gen-Kn.} & {\color[HTML]{333333} Gen-Rs.} & Logiqa1 & Logiqa2 & OCNLI  & {\color[HTML]{333333} Zh\_Winograd} & {\color[HTML]{333333} Vers.} \\ \hline
(0, 16{]} - MHA \& FFN  & 0.5554                           & 0.3282                         & 0.4312                         & 0.3029  & 0.2917  & 0.4005 & 0.7296                              & 0.4383                       \\
(8, 24{]} - MHA \& FFN  & 0.5617                           & 0.3539                         & 0.4535                         & 0.3008  & 0.3304  & 0.4423 & 0.7408                              & 0.4564                       \\
(16, 32{]} - MHA \& FFN & 0.6043                           & 0.3611                         & 0.4616                         & 0.3054  & 0.3449  & 0.4507 & 0.7455                              & 0.4756                       \\
(4, 20{]} - MHA \& FFN  & 0.5579                           & 0.3352                         & 0.4331                         & 0.3085  & 0.2964  & 0.3959 & 0.7314                              & 0.4421                       \\
                        &                                  &                                &                                &         &         &        &                                     &                              \\
(8, 16{]} - MHA \& FFN  & 0.5958                           & 0.3556                         & 0.4551                         & 0.3194  & 0.3235  & 0.4341 & 0.7435                              & 0.4688                       \\
(16, 24{]} - MHA \& FFN & 0.6141                           & 0.3629                         & 0.4646                         & 0.3108  & 0.3486  & 0.4537 & 0.7452                              & 0.4805                       \\
\hline
\end{tabular}
\caption{The detailed versatility performance of Math task under the 7B model in different exploring layer ranges and modules.}
\end{table}

\begin{table}[!h]
\centering
\setlength\tabcolsep{4.5pt}
\fontsize{10}{14}\selectfont 
\begin{tabular}{lccc|ccc|c}
\hline
                        & {\color[HTML]{333333} BERTScore} & {\color[HTML]{333333} Rouge} & {\color[HTML]{333333} BLEU} & Rouge-1 & Rouge-2 & Rouge-L & {\color[HTML]{333333} Spec.} \\ \hline
(0, 16{]} - MHA \& FFN  & 0.7381                           & 0.4794                       & 0.3473                      & 0.5481  & 0.3987  & 0.4913  & 0.5216                       \\
(8, 24{]} - MHA \& FFN  & 0.7395                           & 0.4849                       & 0.3548                      & 0.5517  & 0.4051  & 0.4978  & 0.5264                       \\
(16, 32{]} - MHA \& FFN & 0.7277                           & 0.4604                       & 0.3252                      & 0.5248  & 0.3803  & 0.4762  & 0.5044                       \\
(4, 20{]} - MHA \& FFN  & 0.7376                           & 0.4856                       & 0.3555                      & 0.5519  & 0.4064  & 0.4986  & 0.5262                       \\
                        &                                  &                              &                             &         &         &         &                              \\
(8, 16{]} - MHA \& FFN  & 0.7369                           & 0.4776                       & 0.3457                      & 0.5455  & 0.3973  & 0.4901  & 0.5200                       \\
(16, 24{]} - MHA \& FFN & 0.7261                           & 0.4594                       & 0.3286                      & 0.5234  & 0.3795  & 0.4753  & 0.5047                       \\
\hline
\end{tabular}
\caption{The detailed speciality performance of CAG task under the 7B model in different exploring layer ranges and modules.}
\end{table}

\begin{table}[!h]
\centering
\setlength\tabcolsep{2pt}
\fontsize{10}{14}\selectfont 
\begin{tabular}{lccc|cccc|c}
\hline
                        & {\color[HTML]{333333} Instruct.} & {\color[HTML]{333333} Gen-Kn.} & {\color[HTML]{333333} Gen-Rs.} & Logiqa1 & Logiqa2 & OCNLI  & {\color[HTML]{333333} Zh\_Winograd} & {\color[HTML]{333333} Vers.} \\ \hline
(0, 16{]} - MHA \& FFN  & 0.5508                           & 0.3124                         & 0.4263                         & 0.2992  & 0.3071  & 0.4276 & 0.6713                              & 0.4298                       \\
(8, 24{]} - MHA \& FFN  & 0.5555                           & 0.3497                         & 0.4449                         & 0.3037  & 0.3158  & 0.4308 & 0.7294                              & 0.4500                       \\
(16, 32{]} - MHA \& FFN & 0.5707                           & 0.3576                         & 0.4538                         & 0.3068  & 0.3254  & 0.4354 & 0.7475                              & 0.4607                       \\
(4, 20{]} - MHA \& FFN  & 0.5532                           & 0.3404                         & 0.4366                         & 0.3019  & 0.3059  & 0.4233 & 0.7154                              & 0.4434                       \\
                        &                                  &                                &                                &         &         &        &                                     &                              \\
(8, 16{]} - MHA \& FFN  & 0.5846                           & 0.3508                         & 0.4462                         & 0.3058  & 0.3121  & 0.4294 & 0.7374                              & 0.4605                       \\
(16, 24{]} - MHA \& FFN & 0.5954                           & 0.3589                         & 0.4523                         & 0.3085  & 0.3201  & 0.4411 & 0.7396                              & 0.4689                       \\
\hline
\end{tabular}
\caption{The detailed versatility performance of CAG task under the 7B model in different exploring layer ranges and modules.}
\end{table}

% \paragraph{Knowledge Intensive Tasks}

% 先onecolumn 再 new page
\onecolumn
\newpage
\subsubsection{13B Results}
% \onecolumn
\begin{table}[!h]
\centering
\setlength\tabcolsep{4.5pt}
\fontsize{10}{14}\selectfont 
\begin{tabular}{lccc|ccc|c}
\hline
                        & {\color[HTML]{333333} BERTScore} & {\color[HTML]{333333} Rouge} & {\color[HTML]{333333} BLEU} & Rouge-1 & Rouge-2 & Rouge-L & {\color[HTML]{333333} Spec.} \\ \hline
(0, 20{]} - MHA \& FFN  & 0.7418                           & 0.4045                       & 0.2204                      & 0.5525  & 0.2823  & 0.3786  & 0.4555                       \\
(10, 30{]} - MHA \& FFN & 0.7417                           & 0.4060                       & 0.2223                      & 0.5537  & 0.284   & 0.3802  & 0.4567                       \\
(20, 40{]} - MHA \& FFN & 0.7225                           & 0.3683                       & 0.1802                      & 0.5213  & 0.2424  & 0.3413  & 0.4237                       \\
(5, 25{]} - MHA \& FFN  & 0.7416                           & 0.4071                       & 0.2238                      & 0.5533  & 0.2846  & 0.3829  & 0.4574                       \\
                        &                                  &                              &                             &         &         &         &                              \\
(10, 20{]} - MHA \& FFN & 0.7341                           & 0.3893                       & 0.2056                      & 0.5383  & 0.2651  & 0.3645  & 0.4430                       \\
(20, 30{]} - MHA \& FFN & 0.7203                           & 0.3683                       & 0.1756                      & 0.5192  & 0.2444  & 0.3415  & 0.4214                       \\
\hline
\end{tabular}
\caption{The detailed speciality performance of Finance task under the 13B model in different exploring layer ranges and modules.}
\end{table}

\begin{table}[!h]
\centering
\setlength\tabcolsep{2pt}
\fontsize{10}{14}\selectfont 
\begin{tabular}{lccc|cccc|c}
\hline
                        & {\color[HTML]{333333} Instruct.} & {\color[HTML]{333333} Gen-Kn.} & {\color[HTML]{333333} Gen-Rs.} & Logiqa1 & Logiqa2 & OCNLI  & {\color[HTML]{333333} Zh\_Winograd} & {\color[HTML]{333333} Vers.} \\ \hline
(0, 20{]} - MHA \& FFN  & 0.4511                           & 0.3582                         & 0.4584                         & 0.3023  & 0.3386  & 0.5133 & 0.6793                              & 0.4226                       \\
(10, 30{]} - MHA \& FFN & 0.4643                           & 0.3993                         & 0.4901                         & 0.3235  & 0.3946  & 0.5328 & 0.7095                              & 0.4512                       \\
(20, 40{]} - MHA \& FFN & 0.5215                           & 0.4065                         & 0.5008                         & 0.3581  & 0.4179  & 0.5141 & 0.7137                              & 0.4762                       \\
(5, 25{]} - MHA \& FFN  & 0.4562                           & 0.3839                         & 0.4687                         & 0.3132  & 0.3801  & 0.5042 & 0.6772                              & 0.4363                       \\
                        &                                  &                                &                                &         &         &        &                                     &                              \\
(10, 20{]} - MHA \& FFN & 0.5238                           & 0.4007                         & 0.4935                         & 0.3230  & 0.4093  & 0.5454 & 0.6964                              & 0.4727                       \\
(20, 30{]} - MHA \& FFN & 0.5615                           & 0.4071                         & 0.5001                         & 0.3156  & 0.4021  & 0.5875 & 0.6951                              & 0.4896                       \\
\hline
\end{tabular}
\caption{The detailed versatility performance of Finance task under the 13B model in different exploring layer ranges and modules.}
\end{table}

\begin{table}[!h]
\centering
\setlength\tabcolsep{4.5pt}
\fontsize{10}{14}\selectfont 
\begin{tabular}{lccc|ccc|c}
\hline
                        & {\color[HTML]{333333} BERTScore} & {\color[HTML]{333333} Rouge} & {\color[HTML]{333333} BLEU} & Rouge-1 & Rouge-2 & Rouge-L & {\color[HTML]{333333} Spec.} \\ \hline
(0, 20{]} - MHA \& FFN  & 0.7472                           & 0.4341                       & 0.2261                      & 0.5585  & 0.3142  & 0.4297  & 0.4691                       \\
(10, 30{]} - MHA \& FFN & 0.7486                           & 0.4363                       & 0.2284                      & 0.5594  & 0.3171  & 0.4325  & 0.4711                       \\
(20, 40{]} - MHA \& FFN & 0.7253                           & 0.3946                       & 0.1848                      & 0.5231  & 0.2717  & 0.3890  & 0.4349                       \\
(5, 25{]} - MHA \& FFN  & 0.7459                           & 0.4354                       & 0.2303                      & 0.5577  & 0.3171  & 0.4315  & 0.4705                       \\
                        &                                  &                              &                             &         &         &         &                              \\
(10, 20{]} - MHA \& FFN & 0.7404                           & 0.4201                       & 0.2102                      & 0.5458  & 0.2995  & 0.4151  & 0.4569                       \\
(20, 30{]} - MHA \& FFN & 0.7318                           & 0.3899                       & 0.1748                      & 0.5210  & 0.2645  & 0.3843  & 0.4322                       \\
\hline
\end{tabular}
\caption{The detailed speciality performance of Law task under the 13B model in different exploring layer ranges and modules.}
\end{table}

\begin{table}[!h]
\centering
\setlength\tabcolsep{2pt}
\fontsize{10}{14}\selectfont 
\begin{tabular}{lccc|cccc|c}
\hline
                        & {\color[HTML]{333333} Instruct.} & {\color[HTML]{333333} Gen-Kn.} & {\color[HTML]{333333} Gen-Rs.} & Logiqa1 & Logiqa2 & OCNLI  & {\color[HTML]{333333} Zh\_Winograd} & {\color[HTML]{333333} Vers.} \\ \hline
(0, 20{]} - MHA \& FFN  & 0.4405                           & 0.3614                         & 0.4511                         & 0.3116  & 0.3606  & 0.4666 & 0.6653                              & 0.4176                       \\
(10, 30{]} - MHA \& FFN & 0.4608                           & 0.3943                         & 0.4874                         & 0.3329  & 0.4150  & 0.5273 & 0.6758                              & 0.4462                       \\
(20, 40{]} - MHA \& FFN & 0.5074                           & 0.4065                         & 0.5014                         & 0.3551  & 0.4147  & 0.5441 & 0.6914                              & 0.4718                       \\
(5, 25{]} - MHA \& FFN  & 0.4463                           & 0.3757                         & 0.4646                         & 0.3132  & 0.3939  & 0.4742 & 0.6775                              & 0.4289                       \\
                        &                                  &                                &                                &         &         &        &                                     &                              \\
(10, 20{]} - MHA \& FFN & 0.5006                           & 0.4006                         & 0.4934                         & 0.3547  & 0.4103  & 0.5092 & 0.6993                              & 0.4649                       \\
(20, 30{]} - MHA \& FFN & 0.5474                           & 0.4052                         & 0.5015                         & 0.3690  & 0.4122  & 0.5213 & 0.7034                              & 0.4847                       \\
\hline
\end{tabular}
\caption{The detailed versatility performance of Law task under the 13B model in different exploring layer ranges and modules.}
\end{table}

\begin{table}[!h]
\centering
\setlength\tabcolsep{4.5pt}
\fontsize{10}{14}\selectfont 
\begin{tabular}{l|c}
\hline
                        & \multicolumn{1}{c}{{\color[HTML]{333333} Accuracy}} \\ \hline
(0, 20{]} - MHA \& FFN  & 0.112                                               \\
(10, 30{]} - MHA \& FFN & 0.120                                               \\
(20, 40{]} - MHA \& FFN & 0.068                                               \\
(5, 25{]} - MHA \& FFN  & 0.116                                               \\
                        &                                                     \\
(10, 20{]} - MHA \& FFN & 0.112                                               \\
(20, 30{]} - MHA \& FFN & 0.056                                               \\
(15, 25{]} - MHA \& FFN & 0.076                                               \\
(13, 23{]} - MHA \& FFN & 0.100                                               \\
                        & \multicolumn{1}{c}{}                                \\
(10, 20{]} - FFN        & 0.108                                               \\
(10, 20{]} - MHA        & 0.064                                               \\
(10, 20{]} - down\_proj & 0.076                                               \\
(10, 20{]} - up\_proj   & 0.068                                               \\ \hline
\end{tabular}
\caption{The detailed speciality performance of Math task under the 13B model in different exploring layer ranges and modules.}
\end{table}

\begin{table}[!h]
\centering
\setlength\tabcolsep{2pt}
\fontsize{10}{14}\selectfont 
\begin{tabular}{lccc|cccc|c}
\hline
                        & {\color[HTML]{333333} Instruct.} & {\color[HTML]{333333} Gen-Kn.} & {\color[HTML]{333333} Gen-Rs.} & Logiqa1 & Logiqa2 & OCNLI  & {\color[HTML]{333333} Zh\_Winograd} & {\color[HTML]{333333} Vers.} \\ \hline
(0, 20{]} - MHA \& FFN  & 0.5793                           & 0.3619                         & 0.4528                         & 0.3101  & 0.3493  & 0.4286 & 0.7234                              & 0.4647                       \\
(10, 30{]} - MHA \& FFN & 0.5879                           & 0.3936                         & 0.5011                         & 0.3602  & 0.4021  & 0.5208 & 0.7215                              & 0.4942                       \\
(20, 40{]} - MHA \& FFN & 0.6090                           & 0.4017                         & 0.5115                         & 0.3683  & 0.4158  & 0.5305 & 0.7317                              & 0.5074                       \\
(5, 25{]} - MHA \& FFN  & 0.5838                           & 0.3875                         & 0.4718                         & 0.3294  & 0.3933  & 0.4311 & 0.7331                             & 0.4810                       \\
                        &                                  &                                &                                &         &         &        &                                     &                              \\
(10, 20{]} - MHA \& FFN & 0.6107                           & 0.3971                         & 0.5049                         & 0.3535  & 0.4098  & 0.5188 & 0.7375                              & 0.5043                       \\
(20, 30{]} - MHA \& FFN & 0.6202                           & 0.4032                         & 0.5141                         & 0.3705  & 0.4179  & 0.5273 & 0.7404                              & 0.5125                       \\
\hline
\end{tabular}
\caption{The detailed versatility performance of Math task under the 13B model in different exploring layer ranges and modules.}
\end{table}

\begin{table}[!h]
\centering
\setlength\tabcolsep{4.5pt}
\fontsize{10}{14}\selectfont 
\begin{tabular}{lccc|ccc|c}
\hline
                        & {\color[HTML]{333333} BERTScore} & {\color[HTML]{333333} Rouge} & {\color[HTML]{333333} BLEU} & Rouge-1 & Rouge-2 & Rouge-L & {\color[HTML]{333333} Spec.} \\ \hline
(0, 20{]} - MHA \& FFN  & 0.7487                           & 0.5133                       & 0.3845                      & 0.5791  & 0.4313  & 0.5294  & 0.5488                       \\
(10, 30{]} - MHA \& FFN & 0.7493                           & 0.5185                       & 0.3869                      & 0.5839  & 0.4369  & 0.5345  & 0.5515                       \\
(20, 40{]} - MHA \& FFN & 0.7381                           & 0.4926                       & 0.3604                      & 0.5561  & 0.4108  & 0.5109  & 0.5303                       \\
(5, 25{]} - MHA \& FFN  & 0.7479                           & 0.5173                       & 0.3881                      & 0.5821  & 0.4364  & 0.5333  & 0.5510                       \\
                        &                                  &                              &                             &         &         &         &                              \\
(10, 20{]} - MHA \& FFN & 0.7471                           & 0.5106                       & 0.3761                      & 0.5769  & 0.4286  & 0.5264  & 0.5446                       \\
(20, 30{]} - MHA \& FFN & 0.7379                           & 0.4885                       & 0.3542                      & 0.5542  & 0.4013  & 0.5102  & 0.5268                       \\
\hline
\end{tabular}
\caption{The detailed speciality performance of CAG task under the 13B model in different exploring layer ranges and modules.}
\end{table}

\begin{table}[!h]
\centering
\setlength\tabcolsep{2pt}
\fontsize{10}{14}\selectfont 
\begin{tabular}{lccc|cccc|c}
\hline
                        & {\color[HTML]{333333} Instruct.} & {\color[HTML]{333333} Gen-Kn.} & {\color[HTML]{333333} Gen-Rs.} & Logiqa1 & Logiqa2 & OCNLI  & {\color[HTML]{333333} Zh\_Winograd} & {\color[HTML]{333333} Vers.} \\ \hline
(0, 20{]} - MHA \& FFN  & 0.5543                           & 0.3621                         & 0.4556                         & 0.3062  & 0.3439  & 0.4709 & 0.7014                              & 0.4573                       \\
(10, 30{]} - MHA \& FFN & 0.5641                           & 0.3963                         & 0.4956                         & 0.3424  & 0.3988  & 0.5253 & 0.7159                              & 0.4852                       \\
(20, 40{]} - MHA \& FFN & 0.5852                           & 0.4041                         & 0.5062                         & 0.3632  & 0.4168  & 0.5223 & 0.7224                              & 0.4984                       \\
(5, 25{]} - MHA \& FFN  & 0.5574                           & 0.3857                         & 0.4702                         & 0.3213  & 0.3867  & 0.4676 & 0.7053                              & 0.4711                       \\
                        &                                  &                                &                                &         &         &        &                                     &                              \\
(10, 20{]} - MHA \& FFN & 0.5905                           & 0.3984                         & 0.4987                         & 0.3387  & 0.4101  & 0.5306 & 0.7155                              & 0.4959                       \\
(20, 30{]} - MHA \& FFN & 0.6037                           & 0.4051                         & 0.5071                         & 0.3431  & 0.4203    & 0.5486 & 0.7168                              & 0.5052                       \\
\hline
\end{tabular}
\caption{The detailed versatility performance of CAG task under the 13B model in different exploring layer ranges and modules.}
\end{table}

\onecolumn\newpage
\subsection{Optional Solution}
\label{sec:optional_solution}
% \paragraph{Optional Solution}
% \label{sec:optional_solution}
\begin{table}[!h]
\centering
\normalsize
\begin{tabular}{lccccccccc}
\hline
                                  & \multicolumn{4}{c}{{\color[HTML]{333333} 7B}}                                                                          &           & \multicolumn{4}{c}{{\color[HTML]{333333} 13B}}                                                                         \\ \cline{2-5} \cline{7-10} 
                                  & {\color[HTML]{333333} Finance} & {\color[HTML]{333333} Law} & {\color[HTML]{333333} Math} & {\color[HTML]{333333} CAG} &           & {\color[HTML]{333333} Finance} & {\color[HTML]{333333} Law} & {\color[HTML]{333333} Math} & {\color[HTML]{333333} CAG} \\ \hline
{\color[HTML]{333333} MHA \& FFN} & \textbf{0.8241}                & \textbf{0.8382}            & \textbf{0.4893}             & \textbf{0.9185}            & \textbf{} & \textbf{0.8901}                & \textbf{0.9039}            & \textbf{0.5632}             & \textbf{0.9931}            \\
{\color[HTML]{333333} FFN}        & 0.8204                         & 0.8342                     & 0.4871                      & 0.9156                     &           & 0.8842                         & 0.8994                     & 0.5630                      & 0.9904                     \\
{\color[HTML]{333333} MHA}        & 0.8066                         & 0.8207                     & 0.4650                      & 0.9033                     &           & 0.8667                         & 0.8807                     & 0.5190                      & 0.9771                     \\
{\color[HTML]{333333} down\_proj} & 0.8138                         & 0.8258                     & 0.4778                      & 0.9082                     &           & 0.8743                         & 0.8882                     & 0.5299                      & 0.9833                     \\
\hline
\end{tabular}
\vspace{-0.1cm}
\caption{The \emph{Uni. w/o instruct} scores in various trainable modules within the $(N \times 25\%, N \times 50\%]$ layer range of the 7B and 13B models.}
\vspace{-0.4cm}
\label{tab:optional_solution}
\end{table}
In our context, instruction following is a crucial aspect of LLM's versatility. 
However, in scenarios where users can provide clear and concise instructions, the model's precision in this aspect may be less critical. 
We conduct additional analysis on the results of LLM's versatility without instruction following (\emph{Vers. w/o instruct}), considering different trainable modules in the layer range $(N\times 25\%, N \times 50\%]$ under 7B and 13B models. The derived Uni. score without instruction following (\emph{Uni. w/o instruct}) shows that ``$(N \times 25\%, N \times 50\%]$ - MHA \& FFN'' generally outperforms ``$(N \times 25\%, N \times 50\%]$ - FFN''. This suggests that ``$(N \times 25\%, N \times 50\%]$ - MHA \& FFN'' could be a viable alternative when users value instruction following aspect less.

\subsection{Detailed Results of \emph{CoFiTune} and Baselines under BLOOMZ Backbone}
\label{sec:appendix_main_results_bloomz}
\begin{table}[!h]
\centering
\setlength\tabcolsep{4.5pt}
\fontsize{10}{14}\selectfont 
\begin{tabular}{lccc|ccc|c}
\hline
                    & {\color[HTML]{333333} BERTScore} & {\color[HTML]{333333} Rouge}  & {\color[HTML]{333333} BLEU}   & Rouge-1                       & Rouge-2                       & Rouge-L                       & {\color[HTML]{333333} Spec.}  \\ \hline
\textit{ZeroShot}   & {\color[HTML]{333333} 0.6814}    & {\color[HTML]{333333} 0.2869} & {\color[HTML]{333333} 0.1042} & {\color[HTML]{333333} 0.4177} & {\color[HTML]{333333} 0.1639} & {\color[HTML]{333333} 0.2792} & 0.3575                        \\
\textit{Full SFT}   & {\color[HTML]{333333} 0.7106}    & {\color[HTML]{333333} 0.3581} & {\color[HTML]{333333} 0.1893} & {\color[HTML]{333333} 0.4985} & {\color[HTML]{333333} 0.2292} & {\color[HTML]{333333} 0.3468} & {\color[HTML]{333333} 0.4193} \\
\textit{LoRA}       & {\color[HTML]{333333} 0.6932}    & {\color[HTML]{333333} 0.3167} & {\color[HTML]{333333} 0.1406} & {\color[HTML]{333333} 0.4536} & {\color[HTML]{333333} 0.1944} & {\color[HTML]{333333} 0.3023} & 0.3835                        \\
\textit{Wise-FT}    & {\color[HTML]{333333} 0.7015}    & 0.3296                        & {\color[HTML]{333333} 0.1614} & {\color[HTML]{333333} 0.4651} & {\color[HTML]{333333} 0.2047} & {\color[HTML]{333333} 0.3192} & 0.3975                        \\
\textit{V-SoftMask} & {\color[HTML]{333333} 0.7113}    & 0.3594                        & {\color[HTML]{333333} 0.1885} & {\color[HTML]{333333} 0.5002} & {\color[HTML]{333333} 0.2295} & {\color[HTML]{333333} 0.3485} & 0.4197                        \\
\textit{L1}         & {\color[HTML]{333333} 0.7007}    & 0.3289                        & {\color[HTML]{333333} 0.1572} & {\color[HTML]{333333} 0.4685} & {\color[HTML]{333333} 0.2036} & {\color[HTML]{333333} 0.3161} & 0.3956                        \\
\hline
\end{tabular}
\caption{The detailed speciality performance of Finance task under the 7B model in different methods.}
\end{table}

\begin{table}[!h]
\centering
\setlength\tabcolsep{2pt}
\fontsize{10}{14}\selectfont 
\begin{tabular}{lccc|cccc|c}
\hline
                    & {\color[HTML]{333333} Instruct.} & {\color[HTML]{333333} Gen-Kn.} & {\color[HTML]{333333} Gen-Rs.} & Logiqa1                       & Logiqa2                       & OCNLI                         & {\color[HTML]{333333} Zh\_Winograd} & {\color[HTML]{333333} Vers.}  \\ \hline
\textit{ZeroShot}   & {\color[HTML]{333333} 0.6109}    & {\color[HTML]{333333} 0.3457}  & {\color[HTML]{333333} 0.4232}  & {\color[HTML]{333333} 0.2831} & {\color[HTML]{333333} 0.2947} & {\color[HTML]{333333} 0.4168} & {\color[HTML]{333333} 0.6984}       & 0.4599                        \\
\textit{Full SFT}   & {\color[HTML]{333333} 0.4471}    & {\color[HTML]{333333} 0.2943}  & 0.3787                         & {\color[HTML]{333333} 0.2518} & {\color[HTML]{333333} 0.2563} & {\color[HTML]{333333} 0.3742} & {\color[HTML]{333333} 0.6325}       & 0.3733                        \\
\textit{LoRA}       & {\color[HTML]{333333} 0.5968}    & {\color[HTML]{333333} 0.2965}  & {\color[HTML]{333333} 0.3781}  & {\color[HTML]{333333} 0.2573} & {\color[HTML]{333333} 0.2604} & {\color[HTML]{333333} 0.3216} & {\color[HTML]{333333} 0.6728}       & 0.4238                        \\
\textit{Wise-FT}    & {\color[HTML]{333333} 0.5884}    & {\color[HTML]{333333} 0.3216}  & {\color[HTML]{333333} 0.4028}  & {\color[HTML]{333333} 0.2669} & {\color[HTML]{333333} 0.2771} & {\color[HTML]{333333} 0.3865} & {\color[HTML]{333333} 0.6806}       & 0.4376                        \\
\textit{V-SoftMask} & {\color[HTML]{333333} 0.4508}    & {\color[HTML]{333333} 0.2981}  & 0.3839                         & {\color[HTML]{333333} 0.2562} & {\color[HTML]{333333} 0.2675} & {\color[HTML]{333333} 0.3804} & {\color[HTML]{333333} 0.6313}       & 0.3776                        \\
\hline
\end{tabular}
\caption{The detailed versatility performance of Finance task under the 7B model in different methods.}
\end{table}

\onecolumn\newpage
\subsection{Detailed Results of \emph{CoFiTune} and Baselines under Chinese-Alpaca-Pro Backbone}
\label{sec:appendix_main_results}
\subsubsection{7B Results}
% \onecolumn
\begin{table}[!h]
\centering
\setlength\tabcolsep{4.5pt}
\fontsize{10}{14}\selectfont 
\begin{tabular}{lccc|ccc|c}
\hline
                    & {\color[HTML]{333333} BERTScore} & {\color[HTML]{333333} Rouge} & {\color[HTML]{333333} BLEU} & Rouge-1 & Rouge-2 & Rouge-L & {\color[HTML]{333333} Spec.} \\ \hline
\textit{ZeroShot}   & 0.6912                           & 0.3038                       & 0.1189                      & 0.4384  & 0.1766  & 0.2963  & 0.3713                       \\
\textit{Full SFT}   & 0.7376                           & 0.3973                       & 0.2163                      & 0.5440   & 0.2752  & 0.3728  & 0.4504                       \\
\textit{LoRA}       & 0.7159                           & 0.3487                       & 0.1527                      & 0.5047  & 0.2233  & 0.3182  & 0.4058                       \\
\textit{Wise-FT}    & 0.7208                           & 0.3597                       & 0.1764                      & 0.5134  & 0.2331  & 0.3327  & 0.4189                       \\
\textit{V-SoftMask} & 0.7372                           & 0.3980                       & 0.2175                      & 0.5456  & 0.2764  & 0.3720   & 0.4509                       \\
\textit{L1}         & 0.7181                           & 0.3551                       & 0.1737                      & 0.5091  & 0.2282  & 0.3277  & 0.4156                       \\
\textit{L2}         & 0.7254                           & 0.3649                       & 0.1783                      & 0.5192  & 0.2388  & 0.3368  & 0.4229                       \\
\textit{CoFiTune}   & 0.7222                           & 0.3645                       & 0.1788                      & 0.5197  & 0.2374  & 0.3365  & 0.4217                       \\ \hline
\end{tabular}
\caption{The detailed speciality performance of Finance task under the 7B model across different methods.}
\end{table}

\begin{table}[!h]
\centering
\setlength\tabcolsep{2pt}
\fontsize{10}{14}\selectfont 
\begin{tabular}{lccc|cccc|c}
\hline
                    & {\color[HTML]{333333} Instruct.} & {\color[HTML]{333333} Gen-Kn.} & {\color[HTML]{333333} Gen-Rs.} & Logiqa1 & Logiqa2 & OCNLI  & {\color[HTML]{333333} Zh\_Winograd} & {\color[HTML]{333333} Vers.} \\ \hline
\textit{ZeroShot}   & 0.6237                           & 0.3671                         & 0.4639                         & 0.3101  & 0.3379  & 0.4622 & 0.7455                              & 0.4849                       \\
\textit{Full SFT}   & 0.4608                           & 0.3158                         & 0.4188                         & 0.2666  & 0.2969  & 0.4441 & 0.6673                              & 0.3984                       \\
\textit{LoRA}       & 0.6074                           & 0.3145                         & 0.4211                         & 0.2738  & 0.2832  & 0.4039 & 0.7234                              & 0.4476                       \\
\textit{Wise-FT}    & 0.5979                           & 0.3436                         & 0.4424                         & 0.2828  & 0.3014  & 0.4542 & 0.7314                              & 0.4613                       \\
\textit{V-SoftMask} & 0.4643                           & 0.3204                         & 0.4237                         & 0.2708  & 0.3015  & 0.4467 & 0.6759                              & 0.4028                       \\
\textit{L1}         & 0.5767                           & 0.3213                         & 0.4293                         & 0.2733  & 0.2916  & 0.4376 & 0.7148                              & 0.4424                       \\
\textit{L2}         & 0.5256                           & 0.3179                         & 0.4304                         & 0.2762  & 0.2927  & 0.4262 & 0.7261                              & 0.4246                       \\
\hline
\end{tabular}
\caption{The detailed versatility performance of Finance task under the 7B model across different methods.}
\end{table}

\begin{table}[!h]
\centering
\setlength\tabcolsep{4.5pt}
\fontsize{10}{14}\selectfont 
\begin{tabular}{lccc|ccc|c}
\hline
                    & {\color[HTML]{333333} BERTScore} & {\color[HTML]{333333} Rouge} & {\color[HTML]{333333} BLEU} & Rouge-1 & Rouge-2 & Rouge-L & {\color[HTML]{333333} Spec.} \\ \hline
\textit{ZeroShot}   & 0.6934                           & 0.3083                       & 0.1247                      & 0.4375  & 0.1818  & 0.3057  & 0.3755                       \\
\textit{Full SFT}   & 0.7407                           & 0.4039                       & 0.2282                      & 0.5329  & 0.2754  & 0.4036  & 0.4576                       \\
\textit{LoRA}       & 0.7195                           & 0.3594                       & 0.1725                      & 0.4926  & 0.2335  & 0.3521  & 0.4171                       \\
\textit{Wise-FT}    & 0.7266                           & 0.3751                       & 0.1944                      & 0.5093  & 0.2477  & 0.3684  & 0.4320                       \\
\textit{V-SoftMask} & 0.7402                           & 0.4046                       & 0.2289                      & 0.5365  & 0.2769  & 0.4005  & 0.4579                       \\
\textit{L1}         & 0.7243                           & 0.3695                       & 0.1872                      & 0.5046  & 0.2443  & 0.3595  & 0.4270                       \\
\textit{L2}         & 0.7315                           & 0.3756                       & 0.1953                      & 0.5103  & 0.2467  & 0.3699  & 0.4341                       \\
\textit{CoFiTune}   & 0.7271                           & 0.3772                       & 0.1942                      & 0.5119  & 0.2491  & 0.3706  & 0.4328                       \\ \hline
\end{tabular}
\caption{The detailed speciality performance of Law task under the 7B model across different methods.}
\end{table}

\begin{table}[!h]
\centering
\setlength\tabcolsep{2pt}
\fontsize{10}{14}\selectfont 
\begin{tabular}{lccc|cccc|c}
\hline
                    & {\color[HTML]{333333} Instruct.} & {\color[HTML]{333333} Gen-Kn.} & {\color[HTML]{333333} Gen-Rs.} & Logiqa1 & Logiqa2 & OCNLI  & {\color[HTML]{333333} Zh\_Winograd} & {\color[HTML]{333333} Vers.} \\ \hline
\textit{ZeroShot}   & 0.6237                           & 0.3671                         & 0.4639                         & 0.3101  & 0.3379  & 0.4622 & 0.7455                              & 0.4849                       \\
\textit{Full SFT}   & 0.4280                           & 0.3162                         & 0.4212                         & 0.3039  & 0.3046  & 0.3941  & 0.6824                              & 0.3884                       \\
\textit{LoRA}       & 0.6151                           & 0.3233                         & 0.4335                         & 0.3178  & 0.3109  & 0.3578 & 0.7475                              & 0.4573                       \\
\textit{Wise-FT}    & 0.5925                           & 0.3335                         & 0.4475                         & 0.3006  & 0.3222  & 0.4257 & 0.7415                              & 0.4578                       \\
\textit{V-SoftMask} & 0.4321                           & 0.3209                         & 0.4277                         & 0.3101  & 0.3142  & 0.3982 & 0.6885                              & 0.3936                       \\
\textit{L1}         & 0.5657                           & 0.3298                         & 0.4378                         & 0.3177  & 0.3193  & 0.4133 & 0.7011                             & 0.4444                       \\
\textit{L2}         & 0.5243                           & 0.3246                         & 0.4335                         & 0.3058  & 0.3269  & 0.4065 & 0.6946                             & 0.4275                       \\
\hline
\end{tabular}
\caption{The detailed versatility performance of Law task under the 7B model across different methods.}
\end{table}

\begin{table}[!h]
\centering
\begin{tabular}{l|c}
\hline
                    & {\color[HTML]{333333} Accuracy} \\ \hline
\textit{ZeroShot}   & 0.048                           \\
\textit{Full SFT}   & 0.100                           \\
\textit{LoRA}       & 0.056                           \\
\textit{Wise-FT}    & 0.064                           \\
\textit{V-SoftMask} & 0.096                           \\
\textit{L1}         & 0.068                           \\
\textit{L2}         & 0.076                           \\
\textit{CoFiTune}   & 0.080                           \\ \hline
\end{tabular}
\caption{The detailed speciality performance of Math task under the 7B model across different methods.}
\end{table}

\begin{table}[!h]
\centering
\setlength\tabcolsep{2pt}
\fontsize{10}{14}\selectfont 
\begin{tabular}{lccc|cccc|c}
\hline
                    & {\color[HTML]{333333} Instruct.} & {\color[HTML]{333333} Gen-Kn.} & {\color[HTML]{333333} Gen-Rs.} & Logiqa1 & Logiqa2 & OCNLI  & {\color[HTML]{333333} Zh\_Winograd} & {\color[HTML]{333333} Vers.} \\ \hline
\textit{ZeroShot}   & 0.6237                           & 0.3671                         & 0.4639                         & 0.3101  & 0.3379  & 0.4622 & 0.7455                              & 0.4849                       \\
\textit{Full SFT}   & 0.4874                           & 0.3227                         & 0.4450                         & 0.3194  & 0.3071  & 0.4421 & 0.7114                              & 0.4184                       \\
\textit{LoRA}       & 0.6077                           & 0.3133                         & 0.4326                         & 0.2899  & 0.2782  & 0.4311 & 0.7315                              & 0.4512                       \\
\textit{Wise-FT}    & 0.6156                           & 0.3515                         & 0.4515                         & 0.3116  & 0.3486  & 0.3965 & 0.7495                              & 0.4729                       \\
\textit{V-SoftMask} & 0.4921                           & 0.3267                         & 0.4496                         & 0.3218  & 0.3105  & 0.4453 & 0.7213                              & 0.4226                       \\
\textit{L1}         & 0.5888                           & 0.3273                         & 0.4464                         & 0.3153  & 0.3067  & 0.4351 & 0.7285                              & 0.4542                       \\
\textit{L2}         & 0.5464                           & 0.3215                         & 0.4447                         & 0.3083  & 0.3113  & 0.4291 & 0.7301                              & 0.4375                       \\
\hline
\end{tabular}
\caption{The detailed versatility performance of Math task under the 7B model across different methods.}
\end{table}

\begin{table}[!h]
\centering
\setlength\tabcolsep{4.5pt}
\fontsize{10}{14}\selectfont 
\begin{tabular}{lccc|ccc|c}
\hline
                    & {\color[HTML]{333333} BERTScore} & {\color[HTML]{333333} Rouge} & {\color[HTML]{333333} BLEU} & Rouge-1 & Rouge-2 & Rouge-L & {\color[HTML]{333333} Spec.} \\ \hline
\textit{ZeroShot}   & 0.7012                           & 0.4056                       & 0.2712                      & 0.4559  & 0.3241  & 0.4369  & 0.4593                       \\
\textit{Full SFT}   & 0.7486                           & 0.5005                       & 0.3703                      & 0.5632  & 0.4231  & 0.5153  & 0.5398                       \\
\textit{LoRA}       & 0.7314                           & 0.4632                       & 0.3308                      & 0.5334  & 0.3796  & 0.4765  & 0.5085                       \\
\textit{Wise-FT}    & 0.7323                           & 0.4664                       & 0.3355                      & 0.5367  & 0.3833  & 0.4791  & 0.5114                       \\
\textit{V-SoftMask} & 0.7483                           & 0.5004                       & 0.3714                      & 0.5618  & 0.4232  & 0.5162  & 0.5400                       \\
\textit{L1}         & 0.7324                           & 0.4660                       & 0.3276                      & 0.5389  & 0.3824  & 0.4767  & 0.5087                       \\
\textit{L2}         & 0.7355                           & 0.4713                       & 0.3407                      & 0.5411  & 0.3913  & 0.4815  & 0.5158                       \\
\textit{CoFiTune}   & 0.7356                           & 0.4720                       & 0.3422                      & 0.5414  & 0.3905  & 0.4840   & 0.5166                       \\ \hline
\end{tabular}
\caption{The detailed speciality performance of CAG task under the 7B model across different methods.}
\end{table}

\begin{table}[!h]
\centering
\setlength\tabcolsep{2pt}
\fontsize{10}{14}\selectfont 
\begin{tabular}{lccc|cccc|c}
\hline
                    & {\color[HTML]{333333} Instruct.} & {\color[HTML]{333333} Gen-Kn.} & {\color[HTML]{333333} Gen-Rs.} & Logiqa1 & Logiqa2 & OCNLI  & {\color[HTML]{333333} Zh\_Winograd} & {\color[HTML]{333333} Vers.} \\ \hline
\textit{ZeroShot}   & 0.6237                           & 0.3671                         & 0.4639                         & 0.3101  & 0.3379  & 0.4622 & 0.7455                              & 0.4849                       \\
\textit{Full SFT}   & 0.4517                           & 0.3111                         & 0.4123                         & 0.2822  & 0.3021  & 0.4035 & 0.6613                              & 0.3917                       \\
\textit{LoRA}       & 0.6178                           & 0.3215                         & 0.4319                         & 0.2945  & 0.3058  & 0.3819 & 0.7455                              & 0.4571                       \\
\textit{Wise-FT}    & 0.6009                           & 0.3423                         & 0.4460                         & 0.2930  & 0.3336  & 0.4321 & 0.7255                              & 0.4631                       \\
\textit{V-SoftMask} & 0.4549                           & 0.3165                         & 0.4172                         & 0.2853  & 0.3102  & 0.4040  & 0.6693                              & 0.3962                       \\
\textit{L1}         & 0.5746                           & 0.3283                         & 0.4334                         & 0.3012  & 0.3153  & 0.4137 & 0.7033                              & 0.4454                       \\
\textit{L2}         & 0.5205                           & 0.3252                         & 0.4307                         & 0.3034  & 0.3128  & 0.4173 & 0.6894                              & 0.4255                       \\
\hline
\end{tabular}
\caption{The detailed versatility performance of CAG task under the 7B model across different methods.}
\end{table}

% \paragraph{Knowledge Intensive Tasks}

% 先onecolumn 再 new page
\onecolumn
\newpage
\subsubsection{13B Results}
% \onecolumn
\begin{table}[!h]
\centering
\setlength\tabcolsep{4.5pt}
\fontsize{10}{14}\selectfont 
\begin{tabular}{lccc|ccc|c}
\hline
                    & {\color[HTML]{333333} BERTScore} & {\color[HTML]{333333} Rouge} & {\color[HTML]{333333} BLEU} & Rouge-1 & Rouge-2 & Rouge-L & {\color[HTML]{333333} Spec.} \\ \hline
\textit{ZeroShot}   & 0.6934                           & 0.3124                       & 0.1239                      & 0.4457  & 0.1881  & 0.3035  & 0.3766                       \\
\textit{Full SFT}   & 0.7475                           & 0.4192                       & 0.2415                      & 0.5614  & 0.2997  & 0.3964  & 0.4761                       \\
\textit{LoRA}       & 0.7217                           & 0.3628                       & 0.1692                      & 0.5209  & 0.2365  & 0.3311  & 0.4179                       \\
\textit{Wise-FT}    & 0.7258                           & 0.3729                       & 0.1936                      & 0.5226  & 0.2478  & 0.3485  & 0.4308                       \\
\textit{V-SoftMask} & 0.7477                           & 0.4172                       & 0.2408                      & 0.5595  & 0.2978  & 0.3943  & 0.4752                       \\
\textit{L1}         & 0.7265                           & 0.3741                       & 0.1857                      & 0.5242  & 0.2486  & 0.3494  & 0.4287                       \\
\textit{L2}         & 0.7302                           & 0.3817                       & 0.1985                      & 0.5312  & 0.2574  & 0.3566  & 0.4368                       \\
\textit{CoFiTune}   & 0.7295                           & 0.3799                       & 0.1958                      & 0.5294  & 0.2561  & 0.3542  & 0.4351                       \\ \hline
\end{tabular}
\caption{The detailed speciality performance of Finance task under the 13B model in different methods.}
\end{table}

\begin{table}[!h]
\centering
\setlength\tabcolsep{2pt}
\fontsize{10}{14}\selectfont 
\begin{tabular}{lccc|cccc|c}
\hline
                    & {\color[HTML]{333333} Instruct.} & {\color[HTML]{333333} Gen-Kn.} & {\color[HTML]{333333} Gen-Rs.} & Logiqa1 & Logiqa2 & OCNLI  & {\color[HTML]{333333} Zh\_Winograd} & {\color[HTML]{333333} Vers.} \\ \hline
\textit{ZeroShot}   & 0.6320                           & 0.4057                         & 0.5224                         & 0.3752  & 0.4141  & 0.5448 & 0.7555                              & 0.5201                       \\
\textit{Full SFT}   & 0.4042                           & 0.3555                         & 0.4431                         & 0.3333  & 0.3228  & 0.4652 & 0.6513                              & 0.4009                       \\
\textit{LoRA}       & 0.5986                           & 0.3721                         & 0.4704                         & 0.3147  & 0.3486  & 0.4867 & 0.7314                              & 0.4804                       \\
\textit{Wise-FT}    & 0.5777                           & 0.3908                         & 0.4873                         & 0.3240   & 0.3801  & 0.5318 & 0.7136                              & 0.4853                       \\
\textit{V-SoftMask} & 0.4082                           & 0.3593                         & 0.4484                         & 0.3311  & 0.3279  & 0.4742 & 0.6603                              & 0.4055                       \\
\textit{L1}         & 0.5577                           & 0.3779                         & 0.4742                         & 0.3214  & 0.3564  & 0.4902 & 0.7288                              & 0.4701                       \\
\textit{L2}         & 0.4786                           & 0.3723                         & 0.4702                         & 0.3135  & 0.3464  & 0.5015 & 0.7192                              & 0.4404                       \\
\hline
\end{tabular}
\caption{The detailed versatility performance of Finance task under the 13B model in different methods.}
\end{table}

\begin{table}[!h]
\centering
\setlength\tabcolsep{4.5pt}
\fontsize{10}{14}\selectfont 
%
\caption{The detailed speciality performance of Law task under the 13B model in different methods.}
\end{table}

\begin{table}[!h]
\centering
\setlength\tabcolsep{2pt}
\fontsize{10}{14}\selectfont 
%
\caption{The detailed versatility performance of Law task under the 13B model in different methods.}
\end{table}

\begin{table}[!h]
\centering
\setlength\tabcolsep{4.5pt}
\fontsize{10}{14}\selectfont 
\begin{tabular}{l|c}
\hline
                    & {\color[HTML]{333333} Accuracy} \\ \hline
\textit{ZeroShot}   & 0.076                           \\
\textit{Full SFT}   & 0.140                           \\
\textit{LoRA}       & 0.084                           \\
\textit{Wise-FT}    & 0.100                           \\
\textit{V-SoftMask} & 0.140                           \\
\textit{L1}         & 0.092                           \\
\textit{L2}         & 0.108                           \\
\textit{CoFiTune}   & 0.112                           \\ \hline
\end{tabular}
\caption{The detailed speciality performance of Math task under the 13B model in different methods.}
\end{table}

\begin{table}[!h]
\centering
\setlength\tabcolsep{2pt}
\fontsize{10}{14}\selectfont 
%
\caption{The detailed versatility performance of Math task under the 13B model in different methods.}
\end{table}

\begin{table}[!h]
\centering
\setlength\tabcolsep{4.5pt}
\fontsize{10}{14}\selectfont 
%
\caption{The detailed speciality performance of CAG task under the 13B model in different methods.}
\end{table}

\begin{table}[!h]
\centering
\setlength\tabcolsep{2pt}
\fontsize{10}{14}\selectfont 
%
\caption{The detailed versatility performance of CAG task under the 13B model in different methods.}
\end{table}

\onecolumn
\newpage
\subsubsection{33B Results}
\label{sec:appendix_main_results_33b}
% \onecolumn
\begin{table}[!h]
\centering
\setlength\tabcolsep{4.5pt}
\fontsize{10}{14}\selectfont 
%
\caption{The detailed speciality performance of Finance task under the 33B model in different methods.}
\end{table}

\begin{table}[!h]
\centering
\setlength\tabcolsep{2pt}
\fontsize{10}{14}\selectfont 
%
\caption{The detailed versatility performance of Finance task under the 33B model in different methods.}
\end{table}

\begin{table}[!h]
\centering
\setlength\tabcolsep{4.5pt}
\fontsize{10}{14}\selectfont 
%
\caption{The detailed speciality performance of Law task under the 33B model in different methods.}
\end{table}

\begin{table}[!h]
\centering
\setlength\tabcolsep{2pt}
\fontsize{10}{14}\selectfont 
%
\caption{The detailed versatility performance of Law task under the 33B model in different methods.}
\end{table}

\begin{table}[!h]
\centering
\setlength\tabcolsep{4.5pt}
\fontsize{10}{14}\selectfont 
\begin{tabular}{l|c}
\hline
                    & {\color[HTML]{333333} Accuracy} \\ \hline
\textit{ZeroShot}   & 0.120                           \\
\textit{Full SFT}   & 0.192                           \\
\textit{LoRA}       & 0.128                           \\
\textit{Wise-FT}    & 0.144                           \\
\textit{V-SoftMask} & 0.188                           \\
\textit{L1}         & 0.140                           \\
\textit{L2}         & 0.156                           \\
\textit{CoFiTune}   & 0.152                           \\ \hline
\end{tabular}
\caption{The detailed speciality performance of Math task under the 33B model in different methods.}
\end{table}

\begin{table}[!h]
\centering
\setlength\tabcolsep{2pt}
\fontsize{10}{14}\selectfont 
%
\caption{The detailed versatility performance of Math task under the 33B model in different methods.}
\end{table}

\begin{table}[!h]
\centering
\setlength\tabcolsep{4.5pt}
\fontsize{10}{14}\selectfont 
%
\caption{The detailed speciality performance of CAG task under the 33B model in different methods.}
\end{table}

\begin{table}[!h]
\centering
\setlength\tabcolsep{2pt}
\fontsize{10}{14}\selectfont 
%
\caption{The detailed versatility performance of CAG task under the 33B model in different methods.}
\end{table}

% \subsection{}
% \label{sec:appendix_layer_param}
% \input{appendix_layer_param}

\onecolumn\newpage
\subsection{Module Importance for Speciality}
\label{sec:appendix_module_impt}
\begin{figure}[!thbp]
    \centerline{\includegraphics[scale=0.6]{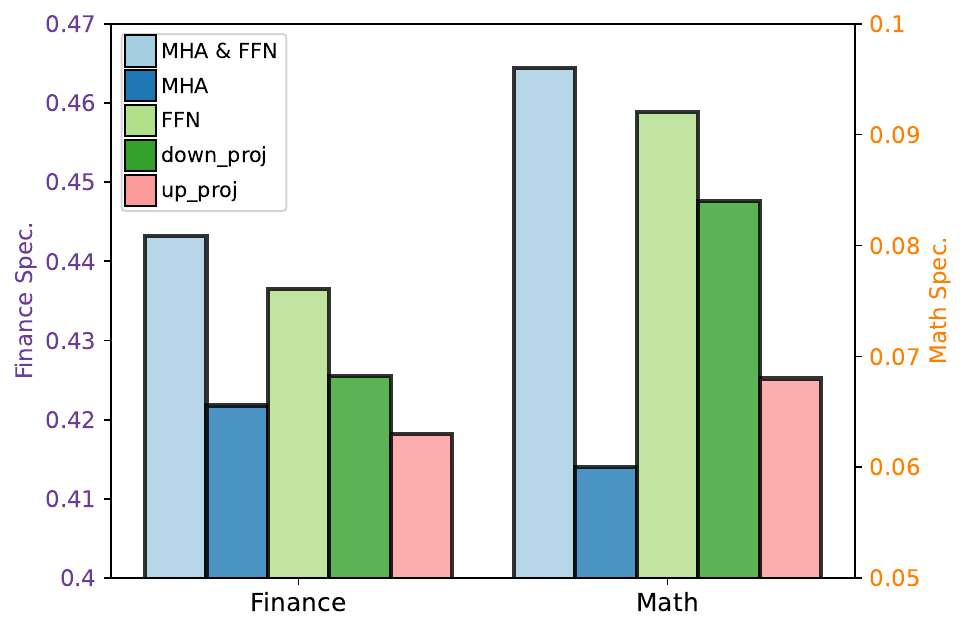}}
    \caption{The Spec. score of different modules trained in all layers for Finance and Math tasks under the 7B model.}
    \label{fig:all_layer_impt_7b}
    \vspace{-0.4cm}
\end{figure}

% \onecolumn\newpage
\subsection{Exploring CF in LLM's Versatility}
\label{sec:appendix_speculate_harm}
\begin{figure}[!thbp]
    \centerline{\includegraphics[scale=0.6]{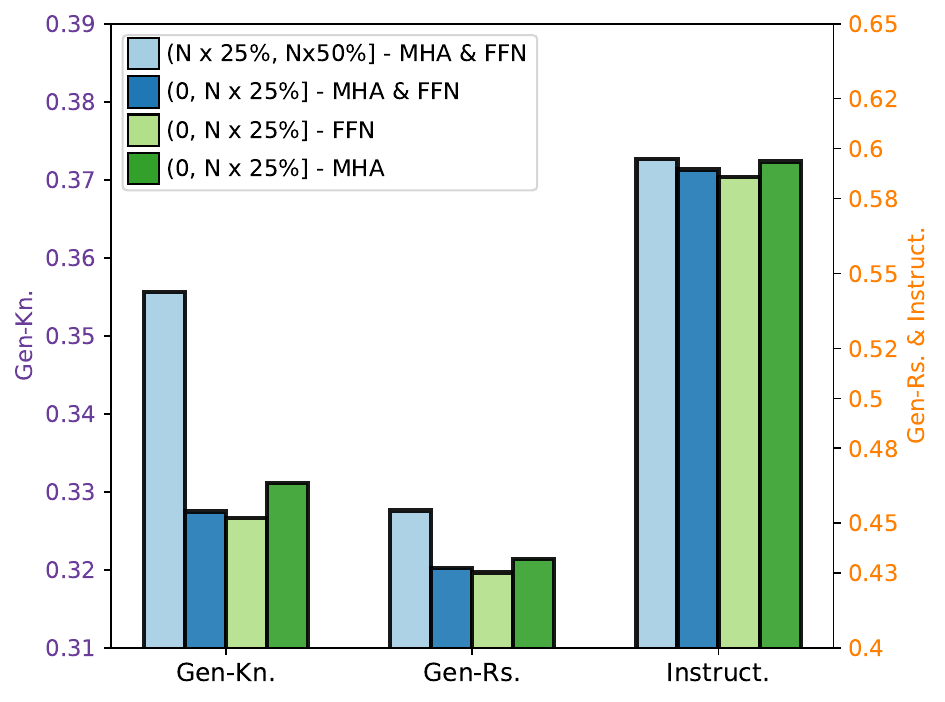}}
    \vspace{-0.2cm}
    \caption{The Gen-Kn., Gen-Rs., and Instruct. scores in Math task under the 7B model.}
    \label{fig:bottom_layer_general_math_7b}
    \vspace{-0.3cm}
\end{figure}

\begin{figure}[!thbp]
    \centerline{\includegraphics[scale=0.6]{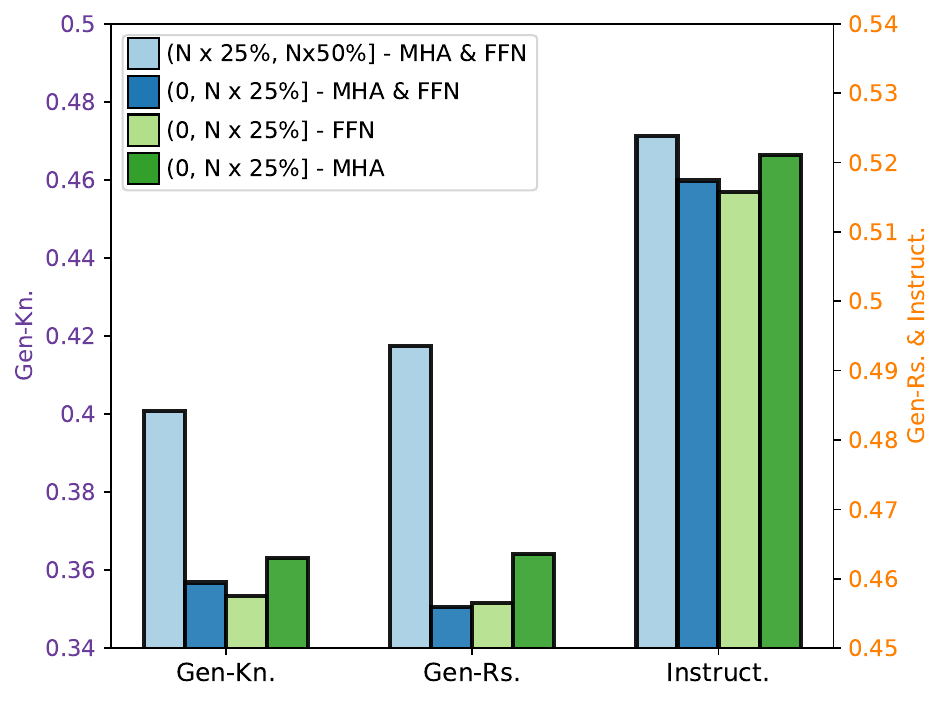}}
    \vspace{-0.2cm}
    \caption{The Gen-Kn., Gen-Rs., and Instruct. scores in Finance task under the 13B model.}
    \label{fig:bottom_layer_general_fiance_13b}
    \vspace{-0.3cm}
\end{figure}

\begin{figure}[!thbp]
    \centerline{\includegraphics[scale=0.6]{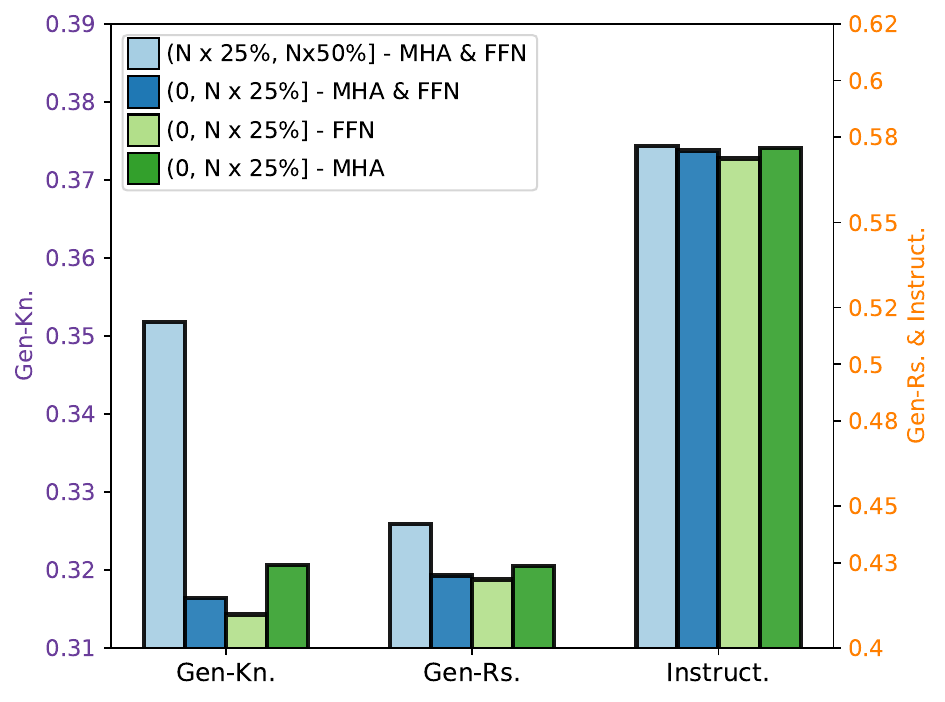}}
    \vspace{-0.2cm}
    \caption{The Gen-Kn., Gen-Rs., and Instruct. scores in Finance task under the 7B model.}
    \label{fig:bottom_layer_general_fiance_7b}
    \vspace{-0.3cm}
\end{figure}

\onecolumn\newpage
\subsection{Details of Our Evaluation Setting}
\label{sec:appendix_evaluation_validation}
\subsubsection{A New Strategy for Improving the Reliability of Evaluation Results}
\label{sec:appendix_new_evaluation_strategy}
\begin{table*}[h]
\centering
\begin{tabular}{lccccccc}
\toprule[1.5pt]
{\color[HTML]{333333} }                    & \multicolumn{3}{c}{{\color[HTML]{333333} Test set}}                                              &  & \multicolumn{3}{c}{{\color[HTML]{333333} Sample-Then-Paraphrase (Ours)}}                         \\ \addlinespace[2pt] \cline{2-4} \cline{6-8} \addlinespace[2pt]
                                           & {\color[HTML]{333333} BERTScore} & {\color[HTML]{333333} Rouge}  & {\color[HTML]{333333} BLEU}   &  & {\color[HTML]{333333} BERTScore} & {\color[HTML]{333333} Rouge}  & {\color[HTML]{333333} BLEU}   \\ \hline \addlinespace[2pt]
{\color[HTML]{333333} \textit{Full SFT}}   & {\color[HTML]{333333} 0.6325}    & {\color[HTML]{333333} 0.2218} & {\color[HTML]{333333} 0.0992} &  & {\color[HTML]{333333} 0.7376}    & {\color[HTML]{333333} 0.3973} & {\color[HTML]{333333} 0.2163} \\
{\color[HTML]{333333} \textit{LoRA}}       & {\color[HTML]{333333} 0.6314}    & {\color[HTML]{333333} 0.2239} & {\color[HTML]{333333} 0.1026} &  & {\color[HTML]{333333} 0.7159}    & {\color[HTML]{333333} 0.3487} & {\color[HTML]{333333} 0.1527} \\
{\color[HTML]{333333} \textit{Wise-FT}}    & {\color[HTML]{333333} 0.6293}    & {\color[HTML]{333333} 0.2195} & {\color[HTML]{333333} 0.1003} &  & {\color[HTML]{333333} 0.7208}    & {\color[HTML]{333333} 0.3597} & {\color[HTML]{333333} 0.1764} \\
{\color[HTML]{333333} \textit{V-SoftMask}} & {\color[HTML]{333333} 0.6318}    & {\color[HTML]{333333} 0.2207} & {\color[HTML]{333333} 0.0984} &  & {\color[HTML]{333333} 0.7372}    & {\color[HTML]{333333} 0.3980} & {\color[HTML]{333333} 0.2175} \\
{\color[HTML]{333333} \textit{L1}}         & {\color[HTML]{333333} 0.6298}    & {\color[HTML]{333333} 0.2202} & {\color[HTML]{333333} 0.0977} &  & {\color[HTML]{333333} 0.7181}    & {\color[HTML]{333333} 0.3551} & {\color[HTML]{333333} 0.1737} \\
{\color[HTML]{333333} \textit{L2}}         & {\color[HTML]{333333} 0.6306}    & {\color[HTML]{333333} 0.2225} & {\color[HTML]{333333} 0.1015} &  & {\color[HTML]{333333} 0.7254}    & {\color[HTML]{333333} 0.3649} & {\color[HTML]{333333} 0.1783} \\
{\color[HTML]{333333} \textit{CoFiTune}}   & {\color[HTML]{333333} 0.6315}    & {\color[HTML]{333333} 0.2206} & {\color[HTML]{333333} 0.0988} &  & {\color[HTML]{333333} 0.7222}    & {\color[HTML]{333333} 0.3645} & {\color[HTML]{333333} 0.1788} \\ \bottomrule[1.5pt]
\end{tabular}
\caption{Comparison between directly using the test set and our sample-then-paraphrase strategy. We present the result using the 7B model in the finance task.}
\label{tab:evaluation_data_compariso_finance}
\end{table*}

% In this study, our primary objective is to augment the capacity of Language Models (LLMs) to produce content specific to a given domain while preserving their overall knowledge base. However, assessing the automatically generated content of models has consistently posed a challenge, particularly in the context of evaluating LLMs~\cite{zhang2023benchmarking}. As illustrated in Table~\ref{evaluation data comparison}, our observations reveal that relying on instructions from unseen test sets for evaluation results in similar and generally suboptimal performance across various methods. We attribute this trend to the inadequacy of current automatic evaluation metrics in gauging LLMs' output for unseen and open-ended questions~\cite{chang2023survey}.
% reliability
% fine-tuning 方法可能在 static questions (in the training data)上面的表现比较好

The exploration algorithm mentioned in Sec.~\ref{sec:coarse_level} requires extensive experiments conducted to explore the distinct role of each layer range and modules within it.
This necessitates a fast, cost-effective, and accurate automatic evaluation strategy.
For Finance, Law, and CAG, we employ automatic generation metrics encompassing both semantic alignment and n-gram matching, including BERTScore~\citep{bert-score}, Rouge~\citep{lin2004rouge}, and BLEU~\citep{papineni2002bleu}.
% We report the results of the automatic generation metric of Finance under the 7B model in the unseen test set in Table~\ref{tab:evaluation_data_compariso_finance}.
The automatic results for Finance under the 7B model in the unseen test set are presented in Table~\ref{tab:evaluation_data_compariso_finance}. 
% As illustrated in Table~\ref{tab:evaluation_data_compariso_finance}, our observations reveal that the evaluation results of different methods from unseen test set are similar and generally suboptimal performance across various methods.
As depicted in Table~\ref{tab:evaluation_data_compariso_finance}, our observations indicate that the evaluation results across different methods for the unseen test set are similar and generally demonstrate suboptimal performance.
This finding is not unique, as previous studies~\citep{ovadia2023fine,wang2023survey} in question-answering have encountered similar issues\footnote{Fine-tuned models become static data snapshots during training and may swiftly become inadequate for effectively supporting dynamic scenarios, making them less capable of handling unseen domain-specific questions.}, and diverse retrieval-augmented generation (RAG) approaches~\cite{ovadia2023fine,gao2023retrieval} are proposed to enhance LLMs' performance in addressing this issue.
Another possible reason is the inadequacy of current automatic evaluation metrics in gauging LLMs' output for unseen and dynamic questions~\cite{chang2023survey,wang2023survey}.

Consequently, to enhance the reliability of the automatic evaluation results, we advocate for a novel evaluation approach when evaluating Finance, Law, and CAG\footnote{The test set of CAG task comprises both positive and negative samples, and their evaluation methods slightly differ; for more details, refer to the CAG task description in Appendix~\ref{sec:CAG_task}.} tasks: initially sampling instructions from the training set and subsequently paraphrasing them to form the test questions, i.e., sample-then-paraphrase strategy.
Intuitively, this strategy facilitates valid automatic evaluation of LLMs' generation by testing in-domain knowledge while addressing concerns related to data contamination through test instruction rephrasing~\cite{schaeffer2023pretraining, oren2023proving}.

\subsubsection{Validation of the Automatic Results}

% To further validate the effectiveness of the automatic metrics we used in accessing LLMs' generation quality, we randomly sample a set of samples from the financial 7B-CoFiTune's generation result and evaluate their Spec. score with both GPT-4 and human annotators.
To further validate the effectiveness of the automatic metrics used in assessing the quality of LLMs' generation, we randomly sample a set of outputs from the financial \emph{CoFiTune} 7B model and evaluate their Spec. score with both GPT-4 and human annotators.

% For human evaluation, we randomly select 100 samples and ask a qualified annotator to label each output's quality based on the following aspects: accuracy, relevance, and coherence. Each score can range from 1 (very poor) to 5 (very good).
For human evaluation, we randomly select 100 samples and ask a qualified annotator\footnote{The annotator has scored above 600 on the College English Test 6 level (CET-6) and is compensated with 5 RMB per sample.} to assess each output's quality based on aspects such as accuracy, relevance, and coherence. Each score can range from 1 (very poor) to 5 (very good).
For GPT-4 evaluation, we adhere to the same criteria and create a prompt instructing GPT-4 to assign a rating to each model output. An example of our evaluation prompt for GPT-4 is presented in Table~\ref{GPT-4 Prompt}.
% For GPT-4 evaluation, we also follow the aforementioned criterion and curate a prompt to instruct GPT-4 to give a rating to each of the model's outputs. Table~\ref{GPT-4 Prompt} presents an example of our evaluation prompt for GPT-4.

\begin{table*}[h]
\centering
\begin{tabular}{ll}
\hline
Prompt & \begin{tabular}[c]{@{}l@{}}You serve as an impartial evaluator. Please adhere to the criteria outlined below and assess\\ the provided output on a scale ranging from 1 (very poor) to 5 (very good). \\Evaluate based on the following dimensions:\\ Accuracy: Assess the truthfulness and factual correctness of the candidate's response.\\ Relevance: Examine how well the response aligns with the topic of the question.\\ Coherence: Evaluate how seamlessly the response integrates into the context, considering \\consistency with previous statements and overall flow of the answer.\\ Please apply these criteria to the following question and output:\\ Question: \{Question\}\\ Output: \{Output\}\\ \\ Accuracy:\\ Relevance:\\ Coherence:\\ Overall:\end{tabular} \\
\hline
\end{tabular}
\caption{The prompt we adopt for GPT-4 evaluation. We use ``\{Question\}'' and ``\{Output\}'' to replace the specific question and output.}
\label{GPT-4 Prompt}
\end{table*}

Upon acquiring individual scores from both the annotator and GPT-4, we normalize each score to a range between 0 and 1. Subsequently, we employ the Inter Annotator Agreement (IAA) metric to assess the concordance between our automatic Spec. scores and the scores obtained from both the annotator and GPT-4.
As illustrated in Table~\ref{IAA}, our Spec. score reveals a substantial level of agreement with evaluations conducted by both human annotators and GPT-4. This alignment underscores the efficacy and robustness of the Spec. score as a reliable metric for assessing the quality of outputs generated by LLMs in our evaluation setting (mentioned in Appendix~\ref{sec:appendix_new_evaluation_strategy}).

\begin{table*}[h]
\centering
\begin{tabular}{lcc}
\toprule[1.5pt]
   & GPT-4 & Human Annotators \\ \addlinespace[2pt] \hline \addlinespace[2pt]
\textit{k}  & 0.689      &   0.653               \\
\textit{$p_0$} &   0.800    &   0.756               \\ \bottomrule[1.5pt]
\end{tabular}
\caption{Inter-Annotator Agreement (IAA) measured by Cohen’s Kappa, and the agreement rate between GPT-4 score, human annotators and our Spec. score.}
\label{IAA}
\end{table*}

% \appendix

% \section{Example Appendix}
% \label{sec:appendix}

% This is an appendix.

\end{document}